\documentclass[journal]{new-aiaa}
\usepackage[utf8]{inputenc}
\usepackage{textcomp}

\usepackage{graphicx}

\usepackage{amsmath,amssymb,amsthm,mathtools}
\usepackage{tikz}
\usetikzlibrary{shapes.geometric, arrows.meta, positioning, calc}
\usepackage[version=4]{mhchem}
\usepackage{siunitx}
\usepackage{longtable,tabularx}
\usepackage{geometry}
\usepackage{xcolor}
\usepackage{algorithm}
\usepackage{algpseudocode}
\usepackage{float}
\usepackage{subcaption}
\usepackage{pgfcalendar}
\usepackage{pgfplotstable}
\usepackage{booktabs}
\usepackage{graphicx}
\usepackage{subcaption}
\usepackage{xcolor}
\usepackage{rotating}
\usepackage{caption}
\usepackage{tikz}
\usetikzlibrary{patterns,shapes.misc}
\usepackage{pgfplots}
\usepgfplotslibrary{dateplot}
\usepgfplotslibrary{groupplots}
\pgfplotsset{compat=1.18}

\usepackage{graphicx}
\usepackage{comment}

      \usepackage{booktabs}
      \usepackage{caption}
      \usepackage{csvsimple}
\usepackage{float}

\newtheorem{theorem}{Theorem}
\newtheorem{lemma}{Lemma}
\newtheorem{proposition}{Proposition}
\newtheorem{corollary}{Corollary}
\newtheorem{remark}{Remark}

\usepackage{pgfplotstable}
\usepgfplotslibrary{groupplots, fillbetween}
\usepgfplotslibrary{dateplot}

\pgfplotsset{compat=1.18}

\definecolor{cbTruth}{HTML}{333333}
\definecolor{cbGuess}{HTML}{56B4E9}
\definecolor{cbOpt}{HTML}{2c7bb6}
\definecolor{cbObs}{HTML}{980043}
\definecolor{cbError}{HTML}{CC79A7}

\definecolor{romForecastColor}{HTML}{0072B2}      
\definecolor{obsColor}{HTML}{000000}              
\definecolor{assimColor}{HTML}{E69F00}            
\definecolor{truthColor}{HTML}{009E73}            
\definecolor{corrWindowColor}{HTML}{CC79A7}       
\definecolor{triggerArrowColor}{HTML}{cc0000}     
\definecolor{triggerthreshold}{HTML}{6a6aff}
\definecolor{fullstatecolor}{HTML}{D55E00}
\definecolor{trainingtesting}{HTML}{469374}
\definecolor{phaseboxgray}{HTML}{EFEFEF}          
\definecolor{phaseboxxbrown}{HTML}{F5E6D3}        
\definecolor{obslabelblue}{HTML}{E6F2FF}          
\definecolor{gridcolor}{HTML}{E0E0E0}             
\definecolor{timestamps}{HTML}{5e239d}
\definecolor{EqColor}{HTML}{B6A7FF}
\definecolor{P1Color}{HTML}{003A7D}
\definecolor{P2Color}{HTML}{BA6B4D}
\definecolor{P3Color}{HTML}{FFD166}
\definecolor{ICDAColor}{HTML}{E25759}

\tikzset{
    lineTrue/.style={color=cbTruth},
    lineGuess/.style={color=cbGuess},
    lineAssim/.style={color=cbOpt},
    markObs/.style={color=cbObs}
}

\hypersetup{
    colorlinks=true,
    linkcolor=black,
    citecolor=black,
    urlcolor=black
}

\usepackage{authblk}

\newcommand{\ak}[1]{{\color{black}{#1}}}

\title{Extrapolative Operator Inference with State Correction: An Error-Triggered Windowed 4D-Var Framework}

\author[1]{Madduri Sai Hemanth Yagna Kasyap\footnote{Graduate Student}}
\author[1]{Manikandan Mathur\footnote{Professor}}
\author[1]{Aniketh Kalur\footnote{Assistant Professor}}
\affil[1]{Department of Aerospace Engineering, Indian Institute of Technology Madras}

\date{}

\begin{document}
\maketitle

\begin{abstract}
    Reduced-order models (ROMs) can lose accuracy over long forecast horizons, particularly outside their training regime. We show that accurate long-horizon prediction remains possible when the ROM retains sufficient short-horizon fidelity and is intermittently corrected using sparse observations. We achieve this by coupling Operator Inference (OpInf) with correction-windowed 4D-Variational data assimilation and strategically selected measurements. Assimilation is triggered only when the model--observation discrepancy exceeds a prescribed threshold, while the polynomial structure of OpInf provides an explicit reduced-order Jacobian and adjoint for efficient gradient evaluation entirely in reduced coordinates. Sensor locations are selected using column-pivoted QR decomposition of the reduced basis.

We derive a sufficient condition that determines when the reduced-order OpInf gradient remains a valid descent direction for the full-order objective within the reduced subspace. The analysis shows that shorter correction horizons limit the amplification of forward-model and adjoint errors entering this condition, explaining why reduced-order assimilation can remain effective despite poor long-horizon ROM predictions.

The framework is demonstrated on the Duffing oscillator, the one-dimensional viscous Burgers equation, and high-dimensional sea-surface-temperature prediction. Across these problems, intermittent assimilation extends the useful predictive horizon of the ROM, while sparse strategically placed measurements recover performance close to full-state assimilation. These results show that long-horizon ROM accuracy is not required for effective prediction when sufficient short-horizon fidelity is retained for repeated state correction.

\end{abstract}

\section{Introduction}\label{sec:intro}

Reduced-order models (ROMs) can provide substantial computational savings for high-dimensional dynamical systems, but their predictive accuracy often deteriorates when deployed outside the regime represented by the training data. This limitation is particularly severe for long-horizon prediction, where model errors can accumulate and eventually render an otherwise useful ROM unreliable. Rather than rebuilding the reduced basis, retraining the reduced operators, or returning to the full-order model, we ask whether an existing ROM can continue to be used once it enters such an extrapolative regime. In this work, we demonstrate that future measurements from sparse sensors can be used to intermittently correct the evolving reduced trajectory, allowing the same pre-trained ROM to remain predictive beyond the regime in which its autonomous forecast is accurate. The central premise is that a ROM need not remain accurate over the entire prediction horizon: sufficient short-horizon fidelity can be exploited to perform corrective updates that limit the accumulation of long-term prediction error. This strategy is particularly compelling for digital twin applications of complex high-dimensional systems, where lightweight surrogates must maintain physical trajectory alignment over extended deployments using streaming, sparse observations.

Projection-based reduced-order modeling constructs a low-dimensional state
representation from dominant modes of the full-order dynamics. Proper
Orthogonal Decomposition (POD), for example, extracts dominant spatial
structures from snapshot data and provides a reduced basis onto which the
full-order state can be projected
~\cite{sirovich1987turbulence,benner2015survey,hesthaven2016certified}.
Classical projection-based ROMs typically require access to the governing
full-order operators in order to project the dynamics onto this reduced
subspace. This requirement can limit their applicability when the full-order
model is available only through legacy codes, commercial software, or other
black-box simulation environments.

Non-intrusive ROM methods address this limitation by learning reduced dynamics directly from data. Dynamic Mode Decomposition and related approaches construct linear approximations associated with the Koopman operator~\cite{schmid2010dmd,rowley2009spectral}, while Operator Inference (OpInf), introduced by Peherstorfer and Willcox~\cite{peherstorfer2016data}, learns polynomial reduced-order operators from state trajectory data without requiring access to the underlying full-order operators. Since its introduction, OpInf has been extended to accommodate broader classes of nonlinear systems, limited or noisy training data, stability constraints, and parametric dependence. Examples include lifting transformations~\cite{qian2020lift}, trajectory-based regression objectives~\cite{uy2022rollouts}, physics-informed constraints~\cite{sawant2023physics}, and parametric formulations~\cite{mcquarrie2021parametric}. More recent work has also considered iterative and nested learning strategies~\cite{aretz2026nested}. Kramer et al.~\cite{kramer2024learning} provide a broader review of developments in Operator Inference.

Despite these advances, a practical difficulty remains when a learned ROM is
deployed beyond the regime represented by its training data: a fixed reduced
model may progressively lose predictive accuracy as its trajectory evolves
into an extrapolative regime. Rather than adapting the basis, retraining the
reduced operators, or enlarging the model class during deployment, we consider
whether subsequently acquired observations can instead be used to correct the
evolving reduced state while retaining the original learned model.

The proposed framework exploits the distinction between long-horizon predictive
accuracy and short-horizon model fidelity. Even after the autonomous ROM
trajectory begins to deteriorate outside the regime represented by the
training data, its dynamics may remain sufficiently accurate over shorter
time horizons to support state correction from subsequently acquired
observations. We therefore retain the pre-trained reduced model throughout
deployment: the POD basis, learned OpInf operators, and underlying model
parameters remain fixed, while variational data assimilation is used only to
correct the evolving reduced state. In this work, the extrapolative regime
thus refers to state-trajectory evolution beyond that represented in the
training data under fixed model parameters and learned dynamics. Persistent
parameter drift or changes in the governing dynamics are not estimated by the
present framework and would instead require joint state--parameter estimation
or online model adaptation. Within this setting, the polynomial structure of
OpInf is particularly advantageous because the reduced-space model Jacobian
can be evaluated analytically and inexpensively, enabling low-dimensional
adjoint equations to provide the gradients required for variational state
correction.

Variational data assimilation combines dynamical models with observations by solving an optimisation problem constrained by the governing dynamics. Following the early formulations of Sasaki~\cite{Sasaki1958} and the development of adjoint-based variational methods~\cite{LeDimet1982,lewis1985variational,courtier1985_exp,le1986variational,talagrand1987variational}, four-dimensional variational assimilation (4D-Var) has become a standard approach for estimating an initial state by incorporating observations distributed over a finite time window; see Freitag~\cite{freitag2009numerical} for a detailed mathematical treatment. Because the model dynamics explicitly constrain the optimisation trajectory, 4D-Var can extract information from temporally distributed measurements but requires repeated forward and adjoint integrations, making the computational cost substantial for high-dimensional systems. Incremental 4D-Var partially alleviates this difficulty by solving approximate linearised problems within inner loops~\cite{Courtier1994-yr}, but the underlying optimisation remains closely tied to the high-dimensional dynamical model.

This computational challenge has motivated several forms of reduced-order variational assimilation. Some approaches reduce the control space while retaining full-order forward and adjoint models~\cite{Parrish1992-nk,robert2005reduced}, whereas others reduce the dynamical model itself. Cao et al.~\cite{cao2007reduced}, for example, developed a POD--Galerkin ROM and corresponding reduced adjoint for 4D-Var, while Daescu and Navon~\cite{daescu2006reduced} developed reduced second-order adjoint methods. K{\"a}rcher et al.~\cite{karcher2018reduced} subsequently introduced certified reduced-basis approximations for strong- and weak-constraint 4D-Var, including reduced approximations of the state and adjoint variables together with efficiently computable a posteriori error bounds. More recently, differentiable data-driven surrogates have also been used within variational assimilation; Maulik et al.~\cite{Maulik2022}, for example, employed an encoder--decoder LSTM surrogate with gradients obtained using automatic differentiation. Arcucci et al.~\cite{arcucci2023reduced} developed Reduced Order Data Assimilation (RODA), combining POD reduction, radial-basis-function interpolation, and 3D-Var assimilation.

These studies demonstrate the computational advantages of performing
variational assimilation in reduced spaces, where the ROM is used primarily
as a surrogate for lowering the cost of the underlying assimilation problem.
The objective here is different. We use assimilation to maintain the
predictive utility of a fixed learned ROM after its autonomous trajectory has
begun to deteriorate outside the regime represented by the training data.
Throughout deployment, the reduced basis and learned operators remain
unchanged; only the evolving reduced state is corrected using subsequently
acquired observations. The resulting framework therefore treats data
assimilation not simply as a reduced-cost alternative to full-order 4D-Var,
but as a mechanism for limiting the accumulation of state-trajectory error and
extending the useful predictive horizon of an existing ROM in an
extrapolative regime.

Two additional considerations are important for maintaining predictive accuracy
in the extrapolative regime. First, although the autonomous ROM may lose
accuracy over long prediction horizons, its short-horizon dynamics and
sensitivities may remain sufficiently accurate to support state correction.
This motivates performing variational assimilation over short correction
windows, which limits the amplification of forward-model and adjoint errors
over the optimisation horizon. Related receding-horizon strategies repeatedly
solve finite-window optimisation problems as the system evolves, an idea
closely connected to moving-horizon control~\cite{Richalet1978-xa}. The
sensitivity of finite-window variational methods to accumulated model error has
also been documented in the data-assimilation literature
~\cite{gilmour2001linear,Tremolet2007}. In the present framework, these short
correction windows are combined with an event-triggered strategy so that
assimilation is performed only when the model--observation disagreement exceeds
a prescribed threshold. This avoids unnecessary optimisation solves while
allowing accumulated trajectory error to be corrected when required, consistent
with the broader idea of event-triggered estimation~\cite{Lieventtriggered2022}.
When observations are available only at a limited number of physical locations,
we use column-pivoted QR (CPQR) decomposition of the POD basis to select
informative sensor locations~\cite{BusingerGolub1965,manohar2018data}, providing
a practical sparse observation map for the reduced-order assimilation.

During deployment, the pre-trained ROM is propagated without modifying its
POD basis or learned OpInf operators. Its predictions are compared with the
available measurements, and when the model--observation mismatch exceeds a
prescribed threshold, a finite-window 4D-Var problem is solved in reduced
coordinates to correct the evolving reduced state. The corrected state is then
used to resume forecasting with the same ROM. Sparse measurements may be used
through the CPQR-based observation map introduced above, while the explicit
polynomial structure of OpInf provides the reduced Jacobian and adjoint needed
for the variational correction. The principal contributions of this work are:

\begin{enumerate}

\item We develop an intermittent reduced-order state-correction framework for
extending the useful predictive horizon of a fixed, pre-trained ROM into
extrapolative regimes, without retraining the reduced operators or modifying
the reduced basis.

\item We formulate the correction entirely in reduced coordinates using
classical polynomial Operator Inference, whose explicit structure provides
inexpensive reduced-order Jacobian and adjoint evaluations for the variational
assimilation problem.

\item We derive a sufficient condition under which the reduced-order OpInf
gradient defines a descent direction for the full-order objective restricted
to the POD subspace. We further show that restricting the correction-window
horizon suppresses the amplification of forward and adjoint errors entering
this condition, providing analytical support for short-window state correction
in the extrapolative regime.

\item We combine the reduced-order correction with error-triggered assimilation
and CPQR-based sparse sensor placement, so that corrective updates are performed
only when required and can be driven by a limited number of informative
measurements.

\end{enumerate}

We evaluate the framework on three problems chosen to progressively test the
mechanism and scope of the proposed state-correction strategy. A nonlinear
oscillator first provides a controlled setting in which the accumulation of
forecast error, the effect of intermittent correction, and the quantities
appearing in the theoretical analysis can be examined directly. Burgers'
equation then tests the approach on a nonlinear, transport-dominated PDE in
which the ROM is propagated beyond the regime represented by its training data.
Finally, sea-surface-temperature observations provide a realistic
high-dimensional setting in which the same correction principle is assessed
using sparse measurements. Across these examples, we examine how effectively
intermittent state correction extends the useful predictive horizon of a fixed
ROM beyond the regime in which its autonomous forecast remains reliable, while
also quantifying the associated assimilation frequency, sensing, and
computational requirements.

The remainder of the paper is organised as follows. Section~\ref{sec:math} introduces the mathematical formulation, including Operator Inference, variational data assimilation, and CPQR-based sensor placement. Section~\ref{sec:proposed} presents the proposed reduced-order assimilation framework and the analysis connecting short assimilation windows with extrapolative forecasting. Section~\ref{sec:illustrative} illustrates the correction mechanism using the nonlinear oscillator, and Section~\ref{sec:results} presents results for Burgers' equation and sea-surface-temperature data. Section~\ref{sec:conclusion} summarises the principal findings, limitations, and directions for future work.

\section{Problem Setting and Methodological Background}\label{sec:math}

To motivate the proposed framework, we first introduce the principal ingredients on which it is built: a high-dimensional dynamical system with sparse observations, a POD representation of its dominant state subspace, a non-intrusive Operator Inference model for the reduced dynamics, and variational data assimilation for state correction. Our purpose is not to review these methods exhaustively, but to establish notation and identify the computational and observational challenges that motivate the reduced-order assimilation framework developed in Section~\ref{sec:proposed}.

We consider the original discrete-time high-fidelity dynamical system, also 
referred to as the full-order model (FOM), given by
\begin{align}
    \mathbf{x}_{k+1} &= \mathcal{M}(\mathbf{x}_k,\mathbf{u}_k),
    \label{eq:fom}\\
    \mathbf{y}_k &= \mathcal{H}(\mathbf{x}_k) + \boldsymbol{\rho}_k.
    \label{eq:fom_op}
\end{align}
Here, $\mathbf{x}_k \in \mathbb{R}^n$ denotes the full-order state at time
$t_k$, $\mathbf{u}_k$ represents external inputs or parameters, and
$\mathcal{M}$ is the discrete-time evolution operator. Measurements
$\mathbf{y}_k \in \mathbb{R}^m$ are obtained through the observation operator
$\mathcal{H}$ with additive noise
$\boldsymbol{\rho}_k \sim \mathcal{N}(\mathbf{0},\mathbf{R})$.
Both $\mathcal{M}$ and $\mathcal{H}$ may, in general, be nonlinear, and we
consider the regime $m \ll n$ in which only sparse observations of the
high-dimensional state are available.

Repeated evaluation of the FOM becomes computationally expensive for
large-scale systems, particularly in variational data assimilation where
forward and sensitivity evaluations are required within an iterative
optimisation procedure. We assume that the trajectories of interest admit an
accurate approximation in a low-dimensional subspace of dimension
$r \ll n$. This motivates replacing the full-order state evolution with a
reduced representation constructed from dominant data-driven modes. In the
present work, this subspace is obtained using Proper Orthogonal Decomposition
(POD), a standard projection-based model-reduction technique
\cite{benner2015survey}.

\subsection{Projection-based Reduced-Order Model via Proper Orthogonal Decomposition}
\label{sec:rom}

Proper Orthogonal Decomposition (POD) is used to construct a low-dimensional
basis from representative full-order state data. Given $N$ state snapshots
$\{\mathbf{x}_k\}_{k=1}^{N}$, we assemble the snapshot matrix
$\mathbf{X}\in\mathbb{R}^{n\times N}$ and compute its singular value
decomposition (SVD),
$\mathbf{X}=\mathbf{U}\boldsymbol{\Sigma}\mathbf{V}^{\top}$, where
$\mathbf{U}\in\mathbb{R}^{n\times p}$,
$\boldsymbol{\Sigma}\in\mathbb{R}^{p\times p}$, and
$\mathbf{V}\in\mathbb{R}^{N\times p}$, with $p=\min(n,N)$ for the economy
SVD. The singular values
$\sigma_1\geq\sigma_2\geq\cdots\geq\sigma_p\geq0$
rank the corresponding left singular vectors according to their contribution
to the snapshot data. Denoting $\mathbf{u}_i:=\mathbf{U}(:,i)$, the POD basis
is formed by retaining the first $r$ left singular vectors:
\begin{equation}
\mathbf{\Phi} = \begin{bmatrix} \mathbf{u}_1 & \mathbf{u}_2 & \cdots & \mathbf{u}_r \end{bmatrix} \in \mathbb{R}^{n \times r}
\label{eq:pod_basis}
\end{equation}
where $r\ll n$ is chosen so that the retained subspace provides an adequate
representation of the trajectories of interest.

The corresponding reduced coordinates are obtained by projecting the
full-order state onto the POD subspace:
\begin{equation}
\widehat{\mathbf{x}}_k = \mathbf{\Phi}^\top \mathbf{x}_k, \quad \widehat{\mathbf{x}}_k \in \mathbb{R}^r.
\label{eq:projection}
\end{equation}
Evolution in these reduced coordinates may then be represented by
\begin{align}
\begin{split}
	\widehat{\mathbf{x}}_{k+1}  &= \widehat{\mathcal{M}} (\widehat{\mathbf{x}}_k, \mathbf{u}_k),\\
	\mathbf{y}_k &= \widehat{\mathcal{H}}(\widehat{\mathbf{x}}_k).
	\end{split}
\label{eq:rom}
\end{align}
where $\widehat{\mathcal{M}}$ and $\widehat{\mathcal{H}}$ denote the reduced
dynamics and observation operators, respectively.

POD therefore provides the reduced coordinate representation but does not, by itself, determine the evolution operator $\widehat{\mathcal{M}}$. In settings where direct projection of the governing equations is unavailable or undesirable, the reduced dynamics must instead be inferred from data. We adopt Operator Inference for this purpose, enabling a non-intrusive construction of $\widehat{\mathcal{M}}$ directly from state snapshots.

\subsection{Operator Inference: Non-intrusive Model Reduction}
\label{sec:opinf}

POD provides a low-dimensional representation of the state but does not, by
itself, determine the corresponding reduced dynamics. When intrusive projection
of the governing equations is unavailable or undesirable, these dynamics can
instead be inferred directly from state data. Operator Inference (OpInf),
introduced by Peherstorfer and Willcox~\cite{peherstorfer2016data}, provides
such a non-intrusive construction by learning reduced operators from snapshot
data without requiring access to the full-order operators or source code.

For systems whose governing dynamics admit a polynomial representation,
projection-based model reduction preserves the polynomial structure in the
reduced coordinates. Motivated by this property, we employ a linear-quadratic
OpInf representation of the reduced dynamics. The present work focuses on this
polynomial setting, although the Operator Inference framework can also be
extended to systems containing non-polynomial terms. The reduced dynamics are
therefore expressed in terms of unknown operators to be inferred from data:
\begin{equation}
\begin{aligned}
\widehat{\mathbf{x}}_{k+1} &= \widehat{\mathcal{M}}(\widehat{\mathbf{x}}_k) \\
&= \widehat{\mathbf{A}}\widehat{\mathbf{x}}_k + \widehat{\mathbf{F}}(\widehat{\mathbf{x}}_k \mathbin{{\tilde{\otimes}}} \widehat{\mathbf{x}}_k), \\
\mathbf{y}_k &= \widehat{\mathbf{C}} \widehat{\mathbf{x}}_k
\end{aligned}
\label{eq:opinf_poly}
\end{equation}
where $\widehat{\mathbf{A}} \in \mathbb{R}^{r \times r}$ and
$\widehat{\mathbf{F}} \in \mathbb{R}^{r \times \frac{r(r+1)}{2}}$ denote the
linear and quadratic reduced operators, respectively, and
$\mathbin{{\tilde{\otimes}}}$ denotes the symmetric Kronecker product.

The observation operator is not inferred through the OpInf regression. Let
$\mathbf{C}\in\mathbb{R}^{m\times n}$ denote the full-state measurement matrix
that selects the physical sensor locations. Since
$\mathbf{x}_k\approx\mathbf{\Phi}\widehat{\mathbf{x}}_k$, the corresponding
observation operator in reduced coordinates is
$\widehat{\mathbf{C}}=\mathbf{C}\mathbf{\Phi}\in\mathbb{R}^{m\times r}$.
Thus, $\mathbf{C}$ acts in the full-order space and contains the physical
sampling locations, whereas $\widehat{\mathbf{C}}$ maps the reduced state
directly to the corresponding measurements. The construction of
$\mathbf{C}$ through sparse sensor placement is discussed subsequently in
Section~\ref{sec:sensor_placement}.

Let $N^{\text{total}}$ denote the total number of snapshots available for
learning the operators, partitioned into training ($N^{\text{train}}$),
validation ($N^{\text{val}}$), and testing ($N^{\text{test}}$) subsets, such
that
$N^{\text{train}}+N^{\text{val}}+N^{\text{test}}=N^{\text{total}}$.
The POD basis is constructed from the training data, and the corresponding
reduced states are assembled into
$\widehat{\mathbf{X}}\in\mathbb{R}^{r\times N^{\text{train}}}$, with their
time-shifted values collected in
$\widehat{\mathbf{X}}'\in\mathbb{R}^{r\times N^{\text{train}}}$.
The reduced operators are obtained from a Tikhonov-regularised least-squares
problem:
\begin{equation}
	\min_{\widehat{\mathbf{A}}, \widehat{\mathbf{F}}} \left\| (\widehat{\mathbf{X}}')^T - \left[ \widehat{\mathbf{X}}^T \quad (\widehat{\mathbf{X}} \mathbin{\widetilde{\otimes}} \widehat{\mathbf{X}})^T \right] \begin{pmatrix} \widehat{\mathbf{A}}^T \\ \widehat{\mathbf{F}}^T \end{pmatrix} \right\|_F^2 + \zeta_A \|\widehat{\mathbf{A}}\|_F^2 + \zeta_F \|\widehat{\mathbf{F}}\|_F^2.
\label{eq:opinf_opt}
\end{equation}
Here, $\zeta_A$ and $\zeta_F$ are non-negative regularisation parameters
associated with the linear and quadratic operators, respectively. We select
$(\zeta_A,\zeta_F)$ using validation-trajectory error: the regression problem
in Eq.~\eqref{eq:opinf_opt} is solved for each candidate parameter pair using
the training data, and the resulting ROM is evaluated on the validation
trajectory using the same initial condition. The parameter pair producing the
lowest validation error is retained.

An important consequence of the polynomial OpInf representation is that the
Jacobian of the reduced dynamics can be evaluated directly and at low
computational cost. This property is particularly useful in the present work,
since the variational data assimilation formulation requires repeated
sensitivity and adjoint evaluations. The resulting reduced polynomial model
therefore provides both an inexpensive forward model and an explicitly
accessible reduced-order Jacobian for the 4D-Var formulation introduced next.

\subsection{4D-Var Data Assimilation on the Full-Order Model}\label{sec:4dvar_fom}

Four-dimensional variational (4D-Var) data assimilation seeks a model trajectory
that reconciles available observations with a prior estimate over a prescribed
assimilation window. In the strong-constraint formulation considered here, the
model dynamics are enforced exactly and the initial condition serves as the
control variable. Following the standard formulation described in
Freitag~\cite{freitag2009numerical}, the full-order 4D-Var problem is written as
\begin{equation}
\begin{aligned}
	& \min_{\mathbf{x}_0} \mathcal{J}(\mathbf{x}_0) = \frac{1}{2} \| \mathbf{x}_0 - \mathbf{x}^b_0 \|^2_{\mathbf{B}^{-1}} + \frac{1}{2} \sum_{k=0}^{N^{\text{asim}}} \| \mathbf{y}_k - \mathbf{H}(\mathbf{x}_k) \|^2_{\mathbf{R}^{-1}} \\
	& \text{subject to} \quad \mathbf{x}_{k+1} = \mathcal{M}(\mathbf{x}_k).
\end{aligned}
\label{eqn:4dvar_cost_fom}
\end{equation}
Here, $\mathbf{x}^b_0$ denotes the background estimate of the initial state,
with associated covariance matrix $\mathbf{B}$, while $\mathbf{R}$ denotes the
observation-error covariance. In the present work, we assume a linear
observation model,
$\mathbf{H}(\mathbf{x}_k):=\mathbf{C}\mathbf{x}_k$, where
$\mathbf{C}\in\mathbb{R}^{m\times n}$ maps the full-order state to the
$m$ available measurements. Optimisation with respect to $\mathbf{x}_0$
therefore adjusts the initial condition so that the resulting model trajectory
remains consistent with both the background estimate and the observations
within the assimilation window.

Since the observation terms depend on $\mathbf{x}_0$ through the entire model
trajectory, gradient-based solution of Eq.~\eqref{eqn:4dvar_cost_fom}
requires sensitivities of the trajectory with respect to the initial state.
These sensitivities are evaluated efficiently through the discrete adjoint
recursion~\cite{freitag2009numerical},
\begin{equation}
\lambda_k = \mathbf{C}^T \mathbf{R}^{-1} (\mathbf{C}\mathbf{x}_k - \mathbf{y}_k) + (\nabla_{\mathbf{x}_k} \mathcal{M}_k)^T \boldsymbol{\lambda}_{k+1}, \quad \text{for}~k = N^{\text{asim}},N^{\text{asim}}-1, \ldots, 0.
\label{eqn:adj_eqn_fom}
\end{equation}
where
$\nabla_{\mathbf{x}_k}\mathcal{M}_k
= \partial\mathcal{M}(\mathbf{x}_k)/\partial\mathbf{x}_k$
denotes the Jacobian of the full-order evolution operator evaluated along the
forward trajectory. Starting from the terminal condition
$\boldsymbol{\lambda}_{N^{\text{asim}}+1}=0$, the adjoint variables are propagated backward
through the assimilation window. The gradient of the 4D-Var objective with
respect to the initial condition is then
\begin{equation}
\nabla_{\mathbf{x}_0} \mathcal{J} = \mathbf{B}^{-1}(\mathbf{x}_0 - \mathbf{x}^b_0) + \boldsymbol{\lambda}_0
\label{eqn:gradient_fom}
\end{equation}
and can be supplied directly to a gradient-based optimisation algorithm.

Although the adjoint formulation avoids explicitly forming trajectory
sensitivities with respect to every component of the initial condition, a
full-order 4D-Var optimisation still requires repeated forward integrations
of the FOM and backward integrations of the associated adjoint model. Moreover,
the adjoint recursion requires repeated application of the transpose of the
full-order Jacobian,
$(\nabla_{\mathbf{x}_k}\mathcal{M}_k)^T$. For large-scale systems, these
forward and adjoint evaluations can dominate the computational cost of the
assimilation procedure. In non-intrusive or black-box settings, direct access
to the full-order Jacobian or adjoint implementation may also be unavailable.

These limitations motivate carrying out the variational correction directly
in reduced coordinates. The polynomial Operator Inference model introduced
above provides both an inexpensive reduced-order forward model and direct
access to its low-dimensional Jacobian, thereby avoiding repeated full-order
forward and adjoint evaluations during online assimilation. A second practical
consideration is that measurements are typically available only at a limited
number of physical locations. The selection of these locations determines how
effectively the reduced state can be constrained and is addressed next.

\subsection{Sensor Placement via Column-Pivoted QR}\label{sec:sensor_placement}

For the reduced-order assimilation framework, measurements need only constrain
the $r$-dimensional reduced state rather than all $n$ components of the
full-order state. Let $\boldsymbol{\Phi}_r\in\mathbb{R}^{n\times r}$ denote the
POD basis and let $\mathbf{C}\in\mathbb{R}^{m\times n}$ denote a sampling
matrix that selects $m$ physical measurement locations. The corresponding
observation operator in reduced coordinates is
\[
\mathbf{H}=\mathbf{C}\boldsymbol{\Phi}_r\in\mathbb{R}^{m\times r},
\]
which maps the reduced state directly to the available sensor measurements.

For unique instantaneous reconstruction of an arbitrary reduced state in the
noise-free linear observation setting, $\mathbf{H}$ must have full column rank,
i.e., $\operatorname{rank}(\mathbf{H})=r$. This requires at least $r$ scalar
measurements, so that $m\geq r$. In the sparse-sensor configurations considered
in this work, we deliberately choose the limiting case $m=r$, thereby using
the minimum number of instantaneous measurements for which the reduced
observation map can be full rank. The remaining task is therefore to determine
which $r$ physical locations should be sampled so that the corresponding rows
of the POD basis remain linearly independent and yield a well-conditioned
reduced observation operator.

We select these locations using a column-pivoted QR (CPQR) decomposition of
$\boldsymbol{\Phi}_r^\top$, following Businger and Golub~\cite{BusingerGolub1965}
and its application to sparse sensor placement by Manohar et
al.~\cite{manohar2018data}. The decomposition is written as
\begin{equation}
\boldsymbol{\Phi}_r^\top \mathbf{P}
=
\mathbf{Q}\mathbf{T},
\label{eq:pivoted_qr}
\end{equation}
where $\mathbf{P}$ is a permutation matrix,
$\mathbf{Q}\in\mathbb{R}^{r\times r}$ is orthogonal, and
$\mathbf{T}\in\mathbb{R}^{r\times n}$ is upper trapezoidal. Column pivoting
produces a rank-revealing ordering of the columns of
$\boldsymbol{\Phi}_r^\top$, equivalently the rows of
$\boldsymbol{\Phi}_r$, and therefore provides a greedy criterion for selecting
informative physical measurement locations.

Since $m=r$ in the sparse-sensor experiments considered here, the first $r$
pivots define the selected measurement locations, and the corresponding rows
are used to construct $\mathbf{C}\in\mathbb{R}^{m\times n}$. Since $\boldsymbol{\Phi}_r$ has rank $r$, the objective is to select $r$
rows that retain this rank while avoiding a poorly conditioned sampled basis.
CPQR provides a rank-revealing greedy selection of these rows, yielding the
square reduced observation operator $\mathbf{H}=\mathbf{C}\boldsymbol{\Phi}_r\in\mathbb{R}^{r\times r}$.
Sensor count and sensor placement therefore play distinct roles in the present
framework. The choice $m=r$ ensures that the instantaneous reduced observation problem
is not structurally underdetermined: provided
$\mathbf{C}\boldsymbol{\Phi}_r$ has full rank, distinct reduced states produce
distinct measurement vectors. This condition is imposed here for the
sparse-sensor experiments and is not asserted to be necessary for state
estimation over an assimilation window, where observations accumulated through
the dynamics may provide additional information. Configurations with $m>r$ are permissible and provide redundant
measurements, including the full-state observation cases used as reference
benchmarks; however, the sparse-sensor experiments in this work focus on
$m=r$.

The preceding developments establish the principal components of the proposed
framework. POD provides a low-dimensional representation of the full-order
state, while Operator Inference supplies a non-intrusive reduced dynamical
model with an explicitly accessible Jacobian. The 4D-Var formulation provides
the mechanism for incorporating observational information through trajectory
correction, and CPQR determines a sparse set of physical measurement locations
through which the reduced state can be constrained. In the next section, these
components are combined into a reduced-order variational data assimilation
framework for correcting model trajectories during deployment.

\section{Proposed Approach: Windowed 4D-Var Data Assimilation with OpInf}
\label{sec:proposed}

Although OpInf ROMs provide inexpensive forecasts, their 
trajectories may progressively lose accuracy when propagated beyond the
regime represented by the training data. We address this limitation using
correction-windowed 4D-Var to intermittently re-estimate the evolving reduced
state from observations while retaining the learned operators and reduced basis
throughout deployment.

The analysis below examines when such reduced-order variational corrections
remain consistent with the corresponding full-order objective. In particular,
we bound the discrepancy between the reduced OpInf gradient and the full-order
gradient projected onto the POD subspace, accounting for forward-model error,
tangent-linear-model mismatch, and adjoint propagation. These bounds show how
restricting the correction horizon limits the amplification of errors entering
the reduced variational gradient, thereby providing theoretical support for
short-window state correction in the extrapolative regime.

\subsection{Mathematical Formulation}

The FOM has state dimension $n$, and we learn an OpInf-based ROM of dimension $r \ll n$ from the data, as described in Section~\ref{sec:math}. Our approach operates entirely in reduced coordinates, where we perform assimilation. The goal is to use 4D-Var to correct OpInf ROM drift in this reduced space, enabling computational efficiency while maintaining accuracy through the assimilation of observed data.

Here, if the length of the window of correction is represented by $N_{\text{window}}$ and ${k_{corr}}$ is the time step at which the correction happens, we formulate the reduced-order 4D-Var problem as a constrained optimization. The reduced dynamics are governed by the OpInf-learned model, 
\begin{equation}
\widehat{\mathbf{x}}_{k+1}
=
\widehat{\mathcal{M}}(\widehat{\mathbf{x}}_k),
\qquad
k=k_{\mathrm{corr}},\ldots,k_{\mathrm{obs}}-1.
\label{eqn:reduced_model}
\end{equation}
and we seek to minimize a cost functional that balances background error and observation misfit over a local correction window ending at observation time $k_{obs} = k_{corr} + N_{\text{window}} - 1$:
\begin{equation}
\begin{aligned}
&\min_{\widehat{\mathbf{x}}_{k_{corr}}} \widehat{\mathcal{J}}(\widehat{\mathbf{x}}_{k_{corr}}) = \frac{1}{2} \| \widehat{\mathbf{x}}_{k_{corr}} - \widehat{\mathbf{x}}^b_{k_{corr}} \|^2_{\widehat{\mathbf{B}}^{-1}} + \frac{1}{2} \sum_{k=k_{corr}}^{k_{obs}} \| \mathbf{y}_k - \mathbf{H}\widehat{\mathbf{x}}_k \|^2_{\mathbf{R}^{-1}} \\
&\text{subject to} \ak{\quad \widehat{\mathbf{x}}_{k+1} = \widehat{\mathcal{M}}(\widehat{\mathbf{x}}_k) \qquad
k=k{\rm corr},\ldots,k_{\rm obs}-1 }.
\end{aligned}
\label{eqn:4dvar_cost}
\end{equation}
Here, $\mathbf{H} = \mathbf{C}\mathbf{\Phi} \in \mathbb{R}^{m \times r}$ is the observation operator in reduced coordinates (determined by the measurement matrix $\mathbf{C}$ from section~\ref{sec:sensor_placement} and the POD basis $\mathbf{\Phi}$), $\widehat{\mathbf{B}}$ and $\mathbf{R}$ are background and observation error covariances in reduced space, and the dynamics constraint is given by Equation~\eqref{eqn:reduced_model}.

To enforce the dynamics constraint, we form the reduced-space Lagrangian:
\begin{equation}
\widehat{\mathcal{L}}
= \widehat{\mathcal{J}}(\widehat{\mathbf{x}}_{k_{corr}})
+ \sum_{k=k_{corr}}^{k_{obs}-1}
\widehat{\boldsymbol{\lambda}}_{k+1}^T
\Bigl(\widehat{\mathcal{M}}(\widehat{\mathbf{x}}_k)- \widehat{\mathbf{x}}_{k+1} \Bigr).
\label{eqn:reduced_lagrangian}
\end{equation}

where $\widehat{\boldsymbol{\lambda}}_{k+1}$ are the adjoint variables.
Taking variations of $\widehat{\mathcal{L}}$ yields the adjoint system.
Variation with respect to $\widehat{\boldsymbol{\lambda}}_{k+1}$ recovers
the dynamics constraint given in Equation~\eqref{eqn:reduced_model}. 
Variation with respect to $\widehat{\mathbf{x}}_k$ for interior points $k \in [k_{corr}, k_{obs}]$ gives the backward adjoint equation:
\begin{equation}
\widehat{\boldsymbol{\lambda}}_k
= \left(\frac{\partial \widehat{\mathcal{M}}}{\partial \widehat{\mathbf{x}}_k}\right)^T \widehat{\boldsymbol{\lambda}}_{k+1}
+ \mathbf{H}^T \mathbf{R}^{-1}(\mathbf{H}\widehat{\mathbf{x}}_k - \mathbf{y}_k),
\label{eqn:adj_eqn}
\end{equation}
with terminal boundary condition $\widehat{\boldsymbol{\lambda}}_{k_{obs}+1} = 0$. Finally, variation with respect to the correction-window initial condition $\widehat{\mathbf{x}}_{k_{corr}}$ yields the gradient:
\begin{equation}
\nabla_{\widehat{\mathbf{x}}_{k_{corr}}} \widehat{\mathcal{J}}
= \widehat{\mathbf{B}}^{-1}(\widehat{\mathbf{x}}_{k_{corr}} - \widehat{\mathbf{x}}^b_{k_{corr}}) + \widehat{\boldsymbol{\lambda}}_{k_{corr}}.
\label{eqn:gradient}
\end{equation}

Once the gradient is computed via the adjoint (Equations~\eqref{eqn:adj_eqn}--\eqref{eqn:gradient}), we pass it to an iterative optimiser that updates the initial state toward the minimum of $\widehat{\mathcal{J}}$. We employ the limited-memory BFGS (L-BFGS) method, a quasi-Newton approach that efficiently exploits gradients for non-linear optimisation.

A key advantage of the OpInf-based ROM is that we explicitly know the polynomial form of $\widehat{\mathcal{M}}$; therefore, the Jacobian $\frac{\partial \widehat{\mathcal{M}}}{\partial \widehat{\mathbf{x}}}$ can be computed analytically without requiring access to the original full-order source code or finite-difference approximations. This analytical Jacobian is then used in Equation~\eqref{eqn:adj_eqn} to propagate sensitivities backward through the correction window.

\subsubsection{Correction Windowing: Correcting when error threshold is exceeded}\label{sec:windowing}

The preceding discussion shows that, even when the reduced adjoint is derived consistently, the backward propagation of mismatch errors can still degrade the gradient as the correction window grows. This is precisely why we do not pursue a single long-horizon assimilation over the full prediction interval. Instead, we adopt a correction-windowed strategy where the assimilation period is divided into short, overlapping intervals, and local 4D-Var corrections are applied only within the active window. By keeping the backward integration horizon short, we can limit the accumulation of adjoint mismatch and preserve gradient quality. In the examples below, this strategy reduces error while retaining the computational advantages of reduced-order variational assimilation.

The correction-windowed assimilation approach introduces several key parameters that must be tuned for a given problem. The observation frequency ($f_{\text{obs}}$) represents the fraction of time steps at which observations are available (e.g., $f_{\text{obs}} = 1/5$ means observations every 5 steps). The trigger threshold ($\tau$) is the relative error tolerance above which a 4D-Var correction window is activated: when forecast error exceeds $\tau$, assimilation is triggered, and this event is also called a trigger event. The correction window length ($N_{\text{window}}$) defines the time span we go back (in time) to make the correction after a measurement becomes available. Specifically, if an observation is available at time step $k_{\text{obs}}$ and we correct the state at time $k_{\text{corr}}$, then $k_{\text{corr}} := k_{\text{obs}} - N_{\text{window}} + 1$. A larger $N_{\text{window}}$ incorporates more temporal information but can increase ROM drift, requiring careful tuning. The choice of these parameters trades off computational cost, accuracy, and ROM predictability. For a given problem, we use a systematic tuning approach via grid search to determine the optimal values for above-mentioned parameters. This is also detailed in Section~\ref{subsubsec:param_opt}.

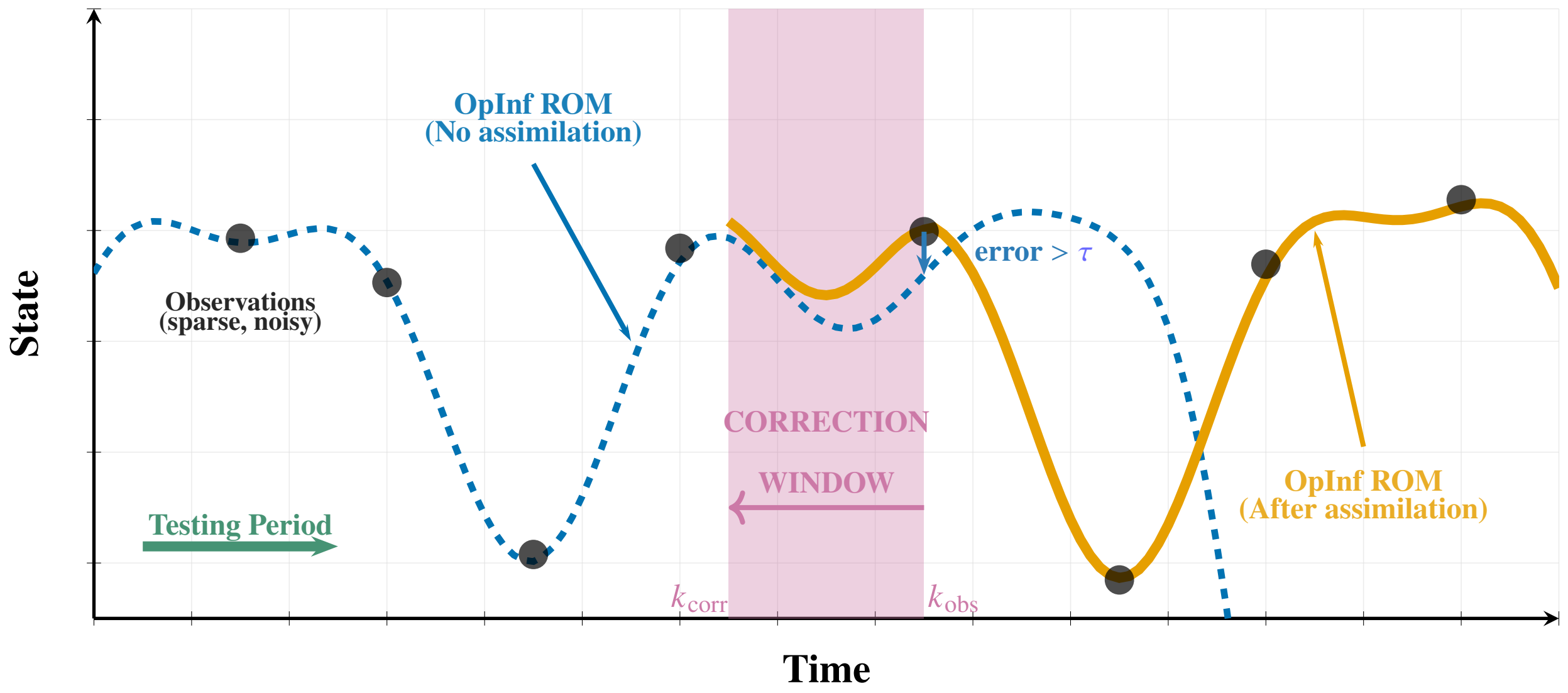
\begin{figure}[H]
    \centering
    \begin{tikzpicture}

\begin{axis}[
width=17cm,
height=8cm,
xmin=0, xmax=150,
ymin=-2.5, ymax=3,
axis on top=false,
axis lines=left,
axis line style={black, line width=1pt},
tick style={black},
xlabel={\textbf{Time}},
ylabel={\textbf{State}},
xticklabels=\empty,
yticklabels=\empty,
xlabel style={font=\large\bfseries},
ylabel style={font=\large\bfseries},
x label style={at={(axis description cs:0.5,-0.04)},anchor=north},
y label style={at={(axis description cs:-0.03,0.5)},anchor=south},
grid=both,
grid style={color=gridcolor, line width=0.3pt, opacity=0.6},
minor grid style={color=gridcolor, line width=0.2pt, opacity=0.4},
clip=true,
clip mode=individual,
]

\fill[corrWindowColor, opacity=0.35] (axis cs:65,-2.5) rectangle (axis cs:85,3.0);

\node[anchor=south, font=\bfseries, text=corrWindowColor]
  at (axis cs:62, -2.55) {$k_{\mathrm{corr}}$};
\node[anchor=south, font=\bfseries, text=corrWindowColor]
  at (axis cs:88, -2.55) {$k_{\mathrm{obs}}$};

\addplot[
  color=romForecastColor,
  dashed,
  line width=2.0pt,
] coordinates {
(0,0.6150) (1,0.7594) (2,0.8761) (3,0.9652) (4,1.0278) (5,1.0658) (6,1.0821)
(7,1.0802) (8,1.0643) (9,1.0384) (10,1.0070) (11,0.9740) (12,0.9432) (13,0.9175)
(14,0.8994) (15,0.8902) (16,0.8906) (17,0.9000) (18,0.9171) (19,0.9398) (20,0.9650)
(21,0.9892) (22,1.0085) (23,1.0186) (24,1.0154) (25,0.9950) (26,0.9539) (27,0.8895)
(28,0.7998) (29,0.6837) (30,0.5414) (31,0.3742) (32,0.1845) (33,-0.0244) (34,-0.2478)
(35,-0.4806) (36,-0.7168) (37,-0.9499) (38,-1.1734) (39,-1.3805) (40,-1.5650)
(41,-1.7211) (42,-1.8438) (43,-1.9290) (44,-1.9739) (45,-1.9868) (46,-1.9374)
(47,-1.8568) (48,-1.7372) (49,-1.5825) (50,-1.3972) (51,-1.1871) (52,-0.9586)
(53,-0.7186) (54,-0.4742) (55,-0.2326) (56,-0.0006) (57,0.2155) (58,0.4101)
(59,0.5787) (60,0.7178) (61,0.8249) (62,0.8991) (63,0.9406) (64,0.9507) (65,0.9321)
(66,0.8881) (67,0.8231) (68,0.7420) (69,0.6501) (70,0.5529) (71,0.4557) (72,0.3635)
(73,0.2810) (74,0.2121) (75,0.1599) (76,0.1265) (77,0.1133) (78,0.1204) (79,0.1474)
(80,0.1926) (81,0.2540) (82,0.3286) (83,0.4132) (84,0.5044) (85,0.5986) (86,0.6924)
(87,0.7826) (88,0.8664) (89,0.9417) (90,1.0069) (91,1.0608) (92,1.1032) (93,1.1342)
(94,1.1545) (95,1.1651) (96,1.1673) (97,1.1624) (98,1.1515) (99,1.1352) (100,1.1141)
(101,1.0875) (102,1.0544) (103,1.0126) (104,0.9593) (105,0.8904) (106,0.8013)
(107,0.6865) (108,0.5402) (109,0.3560) (110,0.1278) (111,-0.1503) (112,-0.4835)
(113,-0.8759) (114,-1.3303) (115,-1.8475) (116,-2.4263) (117,-3.0630)
};

\addplot[
  color=assimColor,
  solid,
  line width=3.0pt,
] coordinates {
(65,1.0821) (66,1.0127) (67,0.9306) (68,0.8405) (69,0.7479) (70,0.6583) (71,0.5767)
(72,0.5082) (73,0.4566) (74,0.4251) (75,0.4157) (76,0.4289) (77,0.4639) (78,0.5187)
(79,0.5895) (80,0.6715) (81,0.7586) (82,0.8438) (83,0.9192) (84,0.9767) (85,1.0076)
(86,1.0273) (87,0.9651) (88,0.8763) (89,0.7599) (90,0.6158) (91,0.4451) (92,0.2501)
(93,0.0342) (94,-0.1983) (95,-0.4420) (96,-0.6911) (97,-0.9390) (98,-1.1790)
(99,-1.4045) (100,-1.6088) (101,-1.7861) (102,-1.9311) (103,-2.0394) (104,-2.1077)
(105,-2.1340) (106,-2.1175) (107,-2.0589) (108,-1.9598) (109,-1.8235) (110,-1.6541)
(111,-1.4566) (112,-1.2369) (113,-1.0012) (114,-0.7563) (115,-0.5086) (116,-0.2645)
(117,-0.0299) (118,0.1900) (119,0.3909) (120,0.5694) (121,0.7231) (122,0.8510)
(123,0.9530) (124,1.0300) (125,1.0841) (126,1.1178) (127,1.1346) (128,1.1383)
(129,1.1329) (130,1.1222) (131,1.1101) (132,1.0998) (133,1.0939) (134,1.0942)
(135,1.1018) (136,1.1167) (137,1.1379) (138,1.1636) (139,1.1909) (140,1.2164)
(141,1.2361) (142,1.2456) (143,1.2403) (144,1.2157) (145,1.1678) (146,1.0928)
(147,0.9881) (148,0.8517) (149,0.6828) (150,0.4821)
};

\addplot[
  only marks,
  mark=*,
  mark size=4pt,
  color=obsColor,
  mark options={fill=obsColor, draw=obsColor, line width=0.8pt},
  opacity=0.7,
] coordinates {
  (15,0.9299)
  (30,0.5304)
  (45,-1.9230)
  (60,0.8396)
  (85,0.9889)
  (105,-2.1527)
  (120,0.6957)
  (140,1.2778)
};



\draw[-{Stealth[length=6pt]}, color=cbOpt, line width=1.5pt]
  (axis cs:85,0.9889) -- (axis cs:85,0.5986);

\node[anchor=west, font=\small\bfseries, text=cbOpt]
  at (axis cs:89, 0.7937) {error $> \textcolor{triggerthreshold}{\tau}$};

\draw[-{Stealth[length=8pt,width=6pt]}, color=trainingtesting, line width=3.0pt]
  (axis cs:5,-1.85) -- (axis cs:25,-1.85);
\node[font=\small\bfseries, text=trainingtesting, anchor=center]
  at (axis cs:15,-1.68) {Testing Period};

\node[
  font=\small\bfseries,
  text=romForecastColor,
  rounded corners=3pt,
  inner sep=3.5pt,
  opacity=0.9,
  anchor=center,
  align=center,
] at (axis cs:45,2.0) {OpInf ROM \\[-10pt] (No assimilation) };

\draw[-{Stealth[length=5pt,width=3pt]}, color=romForecastColor, line width=1.5pt]
  (axis cs:45,1.6) -- (axis cs:55,0.0);

\node[
  font=\small\bfseries,
  text=corrWindowColor,
  anchor=center,
  align=center,
] at (axis cs:75,-1) {CORRECTION\\WINDOW};

\draw[<-, color=corrWindowColor, line width=1.5pt]
   (axis cs:65,-1.5) -- (axis cs:85,-1.5);

\node[
  font=\small\bfseries,
  text=assimColor,
  rounded corners=3pt,
  inner sep=1.5pt,
  opacity=0.85,
  align=center,
  anchor=center,
] at (axis cs:130,-1.40) {OpInf ROM \\[-10pt] (After assimilation)  };

\draw[-{Stealth[length=5pt,width=3pt]}, color=assimColor, line width=1.5pt]
  (axis cs:130,-0.95) -- (axis cs:125,1.0);

\node[
  font=\footnotesize\bfseries,
  text=obsColor,
  rounded corners=3pt,
  inner sep=1.0pt,
  opacity=0.85,
  align=center,
  anchor=center,
] at (axis cs:15,0.25) {Observations\\[-10pt](sparse, noisy)};


\end{axis}
\end{tikzpicture}
    \vspace{-1cm}
    \caption{Illustration of error-triggered correction-windowed 4D-Var strategy during testing period.  The figure is a schematic, not based on a real system. Unassimilated OpInf ROM forecast (\textcolor{romForecastColor}{blue dashed line}), which gradually diverges from the observations. Sparse circles (\textcolor{obsColor}{in black}) mark noisy observations available intermittently. When forecast error exceeds the error threshold $\tau$, a backwards-looking assimilation correction window of length $N_{\text{window}}$ steps is activated (\textcolor{corrWindowColor}{pink-shaded band}). The 4D-Var optimisation propagates information backwards to the start of this correction window and corrects the state prediction (\textcolor{assimColor}{orange solid line }) to better fit the observations. 
    }
    \label{fig:windowed_4dvar_mechanism}
\end{figure}

With these parameters established, the correction-windowing algorithm requires a strategy to determine when corrections are needed. Rather than applying 4D-Var at every time step—which would be computationally redundant when the drift is low—we monitor forecast accuracy and trigger corrections only when forecast error exceeds the $\tau$. This error-triggered approach reduces unnecessary optimisation while maintaining assimilation accuracy during periods of drift. We define the relative forecast error as:
\begin{equation}
	E_{\text{obs}} = \left\|\mathbf{y}_k - \mathbf{H}\widehat{\mathbf{x}}_k\right\|
	\label{eqn:rel_error}
\end{equation}
When $E_{\text{obs}} > \tau$ (user-set trigger threshold), a 4D-Var correction window is triggered to correct the state condition at the start of the correction window. Unlike the projection error discussed later, this error is calculated exclusively at specific observation locations and triggers corrections only when necessary, preventing wasted computation during periods of forecast accuracy. Figure ~\ref{fig:windowed_4dvar_mechanism} and Algorithm ~\ref{algo:windowed_4dvar}  detail the proposed framework pictorially and algorithmically, respectively.

\begin{algorithm}[H]
\caption{Correction-Windowed 4D-Var}
\label{algo:windowed_4dvar}
\begin{algorithmic}[1]
\Require Learned ROM $\widehat{\mathcal{M}}$, reduced observation operator $\mathbf{H} = \mathbf{C}\mathbf{\Phi}$ (from Section~\ref{sec:sensor_placement}), observations $\{\mathbf{y}_{k_{\text{obs}}}\}$, trigger threshold $\tau$, correction length $N_{\text{window}}$, covariances $\widehat{\mathbf{B}},\mathbf{R}$
\State Initialise background trajectory from initial guess $\widehat{\mathbf{x}}_0^b$
\For{each observation time step $k_{\text{obs}}$}
\State Compute $E_{\text{obs}}=\|\mathbf{y}_{k_{\text{obs}}}-\mathbf{H}\widehat{\mathbf{x}}_{k_{\text{obs}}}\|$
\If{$E_{\text{obs}} > \tau$}
\State Set correction-window start $k_{\text{corr}} \leftarrow \max(0,\,k_{\text{obs}}-N_{\text{window}}+1)$ and extract local background $\widehat{\mathbf{x}}_{k_{\text{corr}}}^b$
\State Initialize optimization variable $\widehat{\mathbf{x}}_{k_{\text{corr}}}^b$ and iteration $i \leftarrow 0$
\While{$\|\nabla \mathcal{J}_{\text{window}}(\widehat{\mathbf{x}}_{k_{\text{corr}}}^b)\| > \epsilon$ and $i <$ maximum iteration count}
\State Compute the forward trajectory over $[k_{\mathrm{corr}},k_{\mathrm{obs}}]$
   and evaluate $J_{\mathrm{window}}$ using Equation~\eqref{eqn:4dvar_cost}
\State Solve the adjoint equations over $[k_{\mathrm{corr}},k_{\mathrm{obs}}]$
   and compute $\nabla J_{\mathrm{window}}$ using Equation\eqref{eqn:adj_eqn}---\eqref{eqn:gradient}
\State Update $\widehat{\mathbf{x}}^{\,b}_{k_{\mathrm{corr}}}$ using an
    L-BFGS optimization step and set $i \leftarrow i+1$
\EndWhile

\State Update $\widehat{\mathbf{x}}_{k_{\text{corr}}} \leftarrow \widehat{\mathbf{x}}_{k_{\text{corr}}}^b$ and recompute forecast to $k_{\text{obs}}$
\EndIf
\EndFor

\end{algorithmic}
\end{algorithm}

\subsection{Error Bounds and Stability Mechanisms in Windowed OpInf-4D-Var}
\label{sec:bounds_and_stability}

In extrapolative testing regimes, open-loop OpInf ROM predictions may
progressively depart from the physical trajectory as local reduced-model errors
accumulate. To limit this drift without requiring repeated full-order forward
and adjoint evaluations, we perform 4D-Var data assimilation directly in the
reduced coordinate space using intermittently available observations.

This raises a fundamental theoretical question: \emph{when the reduced-order model optimizes its own initial condition using data-driven, reduced adjoint sensitivities, is that update guaranteed to improve alignment with the true full-order physical dynamics, or does the surrogate gradient introduce spurious directions that corrupt the state?}

In this section, we formulate the error propagation dynamics across a local correction window containing \(N_{\mathrm{window}}\) discrete state levels and ending at time step \(k_{\mathrm{obs}}\). We define $K \triangleq N_{\mathrm{window}}-1$ as the number of model transitions within the window, and use \(k\in\{0,\ldots,K\}\) to index the local window, with \(k=0\) corresponding to \(k_{\mathrm{corr}}\) and \(k=K\) corresponding to \(k_{\mathrm{obs}}\). Thus,

\[
k_{\mathrm{corr}}=k_{\mathrm{obs}}-K.
\]

We derive analytical bounds on the projected 4D-Var gradient that expose
two complementary error-growth mechanisms. Intermittent state correction limits
the horizon over which forward-model defects accumulate, while restricting the
adjoint horizon limits the amplification of sensitivity errors entering the
reduced variational gradient.

\subsubsection{Structural Assumptions}
\label{subsubsec:assumptions}

We first establish the following operational hypotheses governing the dynamics, observation operators, and prior distributions:

\begin{enumerate}
    \item \textbf{Orthonormal Subspace and State Lifting:} The spatial reduction basis $\boldsymbol{\Phi}_r \in \mathbb{R}^{n \times r}$ ($r \ll n$) has orthonormal columns satisfying $\boldsymbol{\Phi}_r^T \boldsymbol{\Phi}_r = \mathbf{I}_r$. The orthogonal projector onto the reduced subspace $\mathcal{V}_r \triangleq \operatorname{range}(\boldsymbol{\Phi}_r)$ is $\mathbf{P} = \boldsymbol{\Phi}_r \boldsymbol{\Phi}_r^T$, and the projector onto its orthogonal complement $\mathcal{V}_r^\perp$ is $\mathbf{P}^\perp = \mathbf{I}_n - \boldsymbol{\Phi}_r \boldsymbol{\Phi}_r^T$.
    
    \item \textbf{Prior Subspace Alignment and Precision Consistency:} The background state satisfies $\mathbf{x}^b \in \operatorname{range}(\boldsymbol{\Phi}_r)$, such that $\mathbf{x}^b = \boldsymbol{\Phi}_r \widehat{\mathbf{x}}^b$ with $\widehat{\mathbf{x}}^b = \boldsymbol{\Phi}_r^T \mathbf{x}^b$. Furthermore, the reduced background precision matrix satisfies $\widehat{\mathbf{B}}^{-1} = \boldsymbol{\Phi}_r^T \mathbf{B}^{-1} \boldsymbol{\Phi}_r$. This eliminates static prior mismatch and isolates dynamic surrogate errors.

    \item \textbf{Consistent Evaluation Point:} Both the full-order gradient $\mathbf{g}_{\mathrm{FOM}}$ and the reduced-order gradient $\mathbf{g}_{\mathrm{ROM}}$ are evaluated at the identical state defined by the current reduced iterate $\widehat{\mathbf{x}}_{k_{\mathrm{corr}}} \in \mathbb{R}^r$. The full-order evaluation point is the lifted state $\mathbf{x}_{k_{\mathrm{corr}}} = \boldsymbol{\Phi}_r \widehat{\mathbf{x}}_{k_{\mathrm{corr}}}$. Consequently, the initial trajectory mismatch at the window start is identically zero: $\rho_{0} \triangleq \|\mathbf{x}_{k_{\mathrm{corr}}} - \boldsymbol{\Phi}_r \widehat{\mathbf{x}}_{k_{\mathrm{corr}}}\|_2 = 0$.

    \item \textbf{Local Domain Invariance and Forward Regularity:} 
Let $\mathcal{D} \subset \mathbb{R}^n$ denote a compact domain containing the physical full-order attractor. Over the local assimilation window $k \in \{0, \dots, K\}$, the full-order trajectory $\{\mathbf{x}_k\}$ and the lifted reduced iterate $\{\boldsymbol{\Phi}_r \widehat{\mathbf{x}}_k\}$ remain within $\mathcal{D}$. On $\mathcal{D}$, the full-order forward operator $\mathcal{M}_k: \mathbb{R}^n \to \mathbb{R}^n$ is locally Lipschitz continuous with constant $C_M \ge 1$:
\begin{equation}
\|\mathcal{M}_k(\mathbf{u}) - \mathcal{M}_k(\mathbf{v})\|_2 \le C_M \|\mathbf{u} - \mathbf{v}\|_2, \quad \forall \mathbf{u}, \mathbf{v} \in \mathcal{D}.
\label{eqn:lipschitz_forward}
\end{equation}

\begin{remark}[Local Regularity vs. Long-Horizon Divergence]
Assumption~4 applies strictly over the short local window duration $K \Delta t$. While unconstrained open-loop OpInf ROM predictions escape $\mathcal{D}$ over extended extrapolative integrations, the present analysis assumes that both trajectories remain within \(\mathcal D\) over each local correction window.
\end{remark}

    \item \textbf{Tangent Linear Model (TLM) Smoothness:} The full-order Jacobian mapping $\mathbf{x} \mapsto \mathbf{M}_k(\mathbf{x}) \triangleq \frac{\partial \mathcal{M}_k}{\partial \mathbf{x}}(\mathbf{x})$ is locally Lipschitz continuous on $\mathcal{D}$ with constant $L_M$:
    \begin{equation}
        \|\mathbf{M}_k(\mathbf{u}) - \mathbf{M}_k(\mathbf{v})\|_2 \le L_M \|\mathbf{u} - \mathbf{v}\|_2, \quad \forall \mathbf{u}, \mathbf{v} \in \mathcal{D}. \label{eqn:lipschitz_tlm}
    \end{equation}

    \item \textbf{Observation Operator Bound:} The full-order observation operators are linear mappings $\mathcal{H}_k(\mathbf{x}) = \mathbf{C} \mathbf{x}$ with covariance $\mathbf{R} \succ 0$. The observation misfit precision factor satisfies:
    \begin{equation}
        \sigma_R \triangleq \|\mathbf{R}^{-1}\|_2 \|\mathbf{C}\|_2^2 < \infty. \label{eqn:sigma_R_def}
    \end{equation}
\end{enumerate}

\subsubsection{The Optimization Target: Subspace Decomposition and Adjoint Alignment}
\label{subsubsec:subspace_decomp}

Let $\mathcal{J}: \mathbb{R}^n \to \mathbb{R}$ denote the full-order 4D-Var objective function defined over the local window:
\begin{equation}
    \mathcal{J}(\mathbf{x}_{k_{\mathrm{corr}}}) = \frac{1}{2} \|\mathbf{x}_{k_{\mathrm{corr}}} - \mathbf{x}_{k_{\mathrm{corr}}}^b\|_{\mathbf{B}^{-1}}^2 + \frac{1}{2} \sum_{k=k_{\mathrm{corr}}}^{k_{\mathrm{obs}}} \|\mathbf{y}_k - \mathbf{C} \mathbf{x}_k\|_{\mathbf{R}^{-1}}^2, \label{eqn:fom_cost_def}
\end{equation}
subject to $\mathbf{x}_{k+1} = \mathcal{M}_k(\mathbf{x}_k)$. The reduced optimization operates entirely in $\mathbb{R}^r$ under the surrogate cost functional $\widehat{\mathcal{J}}(\widehat{\mathbf{x}}_{k_{\mathrm{corr}}})$ given in Equation~\eqref{eqn:4dvar_cost}.

Because the reduced optimizer executes search updates strictly within the span of $\boldsymbol{\Phi}_r$, we benchmark the reduced gradient $\mathbf{g}_{\mathrm{ROM}} \triangleq \nabla_{\widehat{\mathbf{x}}_{k_{\mathrm{corr}}}} \widehat{\mathcal{J}}(\widehat{\mathbf{x}}_{k_{\mathrm{corr}}}) \in \mathbb{R}^r$ against the full-order gradient $\mathbf{g}_{\mathrm{FOM}} \triangleq \nabla_{\mathbf{x}_{k_{\mathrm{corr}}}} \mathcal{J}(\boldsymbol{\Phi}_r \widehat{\mathbf{x}}_{k_{\mathrm{corr}}}) \in \mathbb{R}^n$ evaluated at the lifted state.

\begin{theorem}[Orthogonal Decomposition of Gradient Error]
\label{thm:decomp}
Under Assumptions~1 and 3, the Euclidean error between the full-order gradient and the lifted reduced gradient decomposes orthogonally as:
\begin{equation}
    \|\mathbf{g}_{\mathrm{FOM}} - \boldsymbol{\Phi}_r \mathbf{g}_{\mathrm{ROM}}\|_2^2 = \underbrace{\|\boldsymbol{\Phi}_r^T \mathbf{g}_{\mathrm{FOM}} - \mathbf{g}_{\mathrm{ROM}}\|_2^2}_{\text{Projected Gradient Error}} + \underbrace{\|(\mathbf{I}_n - \boldsymbol{\Phi}_r \boldsymbol{\Phi}_r^T) \mathbf{g}_{\mathrm{FOM}}\|_2^2}_{\text{Out-of-Subspace Gradient Component}}. \label{eqn:pythagoras}
\end{equation}
\end{theorem}

\begin{remark}[Subspace Expressiveness vs. Optimization Fidelity]
The component $\|(\mathbf{I}_n - \boldsymbol{\Phi}_r \boldsymbol{\Phi}_r^T) \mathbf{g}_{\mathrm{FOM}}\|_2$ represents the out-of-subspace gradient sensitivity that cannot be represented by the POD basis $\boldsymbol{\Phi}_r$. However, any update generated by the reduced optimizer takes the form $\Delta \mathbf{x} = -\alpha \boldsymbol{\Phi}_r \mathbf{g}_{\mathrm{ROM}} \in \operatorname{range}(\boldsymbol{\Phi}_r)$. Because $\Delta \mathbf{x}^T (\mathbf{I}_n - \boldsymbol{\Phi}_r \boldsymbol{\Phi}_r^T) \mathbf{g}_{\mathrm{FOM}} \equiv 0$, the out-of-subspace gradient component is completely decoupled from the update step. Therefore, physical optimization fidelity is governed strictly by the in-subspace projected gradient error $\|\boldsymbol{\Phi}_r^T \mathbf{g}_{\mathrm{FOM}} - \mathbf{g}_{\mathrm{ROM}}\|_2$.
\end{remark}

Using the adjoint gradient representations in Equations~\eqref{eqn:adj_eqn}--\eqref{eqn:gradient}, the projected gradient discrepancy evaluates to:
\begin{align}
    \boldsymbol{\Phi}_r^T \mathbf{g}_{\mathrm{FOM}} - \mathbf{g}_{\mathrm{ROM}} 
    &= \boldsymbol{\Phi}_r^T \left( \mathbf{B}^{-1}(\boldsymbol{\Phi}_r \widehat{\mathbf{x}}_{k_{\mathrm{corr}}} - \mathbf{x}_{k_{\mathrm{corr}}}^b) + \boldsymbol{\lambda}_{k_{\mathrm{corr}}} \right) - \left( \widehat{\mathbf{B}}^{-1}(\widehat{\mathbf{x}}_{k_{\mathrm{corr}}} - \widehat{\mathbf{x}}_{k_{\mathrm{corr}}}^b) + \widehat{\boldsymbol{\lambda}}_{k_{\mathrm{corr}}} \right) \nonumber \\
    &= \underbrace{\left( \boldsymbol{\Phi}_r^T \mathbf{B}^{-1} \boldsymbol{\Phi}_r - \widehat{\mathbf{B}}^{-1} \right)(\widehat{\mathbf{x}}_{k_{\mathrm{corr}}} - \widehat{\mathbf{x}}_{k_{\mathrm{corr}}}^b) - \boldsymbol{\Phi}_r^T \mathbf{B}^{-1} (\mathbf{I}_n - \boldsymbol{\Phi}_r \boldsymbol{\Phi}_r^T) \mathbf{x}_{k_{\mathrm{corr}}}^b}_{\mathbf{E}_{\mathrm{prior}}} + \left( \boldsymbol{\Phi}_r^T \boldsymbol{\lambda}_{k_{\mathrm{corr}}} - \widehat{\boldsymbol{\lambda}}_{k_{\mathrm{corr}}} \right).
\end{align}
Under Assumption~2, the prior error term vanishes identically ($\mathbf{E}_{\mathrm{prior}} = \mathbf{0}$). Shifting the window indices to $k \in \{0, \dots, K\}$ for convenience (where $k=0$ corresponds to $k_{\mathrm{corr}}$ and $k=K$ corresponds to $k_{\mathrm{obs}}$), the projected gradient discrepancy reduces to the terminal adjoint error at the start of the window:
\begin{equation}
    \|\boldsymbol{\Phi}_r^T \mathbf{g}_{\mathrm{FOM}} - \mathbf{g}_{\mathrm{ROM}}\|_2 = \|\boldsymbol{\Phi}_r^T \boldsymbol{\lambda}_0 - \widehat{\boldsymbol{\lambda}}_0\|_2 \triangleq \varepsilon_0. \label{eqn:grad_to_adj}
\end{equation}

\subsubsection{Forward Trajectory Drift via Discrete Grönwall Inequality}
\label{subsubsec:forward_drift}

To understand how sensitivity errors accumulate in Equation~\eqref{eqn:grad_to_adj}, we first bound how far the reduced trajectory drifts forward in time away from the true physical dynamics. 
Let $\rho_k \triangleq \|\mathbf{x}_k - \boldsymbol{\Phi}_r \widehat{\mathbf{x}}_k\|_2$ denote the forward state error at step $k$. At each discrete step, the learned OpInf surrogate incurs a local one-step forward modeling defect:
\begin{equation}
    \delta_{\mathrm{OpInf}}(k) \triangleq \|\mathcal{M}_k(\boldsymbol{\Phi}_r \widehat{\mathbf{x}}_k) - \boldsymbol{\Phi}_r \widehat{\mathcal{M}}_k(\widehat{\mathbf{x}}_k)\|_2. \label{eqn:opinf_defect_def}
\end{equation}

\begin{lemma}[Forward State Error Propagation]
\label{lem:forward_gronwall}
Under Assumptions~3 and 4, the forward trajectory drift across the window satisfies:
\begin{equation}
    \rho_k \le \sum_{j=0}^{k-1} C_M^{k-1-j} \delta_{\mathrm{OpInf}}(j). \label{eqn:gronwall_sum}
\end{equation}
If additionally the one-step defect is uniformly bounded by $\delta_{\mathrm{OpInf}}(j) \le \bar{\delta}$ for all $j \in \{0, \dots, K-1\}$, then:
\begin{equation}
    \rho_k \le \begin{cases} 
        \bar{\delta} \, \dfrac{C_M^k - 1}{C_M - 1}, & C_M \neq 1, \\[8pt] 
        k \, \bar{\delta}, & C_M = 1. 
    \end{cases} \label{eqn:gronwall_geom}
\end{equation}
\end{lemma}

\begin{proof}
At time step $j$, decompose the state discrepancy by adding and subtracting $\mathcal{M}_j(\boldsymbol{\Phi}_r \widehat{\mathbf{x}}_j)$:
\begin{equation*}
    \mathbf{x}_{j+1} - \boldsymbol{\Phi}_r \widehat{\mathbf{x}}_{j+1} = \left( \mathcal{M}_j(\mathbf{x}_j) - \mathcal{M}_j(\boldsymbol{\Phi}_r \widehat{\mathbf{x}}_j) \right) + \left( \mathcal{M}_j(\boldsymbol{\Phi}_r \widehat{\mathbf{x}}_j) - \boldsymbol{\Phi}_r \widehat{\mathcal{M}}_j(\widehat{\mathbf{x}}_j) \right).
\end{equation*}
Taking Euclidean norms and invoking Lipschitz continuity (Assumption~4) along with Equation~\eqref{eqn:opinf_defect_def}:
\begin{equation}
    \rho_{j+1} \le C_M \rho_j + \delta_{\mathrm{OpInf}}(j). \label{eqn:gronwall_step}
\end{equation}
Unrolling this linear recurrence from the initial condition $\rho_0 = 0$ (Assumption~3) yields Equation~\eqref{eqn:gronwall_sum}. Under a uniform bound $\bar{\delta}$, factoring out $\bar{\delta}$ yields a finite geometric series $\sum_{m=0}^{k-1} C_M^m$, establishing Equation~\eqref{eqn:gronwall_geom}.
\end{proof}

Lemma~\ref{lem:forward_gronwall} reveals the mechanism of open-loop extrapolation failure: in extrapolative regimes, the local modeling defect $\bar{\delta}$ is strictly non-zero. When $C_M > 1$, these local defects compound exponentially as $\sim C_M^k \bar{\delta}$, causing the ROM forecast to rapidly diverge unless periodically re-anchored.

\subsubsection{Backward Adjoint Recurrence and TLM Mismatch Decomposition}
\label{subsubsec:backward_adj}

Having quantified forward drift $\rho_k$, we now trace how it corrupts backward adjoint sensitivities. The full- and reduced-order adjoint variables evolve backward according to:
\begin{align}
    \boldsymbol{\lambda}_k &= \mathbf{M}_k(\mathbf{x}_k)^T \boldsymbol{\lambda}_{k+1} + \mathbf{d}_k, \quad \boldsymbol{\lambda}_K = \mathbf{d}_K, \label{eqn:adj_fom_def} \\
    \widehat{\boldsymbol{\lambda}}_k &= \widehat{\mathbf{M}}_k(\widehat{\mathbf{x}}_k)^T \widehat{\boldsymbol{\lambda}}_{k+1} + \widehat{\mathbf{d}}_k, \quad \widehat{\boldsymbol{\lambda}}_K = \widehat{\mathbf{d}}_K, \label{eqn:adj_rom_def}
\end{align}
where $\mathbf{d}_k \triangleq \mathbf{C}^T \mathbf{R}^{-1}(\mathbf{C}\mathbf{x}_k - \mathbf{y}_k)$ and $\widehat{\mathbf{d}}_k \triangleq \mathbf{H}^T \mathbf{R}^{-1}(\mathbf{H}\widehat{\mathbf{x}}_k - \mathbf{y}_k)$ represent the observation forcing terms, with $\mathbf{H} = \mathbf{C}\boldsymbol{\Phi}_r$.

\begin{lemma}[One-Step Projected Adjoint Error Recurrence]
\label{lem:adj_recurrence}
Let $\varepsilon_k \triangleq \|\boldsymbol{\Phi}_r^T \boldsymbol{\lambda}_k - \widehat{\boldsymbol{\lambda}}_k\|_2$. For each $k \in \{0, \dots, K-1\}$:
\begin{equation}
    \varepsilon_k \le \|\widehat{\mathbf{M}}_k(\widehat{\mathbf{x}}_k)\|_2 \varepsilon_{k+1} + \|\mathbf{M}_k(\mathbf{x}_k)\boldsymbol{\Phi}_r - \boldsymbol{\Phi}_r \widehat{\mathbf{M}}_k(\widehat{\mathbf{x}}_k)\|_2 \|\boldsymbol{\lambda}_{k+1}\|_2 + \|\boldsymbol{\Phi}_r^T \mathbf{d}_k - \widehat{\mathbf{d}}_k\|_2. \label{eqn:adj_step_bound}
\end{equation}
\end{lemma}

\begin{proof}
Multiplying Equation~\eqref{eqn:adj_fom_def} by $\boldsymbol{\Phi}_r^T$ and subtracting Equation~\eqref{eqn:adj_rom_def}:
\begin{equation*}
    \boldsymbol{\Phi}_r^T \boldsymbol{\lambda}_k - \widehat{\boldsymbol{\lambda}}_k = \left( \boldsymbol{\Phi}_r^T \mathbf{M}_k(\mathbf{x}_k)^T \boldsymbol{\lambda}_{k+1} - \widehat{\mathbf{M}}_k(\widehat{\mathbf{x}}_k)^T \widehat{\boldsymbol{\lambda}}_{k+1} \right) + (\boldsymbol{\Phi}_r^T \mathbf{d}_k - \widehat{\mathbf{d}}_k).
\end{equation*}
Inserting the hybrid cross-term $\widehat{\mathbf{M}}_k(\widehat{\mathbf{x}}_k)^T \boldsymbol{\Phi}_r^T \boldsymbol{\lambda}_{k+1}$ decouples operator mismatch from adjoint error:
\begin{align*}
    \boldsymbol{\Phi}_r^T \mathbf{M}_k^T \boldsymbol{\lambda}_{k+1} - \widehat{\mathbf{M}}_k^T \widehat{\boldsymbol{\lambda}}_{k+1} 
    &= (\boldsymbol{\Phi}_r^T \mathbf{M}_k^T - \widehat{\mathbf{M}}_k^T \boldsymbol{\Phi}_r^T)\boldsymbol{\lambda}_{k+1} + \widehat{\mathbf{M}}_k^T (\boldsymbol{\Phi}_r^T \boldsymbol{\lambda}_{k+1} - \widehat{\boldsymbol{\lambda}}_{k+1}).
\end{align*}
Taking norms, applying the triangle inequality, and utilizing $\|\mathbf{A}^T\|_2 = \|\mathbf{A}\|_2$ gives:
\begin{align*}
    \varepsilon_k &\le \|\widehat{\mathbf{M}}_k\|_2 \|\boldsymbol{\Phi}_r^T \boldsymbol{\lambda}_{k+1} - \widehat{\boldsymbol{\lambda}}_{k+1}\|_2 + \|\boldsymbol{\Phi}_r^T \mathbf{M}_k^T - \widehat{\mathbf{M}}_k^T \boldsymbol{\Phi}_r^T\|_2 \|\boldsymbol{\lambda}_{k+1}\|_2 + \|\boldsymbol{\Phi}_r^T \mathbf{d}_k - \widehat{\mathbf{d}}_k\|_2 \\
    &\leq \|\widehat{\mathbf{M}}_k\|_2 \varepsilon_{k+1} + \|\mathbf{M}_k \boldsymbol{\Phi}_r - \boldsymbol{\Phi}_r \widehat{\mathbf{M}}_k\|_2 \|\boldsymbol{\lambda}_{k+1}\|_2 + \|\boldsymbol{\Phi}_r^T \mathbf{d}_k - \widehat{\mathbf{d}}_k\|_2,
\end{align*}
which establishes Equation~\eqref{eqn:adj_step_bound}.
\end{proof}

We now unpack the two driving perturbations in Equation~\eqref{eqn:adj_step_bound}: the Tangent Linear Model mismatch and the observation forcing discrepancy.

\begin{proposition}[Linear-Quadratic OpInf Jacobian and Adjoint Amplification]
\label{prop:opinf_jacobian_bound}
For the linear-quadratic OpInf surrogate defined in Equation~\eqref{eq:opinf_poly}, the Tangent Linear Model (TLM) Jacobian $\widehat{\mathbf{M}}(\widehat{\mathbf{x}}) \triangleq \frac{\partial \widehat{\mathcal{M}}}{\partial \widehat{\mathbf{x}}}(\widehat{\mathbf{x}}) \in \mathbb{R}^{r \times r}$ evaluates analytically to:
\begin{equation}
\widehat{\mathbf{M}}(\widehat{\mathbf{x}}) = \widehat{\mathbf{A}} + 2\widehat{\mathbf{F}}(\widehat{\mathbf{x}} \otimes \mathbf{I}_r).
\label{eqn:opinf_tlm_formula}
\end{equation}
Over any bounded state trajectory $\|\widehat{\mathbf{x}}_k\|_2 \le R$, the spectral norm of the reduced adjoint operator is directly bounded by the learned OpInf matrix coefficients:
\begin{equation}
\|\widehat{\mathbf{M}}(\widehat{\mathbf{x}}_k)\|_2 \le \mu_k \triangleq \|\widehat{\mathbf{A}}\|_2 + 2\|\widehat{\mathbf{F}}\|_2 \|\widehat{\mathbf{x}}_k\|_2 \le \|\widehat{\mathbf{A}}\|_2 + 2\|\widehat{\mathbf{F}}\|_2 R \triangleq \bar{\mu}.
\label{eqn:mu_derived_bound}
\end{equation}
\end{proposition}

\begin{remark}[Equivalence to Compact Quadratic Monomial Implementations]
\label{rem:compact_monomials}
In numerical implementations of Operator Inference, the redundant symmetric cross-terms in $\widehat{\mathbf{x}} \otimes \widehat{\mathbf{x}} \in \mathbb{R}^{r^2}$ are typically omitted to improve regression conditioning, yielding a compact quadratic representation $\widehat{\mathbf{F}}_{\mathrm{compact}} \widehat{\mathbf{x}}^{\underline{\otimes} 2}$ with $\widehat{\mathbf{F}}_{\mathrm{compact}} \in \mathbb{R}^{r \times \frac{r(r+1)}{2}}$ and $\widehat{\mathbf{x}}^{\underline{\otimes} 2} \triangleq [\widehat{x}_1^2, \widehat{x}_1 \widehat{x}_2, \dots, \widehat{x}_r^2]^T$. 

Because polynomial evaluation along the trajectory diagonal is uniquely governed by the symmetric component of the underlying bilinear form, there exists a unique symmetric matricization $\widehat{\mathbf{F}} \in \mathbb{R}^{r \times r^2}$ satisfying $\widehat{\mathbf{F}}(\widehat{\mathbf{x}} \otimes \widehat{\mathbf{x}}) \equiv \widehat{\mathbf{F}}_{\mathrm{compact}} \widehat{\mathbf{x}}^{\underline{\otimes} 2}$ for all $\widehat{\mathbf{x}} \in \mathbb{R}^r$. The resulting Tangent Linear Model Jacobian $\widehat{\mathbf{M}}(\widehat{\mathbf{x}}) = \frac{\partial \widehat{\mathcal{M}}}{\partial \widehat{\mathbf{x}}}(\widehat{\mathbf{x}}) \in \mathbb{R}^{r \times r}$ and its action on perturbations are invariant to this matricization choice. We adopt the symmetric Kronecker form throughout this derivation without loss of generality, as it allows clean application of the Kronecker spectral identity $\|\widehat{\mathbf{x}} \otimes \mathbf{I}_r\|_2 = \|\widehat{\mathbf{x}}\|_2$ while preserving exact correspondence to compact numerical implementations.
\end{remark}

\begin{proposition}[Decomposition of Tangent Linear Model Mismatch]
\label{prop:tlm_split}
Under Assumptions~1, 4, and 5, the operator mismatch decomposes into three structural components:
\begin{equation}
    \|\mathbf{M}_k(\mathbf{x}_k)\boldsymbol{\Phi}_r - \boldsymbol{\Phi}_r \widehat{\mathbf{M}}_k(\widehat{\mathbf{x}}_k)\|_2 \le \vphantom{\Delta_{\mathrm{TLM}, k}} \eta_{\mathrm{sub}, k} +  L_M \rho_k + \Delta_{\mathrm{TLM}, k}, \label{eqn:tlm_decomp}
\end{equation}
where:
\begin{itemize}
    \item $\eta_{\mathrm{sub}, k} \triangleq \|(\mathbf{I}_n - \boldsymbol{\Phi}_r \boldsymbol{\Phi}_r^T)\mathbf{M}_k(\mathbf{x}_k)\boldsymbol{\Phi}_r\|_2$ is the out-of-subspace TLM projection residual,
    \item $L_M \rho_k$ captures the Jacobian distortion caused by forward trajectory drift $\rho_k = \|\mathbf{x}_k - \boldsymbol{\Phi}_r \widehat{\mathbf{x}}_k\|_2$, and
    \item $\Delta_{\mathrm{TLM}, k} \triangleq \|\widehat{\mathbf{M}}_k(\widehat{\mathbf{x}}_k) - \boldsymbol{\Phi}_r^T \mathbf{M}_k(\boldsymbol{\Phi}_r \widehat{\mathbf{x}}_k)\boldsymbol{\Phi}_r\|_2$ is the intrinsic OpInf Jacobian defect.
\end{itemize}

Under the linear-quadratic model, the intrinsic OpInf Jacobian defect $\Delta_{\mathrm{TLM}, k}$ is governed directly by the coefficient errors of the learned operators:
\begin{equation}
\Delta_{\mathrm{TLM}, k} \le \|\widehat{\mathbf{A}} - \boldsymbol{\Phi}_r^T \mathbf{A} \boldsymbol{\Phi}_r\|_2 + 2\|\widehat{\mathbf{F}} - \boldsymbol{\Phi}_r^T \mathbf{F}(\boldsymbol{\Phi}_r \otimes \boldsymbol{\Phi}_r)\|_2 \|\widehat{\mathbf{x}}_k\|_2 + \|\mathbf{r}_{\mathrm{TLM}, k}\|_2,
\label{eqn:delta_tlm_explicit}
\end{equation}
where $\mathbf{A}$ and $\mathbf{F}$ represent the full-order linear and quadratic operators, and $\mathbf{r}_{\mathrm{TLM}, k}$ denotes any residual non-polynomial Jacobian components.
\end{proposition}

\begin{proof}
Decompose $\mathbf{M}_k(\mathbf{x}_k)\boldsymbol{\Phi}_r$ using $\mathbf{I}_n = (\mathbf{I}_n - \boldsymbol{\Phi}_r \boldsymbol{\Phi}_r^T) + \boldsymbol{\Phi}_r \boldsymbol{\Phi}_r^T$:
\begin{equation*}
    \mathbf{M}_k(\mathbf{x}_k)\boldsymbol{\Phi}_r - \boldsymbol{\Phi}_r \widehat{\mathbf{M}}_k(\widehat{\mathbf{x}}_k) = (\mathbf{I}_n - \boldsymbol{\Phi}_r \boldsymbol{\Phi}_r^T)\mathbf{M}_k(\mathbf{x}_k)\boldsymbol{\Phi}_r + \boldsymbol{\Phi}_r \left( \boldsymbol{\Phi}_r^T \mathbf{M}_k(\mathbf{x}_k)\boldsymbol{\Phi}_r - \widehat{\mathbf{M}}_k(\widehat{\mathbf{x}}_k) \right).
\end{equation*}
Since $\boldsymbol{\Phi}_r$ is an isometry ($\|\boldsymbol{\Phi}_r \mathbf{A}\|_2 = \|\mathbf{A}\|_2$):
\begin{equation*}
    \|\mathbf{M}_k(\mathbf{x}_k)\boldsymbol{\Phi}_r - \boldsymbol{\Phi}_r \widehat{\mathbf{M}}_k(\widehat{\mathbf{x}}_k)\|_2 \le \eta_{\mathrm{sub}, k} + \|\boldsymbol{\Phi}_r^T \mathbf{M}_k(\mathbf{x}_k)\boldsymbol{\Phi}_r - \widehat{\mathbf{M}}_k(\widehat{\mathbf{x}}_k)\|_2.
\end{equation*}
Adding and subtracting the Galerkin projection evaluated at the lifted ROM state, $\boldsymbol{\Phi}_r^T \mathbf{M}_k(\boldsymbol{\Phi}_r \widehat{\mathbf{x}}_k)\boldsymbol{\Phi}_r$:
\begin{align*}
    \|\boldsymbol{\Phi}_r^T \mathbf{M}_k(\mathbf{x}_k)\boldsymbol{\Phi}_r - \widehat{\mathbf{M}}_k(\widehat{\mathbf{x}}_k)\|_2 
    &\le \|\boldsymbol{\Phi}_r^T \left( \mathbf{M}_k(\mathbf{x}_k) - \mathbf{M}_k(\boldsymbol{\Phi}_r \widehat{\mathbf{x}}_k) \right) \boldsymbol{\Phi}_r\|_2 + \Delta_{\mathrm{TLM}, k} \\
    &\le \|\mathbf{M}_k(\mathbf{x}_k) - \mathbf{M}_k(\boldsymbol{\Phi}_r \widehat{\mathbf{x}}_k)\|_2 + \Delta_{\mathrm{TLM}, k} \\
    &\le L_M \|\mathbf{x}_k - \boldsymbol{\Phi}_r \widehat{\mathbf{x}}_k\|_2 + \Delta_{\mathrm{TLM}, k} = L_M \rho_k + \Delta_{\mathrm{TLM}, k},
\end{align*}
where the last inequality follows from Assumption~5.
\end{proof}

\begin{remark}[Derivative Error vs. Forward Accuracy]
Equations~\eqref{eq:opinf_poly} and~\eqref{eqn:delta_tlm_explicit} highlight an essential structural property: quadratic coefficient regression errors enter the forward trajectory defect at order $\mathcal{O}(\|\widehat{\mathbf{x}}\|_2^2)$, whereas they enter the adjoint sensitivity mismatch $\Delta_{\mathrm{TLM}, k}$ linearly at order $\mathcal{O}(\|\widehat{\mathbf{x}}\|_2)$ with an explicit prefactor of 2.
\end{remark}

\begin{lemma}[Observation Misfit Error Bound]
\label{lem:obs_bound}
Under Assumptions~1 and 6, the observation forcing discrepancy satisfies:
\begin{equation}
    \|\boldsymbol{\Phi}_r^T \mathbf{d}_k - \widehat{\mathbf{d}}_k\|_2 \le \sigma_R \rho_k, \quad \forall k \in \{0, \dots, K\}. \label{eqn:obs_error_bound}
\end{equation}
\end{lemma}

\begin{proof}
Substituting the observation forcing terms and noting that $\mathbf{H}^T = \boldsymbol{\Phi}_r^T \mathbf{C}^T$:
\begin{align*}
    \boldsymbol{\Phi}_r^T \mathbf{d}_k - \widehat{\mathbf{d}}_k 
    &= \boldsymbol{\Phi}_r^T \mathbf{C}^T \mathbf{R}^{-1}(\mathbf{C}\mathbf{x}_k - \mathbf{y}_k) - \boldsymbol{\Phi}_r^T \mathbf{C}^T \mathbf{R}^{-1}(\mathbf{C}\boldsymbol{\Phi}_r \widehat{\mathbf{x}}_k - \mathbf{y}_k) \\
    &= \boldsymbol{\Phi}_r^T \mathbf{C}^T \mathbf{R}^{-1}\mathbf{C}(\mathbf{x}_k - \boldsymbol{\Phi}_r \widehat{\mathbf{x}}_k).
\end{align*}
Taking norms with $\|\boldsymbol{\Phi}_r\|_2 = 1$ yields $\|\boldsymbol{\Phi}_r^T \mathbf{d}_k - \widehat{\mathbf{d}}_k\|_2 \le \|\mathbf{R}^{-1}\|_2 \|\mathbf{C}\|_2^2 \|\mathbf{x}_k - \boldsymbol{\Phi}_r \widehat{\mathbf{x}}_k\|_2 \le \sigma_R \rho_k$.
\end{proof}

\begin{remark}[Differentiating Forward Defect from TLM Mismatch]
A critical distinction in our analysis is separating the forward OpInf defect $\delta_{\mathrm{OpInf}}(k)$ from the TLM mismatch $\Delta_{\mathrm{TLM}, k}$. The OpInf regression controls the forward-model fit represented by \(\delta_{\mathrm{OpInf}}(k)\), but does not directly control errors in the learned Jacobian. As shown in Proposition 2, \(\Delta_{\mathrm{TLM},k}\) therefore enters as an independent sensitivity perturbation.
\end{remark}

\subsubsection{Master Bound on Projected Gradient Error and Certified Descent}
\label{subsubsec:master_bound}

We now assemble the forward state drift and backward adjoint error recurrence into a unified projected gradient error bound. Recall that by Proposition~\ref{prop:opinf_jacobian_bound}, the spectral norm of the reduced adjoint operator is directly governed by the learned linear-quadratic OpInf coefficients:
\begin{equation}
\bar{\mu} \triangleq \|\widehat{\mathbf{A}}\|_2 + 2\|\widehat{\mathbf{F}}\|_2 R.
\label{eqn:bar_mu_def}
\end{equation}

\begin{theorem}[Comprehensive Bound on Reduced 4D-Var Gradient Error]
\label{thm:main_bound}
Suppose Assumptions~1--6 hold. Let $\Lambda \triangleq \max_{1 \le k \le K} \|\boldsymbol{\lambda}_k\|_2$ bound the full-order adjoint states, and let $\bar{\mu}$ defined in Equation~\eqref{eqn:bar_mu_def} bound the spectral norm of the reduced Tangent Linear Model over the bounded trajectory $\|\widehat{\mathbf{x}}_k\|_2 \le R$. Then the projected gradient error evaluated at the window start satisfies:
\begin{equation}
\|\boldsymbol{\Phi}_r^T \mathbf{g}_{\mathrm{FOM}} - \mathbf{g}_{\mathrm{ROM}}\|_2 \le \sum_{k=0}^{K-1} \bar{\mu}^k \Big( \Lambda (\eta_{\mathrm{sub}, k} + \Delta_{\mathrm{TLM}, k}) + (\Lambda L_M + \sigma_R)\rho_k \Big) + \bar{\mu}^K \sigma_R \rho_K \triangleq \mathcal{E}_{\mathrm{bound}},
\label{eqn:master_bound}
\end{equation}
where $\rho_k$ is the discrete Grönwall accumulation from Lemma~\ref{lem:forward_gronwall}.

Furthermore, if uniform bounds hold over the correction window ($\delta_{\mathrm{OpInf}}(k) \le \bar{\delta}$, $\eta_{\mathrm{sub}, k} \le \bar{\eta}$, $\Delta_{\mathrm{TLM}, k} \le \bar{\Delta}$), then for $\bar{\mu} > 1$ and $C_M > 1$:
\begin{align}
\mathcal{E}_{\mathrm{bound}} &\le \Lambda(\bar{\eta} + \bar{\Delta})\left(\frac{\bar{\mu}^K - 1}{\bar{\mu} - 1}\right) + (\Lambda L_M + \sigma_R)\frac{\bar{\delta}}{C_M - 1} \left[ \frac{(\bar{\mu} C_M)^K - 1}{\bar{\mu} C_M - 1} - \frac{\bar{\mu}^K - 1}{\bar{\mu} - 1} \right] \nonumber \\
&\quad + \sigma_R \bar{\delta}\bar{\mu}^K \left(\frac{C_M^K - 1}{C_M - 1}\right).
\label{eqn:uniform_bound}
\end{align}
\end{theorem}

\begin{proof}
Substituting the TLM decomposition (Proposition~\ref{prop:tlm_split}) and the observation misfit bound (Lemma~\ref{lem:obs_bound}) into the one-step adjoint difference recurrence (Lemma~\ref{lem:adj_recurrence}), we have for each $k \in \{0, \dots, K-1\}$:
\begin{equation*}
\varepsilon_k \le \bar{\mu} \varepsilon_{k+1} + \Gamma_k, \quad \text{where} \quad \Gamma_k \triangleq \Lambda(\eta_{\mathrm{sub}, k} + \Delta_{\mathrm{TLM}, k}) + (\Lambda L_M + \sigma_R)\rho_k.
\end{equation*}
At the terminal step $K$, the boundary conditions dictate $\boldsymbol{\lambda}_K = \mathbf{d}_K$ and $\widehat{\boldsymbol{\lambda}}_K = \widehat{\mathbf{d}}_K$. By Lemma~\ref{lem:obs_bound}, this implies $\varepsilon_K \le \sigma_R \rho_K$.
Because $\bar{\mu} \ge 0$ and $\Gamma_k \ge 0$, unrolling the scalar recurrence backward from $k = 0$ to $k = K$ preserves the inequality order:
\begin{align*}
\varepsilon_0 &\le \bar{\mu} \varepsilon_1 + \Gamma_0 \\
&\le \bar{\mu}^2 \varepsilon_2 + \bar{\mu} \Gamma_1 + \Gamma_0 \\
&\le \sum_{k=0}^{K-1} \bar{\mu}^k \Gamma_k + \bar{\mu}^K \varepsilon_K \\
&\le \sum_{k=0}^{K-1} \bar{\mu}^k \Big( \Lambda(\eta_{\mathrm{sub}, k} + \Delta_{\mathrm{TLM}, k}) + (\Lambda L_M + \sigma_R)\rho_k \Big) + \bar{\mu}^K \sigma_R \rho_K.
\end{align*}
From Equation~\eqref{eqn:grad_to_adj}, $\|\boldsymbol{\Phi}_r^T \mathbf{g}_{\mathrm{FOM}} - \mathbf{g}_{\mathrm{ROM}}\|_2 = \varepsilon_0$, which establishes Equation~\eqref{eqn:master_bound}.

To obtain the uniform closed-form bound in Equation~\eqref{eqn:uniform_bound}, substitute the uniform upper bounds into the summation:
\begin{equation*}
\sum_{k=0}^{K-1} \bar{\mu}^k \Lambda (\eta_{\mathrm{sub}, k} + \Delta_{\mathrm{TLM}, k}) \le \Lambda(\bar{\eta} + \bar{\Delta}) \sum_{k=0}^{K-1} \bar{\mu}^k = \Lambda(\bar{\eta} + \bar{\Delta}) \left(\frac{\bar{\mu}^K - 1}{\bar{\mu} - 1}\right).
\end{equation*}
Next, insert the uniform forward Grönwall bound $\rho_k \le \bar{\delta} \frac{C_M^k - 1}{C_M - 1}$ into the coupling term:
\begin{align*}
\sum_{k=0}^{K-1} \bar{\mu}^k \rho_k &\le \frac{\bar{\delta}}{C_M - 1} \sum_{k=0}^{K-1} \bar{\mu}^k (C_M^k - 1) \\
&= \frac{\bar{\delta}}{C_M - 1} \left[ \sum_{k=0}^{K-1} (\bar{\mu} C_M)^k - \sum_{k=0}^{K-1} \bar{\mu}^k \right] \\
&= \frac{\bar{\delta}}{C_M - 1} \left[ \frac{(\bar{\mu} C_M)^K - 1}{\bar{\mu} C_M - 1} - \frac{\bar{\mu}^K - 1}{\bar{\mu} - 1} \right].
\end{align*}
Finally, evaluating the terminal observation term $\bar{\mu}^K \sigma_R \rho_K \le \sigma_R \bar{\delta} \bar{\mu}^K \left(\frac{C_M^K - 1}{C_M - 1}\right)$ and combining these three terms yields Equation~\eqref{eqn:uniform_bound}.
\end{proof}

\paragraph{Continuous-Time Scaling across Physical Horizon.}
Let $T = K \Delta t$ denote the physical time duration of the assimilation window, where $\Delta t > 0$ is the discretization step size. For smooth underlying flow dynamics, the discrete forward and adjoint single-step multipliers scale as $C_M = 1 + L_f \Delta t + \mathcal{O}(\Delta t^2)$ and $\bar{\mu} = 1 + L_a \Delta t + \mathcal{O}(\Delta t^2)$, where $L_f$ and $L_a$ represent continuous-time Lyapunov-type forward and adjoint growth rates. Furthermore, the local one-step OpInf defect scales with the time step as $\bar{\delta} = \bar{\varepsilon}_m \Delta t + \mathcal{O}(\Delta t^2)$, where $\bar{\varepsilon}_m$ is the continuous model error rate per unit physical time. 

Under the same smooth-flow assumption, the one-step subspace and tangent-linear-model defects also scale as \(\bar{\eta}_{\rm sub}=\mathcal{O}(\Delta t)\) and \(\bar{\Delta}_{\rm TLM}=\mathcal{O}(\Delta t)\). Indeed, with \(\mathbf{M}_k=\mathbf{I}+\Delta t\,\mathbf{J}_k+\mathcal{O}(\Delta t^2)\),

$$
(\mathbf{I}-\mathbf{\Phi}_r\mathbf{\Phi}_r^T)\mathbf{M}_k\mathbf{\Phi}_r
=
\Delta t\,(\mathbf{I}-\mathbf{\Phi}_r\mathbf{\Phi}_r^T)\mathbf{J}_k\mathbf{\Phi}_r
+\mathcal{O}(\Delta t^2),
$$

since \((\mathbf{I}-\mathbf{\Phi}_r\mathbf{\Phi}_r^T)\mathbf{\Phi}_r=\mathbf{0}\). Likewise, if the reduced and projected full-order tangent maps are consistent one-step approximations of the underlying continuous dynamics, their difference is \(\mathcal{O}(\Delta t)\). Consequently, the factor \((\bar{\eta}_{\rm sub}+\bar{\Delta}_{\rm TLM})(\bar{\mu}^K-1)/(\bar{\mu}-1)\) remains finite as \(\Delta t\rightarrow0\) for fixed \(T=K\Delta t\).

In the limit as $\Delta t \to 0$ with fixed physical duration $T = K \Delta t$, the cross-coupling term in Equation~\eqref{eqn:uniform_bound} scales as:
\begin{equation}
(\bar{\mu} C_M)^K = \left( 1 + (L_f + L_a)\frac{T}{K} + \mathcal{O}(K^{-2}) \right)^K \xrightarrow{K \to \infty} \exp\big((L_f + L_a)T\big).
\label{eqn:exp_scaling}
\end{equation}
Because the discrete denominator scales as $C_M - 1 = L_f \Delta t$, the factor $\Delta t$ cancels cleanly against the discrete defect $\bar{\delta}$:
\begin{equation}
\frac{\bar{\delta}}{C_M - 1} = \frac{\bar{\varepsilon}_m \Delta t}{L_f \Delta t} = \frac{\bar{\varepsilon}_m}{L_f}.
\end{equation}
Consequently, the sensitivity error bound remains well-posed and independent of discretization step refinement:
\begin{itemize}
    \item \textbf{Long Windows ($T \gg (L_f + L_a)^{-1}$):} Cumulative error compounds as $\exp((L_f + L_a)T)$, degrading theoretical accuracy guarantees.
    \item \textbf{Short Windows ($T \ll (L_f + L_a)^{-1}$):} The exponential linearizes as $\exp((L_f + L_a)T) \approx 1 + (L_f + L_a)T$, restricting total sensitivity error to a benign, linear accumulation regime in physical time $T$.
\end{itemize}

\begin{remark}[Interpretation of the characteristic timescale]
The characteristic timescale suggested by the continuous-time bound should be interpreted as a conservative asymptotic guideline rather than a necessary condition on the correction-window length. The estimate is based on worst-case growth rates, through uniform bounds on the forward and adjoint dynamics, and therefore assumes that the largest admissible amplification persists throughout the assimilation window. In practice, the local dynamics may exhibit weaker growth, transient contraction, or cancellation of state and sensitivity errors, so successful correction may occur for windows longer than the nominal timescale. The estimate therefore characterizes when error accumulation is guaranteed to remain limited under worst-case assumptions, rather than defining a sharp threshold for variational correction.
\end{remark}

\begin{corollary}[Sufficient Condition for Physical Descent Direction]
\label{cor:descent}
Let $\mathbf{g}_{\mathrm{FOM}} \triangleq \nabla_{\mathbf{x}_{k_{\mathrm{corr}}}} \mathcal{J}(\boldsymbol{\Phi}_r \widehat{\mathbf{x}}_{k_{\mathrm{corr}}})$ and $\mathbf{g}_{\mathrm{ROM}} \triangleq \nabla_{\widehat{\mathbf{x}}_{k_{\mathrm{corr}}}} \widehat{\mathcal{J}}(\widehat{\mathbf{x}}_{k_{\mathrm{corr}}})$. Assume the current state is non-stationary in the reduced subspace, such that $\|\boldsymbol{\Phi}_r^T \mathbf{g}_{\mathrm{FOM}}\|_2 > 0$. If the correction window length $N_{\mathrm{window}}$ satisfies:
\begin{equation}
\mathcal{E}_{\mathrm{bound}} < \|\boldsymbol{\Phi}_r^T \mathbf{g}_{\mathrm{FOM}}\|_2,
\label{eqn:descent_condition}
\end{equation}
then the search direction $-\mathbf{g}_{\mathrm{ROM}}$ computed by the reduced OpInf adjoint yields a guaranteed descent direction for the true full-order observation functional:
\begin{equation}
\left. \frac{d}{d\alpha} \mathcal{J}(\boldsymbol{\Phi}_r \widehat{\mathbf{x}}_{k_{\mathrm{corr}}} - \alpha \boldsymbol{\Phi}_r \mathbf{g}_{\mathrm{ROM}}) \right|_{\alpha=0} = - \mathbf{g}_{\mathrm{FOM}}^T \boldsymbol{\Phi}_r \mathbf{g}_{\mathrm{ROM}} < 0.
\label{eqn:directional_deriv}
\end{equation}
\end{corollary}

\begin{proof}
Let $\mathbf{a} \triangleq \boldsymbol{\Phi}_r^T \mathbf{g}_{\mathrm{FOM}}$ and $\mathbf{b} \triangleq \mathbf{g}_{\mathrm{ROM}}$. Expanding the inner product via reverse Cauchy--Schwarz:
\begin{equation*}
\mathbf{a}^T \mathbf{b} = \|\mathbf{a}\|_2^2 + \mathbf{a}^T(\mathbf{b} - \mathbf{a}) \ge \|\mathbf{a}\|_2 \left( \|\mathbf{a}\|_2 - \|\mathbf{a} - \mathbf{b}\|_2 \right).
\end{equation*}
By Theorem~\ref{thm:main_bound}, $\|\mathbf{a} - \mathbf{b}\|_2 \le \mathcal{E}_{\mathrm{bound}}$. Since $\|\mathbf{a}\|_2 > 0$ by non-stationarity, condition~\eqref{eqn:descent_condition} ensures $\|\mathbf{a}\|_2 - \mathcal{E}_{\mathrm{bound}} > 0$, which yields $\mathbf{a}^T \mathbf{b} > 0$. Therefore:
\begin{equation*}
\nabla_{\mathbf{x}} \mathcal{J}(\boldsymbol{\Phi}_r \widehat{\mathbf{x}}_{k_{\mathrm{corr}}})^T (-\boldsymbol{\Phi}_r \mathbf{g}_{\mathrm{ROM}}) = -\mathbf{g}_{\mathrm{FOM}}^T \boldsymbol{\Phi}_r \mathbf{g}_{\mathrm{ROM}} = -\mathbf{a}^T \mathbf{b} < 0,
\end{equation*}
guaranteeing a strict reduction in the true physical cost for sufficiently small step size $\alpha > 0$.
\end{proof}

\begin{remark}[Geometric interpretation]
Corollary~\ref{cor:descent} provides an analytical sufficient condition rather than an online certification criterion, since $\mathbf{g}_{\mathrm{FOM}}$ may not be available during deployment. Its purpose is to characterize when the gradient computed using the OpInf model remains a descent direction for the corresponding full-order objective restricted to the POD subspace. Importantly, this condition does not require the ROM trajectory to remain globally accurate; it requires only sufficient local fidelity of the state and sensitivity information entering the variational gradient over the correction window. Geometrically,
\[
(\boldsymbol{\Phi}_r^{\top}\mathbf{g}_{\mathrm{FOM}})^{\top}
\mathbf{g}_{\mathrm{ROM}}
=
\|\boldsymbol{\Phi}_r^{\top}\mathbf{g}_{\mathrm{FOM}}\|_2
\|\mathbf{g}_{\mathrm{ROM}}\|_2 \cos\theta,
\]
where $\theta$ is the angle between the projected full-order gradient
$\boldsymbol{\Phi}_r^{\top}\mathbf{g}_{\mathrm{FOM}}$ and the reduced-order
gradient $\mathbf{g}_{\mathrm{ROM}}$. Hence, the descent property is retained
whenever $\theta < \pi/2$. The sufficient condition in Corollary~1 therefore
guarantees that the ROM gradient does not become sufficiently misaligned with
the full-order gradient projected onto the POD subspace to reverse the descent
direction. This condition is expected to remain satisfied when the state and
sensitivity errors remain controlled over the assimilation window, providing
a theoretical basis for the effectiveness of short correction windows even
when the uncontrolled ROM has lost long-horizon predictive accuracy. As
$\|\boldsymbol{\Phi}_r^{\top}\mathbf{g}_{\mathrm{FOM}}\|_2 \to 0$ near a
stationary point within the reduced subspace, the sufficient condition
naturally becomes more restrictive.
\end{remark}

\subsubsection{Theoretical Mechanism Supporting Extrapolative Forecasting}
\label{subsubsec:mechanism_explanation}
The analytical bounds derived above provide a mechanism for interpreting
why correction-windowed 4D-Var can remain effective in extrapolative regimes.
In particular, the state correction repeatedly resets the local FOM--ROM
trajectory discrepancy at the beginning of a correction window, while the
restricted adjoint horizon limits the amplification of sensitivity errors.

\begin{enumerate}
    \item \textbf{Forward Error Quenching (Gr\"onwall Cutoff)}: In extrapolative testing
regimes, the local OpInf modeling defect $\bar{\delta}$ is generally non-zero.
In open-loop simulation, these local defects accumulate according to
\[
\rho_k \leq
\frac{C_M^k-1}{C_M-1}\bar{\delta},
\qquad C_M \neq 1.
\]
At each state correction, the full-order and lifted reduced trajectories used
in the local analysis are initialized from the same corrected state, so that
$\rho_0=0$ by Assumption~3. The correction therefore restarts the forward-error
accumulation over a local horizon rather than allowing the Gr\"onwall
amplification to compound over the entire prediction interval.

   \item \textbf{Backward Sensitivity Stabilization (Adjoint Horizon Cutoff):} 
In extrapolative testing regimes characterized by elevated state amplitudes ($\|\widehat{\mathbf{x}}_k\|_2 \le R$), the local linear-quadratic OpInf Jacobian norm is elevated:
\begin{equation}
\bar{\mu} \triangleq \|\widehat{\mathbf{A}}\|_2 + 2\|\widehat{\mathbf{F}}\|_2 R.
\label{eqn:mu_extrap_eval}
\end{equation}
Over an extended assimilation horizon of physical duration $T = K \Delta t$, backward sensitivity propagation compounds through the adjoint resolvent as $\bar{\mu}^K = \exp\left(\frac{\ln \bar{\mu}}{\Delta t} T\right) \approx \exp(L_a T)$, where $L_a \approx (\bar{\mu} - 1)/\Delta t$ denotes the continuous logarithmic growth rate of the linearized adjoint dynamics. If $T \gg L_a^{-1}$, this exponential factor amplifies both the local model defects $\bar{\delta}$ and Jacobian mismatches $\bar{\Delta}_{\mathrm{TLM}}$, causing the theoretical error certificate $\mathcal{E}_{\mathrm{bound}}$ to inflate and potentially violating the descent condition $\mathcal{E}_{\mathrm{bound}} < \|\boldsymbol{\Phi}_r^T \mathbf{g}_{\mathrm{FOM}}\|_2$.

By restricting the correction window to a short physical duration satisfying $T \ll L_a^{-1}$, we exploit the first-order Taylor expansion:
\begin{equation}
\bar{\mu}^K = \exp(L_a T) = 1 + L_a T + \mathcal{O}\big((L_a T)^2\big).
\label{eqn:linear_taylor_adjoint}
\end{equation}
Substituting Equation~\eqref{eqn:linear_taylor_adjoint} into the geometric sum of the master bound demonstrates that cumulative sensitivity amplification remains strictly in the linear regime:
\begin{equation}
\frac{\bar{\mu}^K - 1}{\bar{\mu} - 1} \approx \frac{(1 + L_a T) - 1}{L_a \Delta t} = \frac{T}{\Delta t} = K.
\end{equation}
Truncating the backward adjoint horizon restricts the amplification of sensitivity errors and thereby limits the growth of the terms entering \(E_{\rm bound}\). This promotes satisfaction of the sufficient descent condition in Corollary~1, $E_{\rm bound} < \|\mathbf{\Phi}_r^T\mathbf{g}_{\rm FOM}\|_2$, under which the reduced 4D-Var update is guaranteed to provide a descent direction for the full-order objective restricted to the POD subspace.

\end{enumerate}

\subsection{Illustrative Example: The Duffing Oscillator}
\label{sec:illustrative}
We use the Duffing oscillator as a controlled test problem to demonstrate the full workflow: (i) generate reference trajectories, (ii) learn the system dynamics using OpInf, and (iii) apply correction-windowed 4D-Var assimilation to correct state conditions using sparse (in time) noisy observations. Since we use this example to demonstrate the applicability of the framework, we focus on system identification of the Duffing Oscillator rather than model reduction, so the learned model has dimension $r=2$, equal to the full system dimension. This problem is ideal for validation because it is low-dimensional ($n=2$ state variables), has explicit cubic nonlinearity, and allows each component of the method to be examined independently.

The Duffing oscillator is a damped dynamical system defined by a cubic restoring force. We formulate it as a discrete-time dynamical system:
\begin{equation}
\mathbf{x}_{k+1} = \mathbf{x}_k + \Delta t \, \mathbf{f}(\mathbf{x}_k),
\label{eq:duffing_discrete}
\end{equation}
where the state is $\mathbf{x}_k = [s_k~~v_k]^\top \in \mathbb{R}^2$ (position and velocity), $\Delta t = 0.01$ is the time step, and the governing vector field is:
\begin{equation}
\mathbf{f}(\mathbf{x}_k) =
\begin{bmatrix}
v_k \\
-\delta v_k - \alpha s_k - \beta s_k^3
\end{bmatrix}.
\end{equation}
Parameters are set to $\delta = 0.2$, $\alpha = 1.0$, and $\beta = 1.0$, which represent the damping coefficient, linear stiffness, and cubic stiffness, respectively. We generate a reference trajectory by advancing the system from initial condition $\mathbf{x}_0 = [2~~0]^\top$ for $N^{\text{total}} = 1000$ steps (spanning time $t \in [0, 10]$).

We organize the resulting trajectory as a pair of snapshot matrices following the standard notation from Section~\ref{sec:opinf}:
\begin{equation}
\mathbf{X} = \begin{bmatrix}
\mathbf{x}_0 & \mathbf{x}_1 & \cdots & \mathbf{x}_{N^{\text{total}}-1}
\end{bmatrix}
\in \mathbb{R}^{2 \times N^{\text{total}}}, \quad
\mathbf{X}' = \begin{bmatrix}
\mathbf{x}_1 & \mathbf{x}_2 & \cdots & \mathbf{x}_{N^{\text{total}}}
\end{bmatrix}
\in \mathbb{R}^{2 \times N^{\text{total}}},
\end{equation}
where each column $j$ of $\mathbf{X}$ contains $\mathbf{x}_j$ and the corresponding column of $\mathbf{X}'$ contains $\mathbf{x}_{j+1}$. This snapshot pairing directly enables OpInf regression. We allocate 75\% of snapshots for training and 25\% for validation, yielding $N^{\text{train}} = 750$, $N^{\text{val}} = 250$.

\subsubsection{System Identification via OpInf}
After obtaining the ground-truth data, we learn the Duffing dynamics from training snapshots using the same POD projection and regularised OpInf pipeline introduced in Sections~\ref{sec:rom} and~\ref{sec:opinf}. Since the system is already low-dimensional ($n=r=2$), this step is system identification rather than dimensionality reduction: the basis projection acts as a coordinate transform, and all state information is retained.

The Duffing dynamics are defined by a cubic restoring force; this nonlinearity is represented using the symmetric Kronecker product of the reduced state. The discrete-time reduced-order model is written as
\begin{equation}
    \widehat{\mathbf{x}}_{k+1} = \widehat{\mathbf{A}}\widehat{\mathbf{x}}_{k} + \widehat{\mathbf{F}}\bigl(\widehat{\mathbf{x}}_{k}\widetilde{\otimes}\widehat{\mathbf{x}}_{k}\widetilde{\otimes}\widehat{\mathbf{x}}_{k}\bigr),
    \label{eq:duffing_rom_dynamics}
\end{equation}
where $\widetilde{\otimes}$ denotes the symmetric Kronecker product. For a two-dimensional state vector $\widehat{\mathbf{x}}_k = [\widehat{x}_1, \widehat{x}_2]^\top$, the expanded cubic term $\widehat{\mathbf{x}}_k^{\widetilde{\otimes} 3} := \widehat{\mathbf{x}}_{k}\widetilde{\otimes}\widehat{\mathbf{x}}_{k}\widetilde{\otimes}\widehat{\mathbf{x}}_{k}$ evaluates to a $4 \times 1$ vector containing all unique polynomial permutations:
\begin{equation}
    \widehat{\mathbf{x}}_k^{\widetilde{\otimes} 3} = [\widehat{x}_1^3, \, \widehat{x}_1^2 \widehat{x}_2, \, \widehat{x}_1 \widehat{x}_2^2, \, \widehat{x}_2^3]^\top.
\end{equation}
To map this $4 \times 1$ nonlinear basis back to the two-dimensional state space, the inferred operator $\widehat{\mathbf{F}}$ is structured as a $2 \times 4$ matrix. The operators $\widehat{\mathbf{A}} \in \mathbb{R}^{2 \times 2}$ and $\widehat{\mathbf{F}} \in \mathbb{R}^{2 \times 4}$ are determined from data via the regularised least-squares formulation and validation-based tuning of $(\zeta_A,\zeta_F)$ specified in Equation~\eqref{eq:opinf_opt}. This estimated polynomial model subsequently serves as the model forecast operator for the windowed 4D-Var data assimilation step.

Using the identified model, we now demonstrate error-triggered, windowed 4D-Var assimilation. Rather than performing the optimisation across the entire trajectory, we employ an adaptive strategy: the background forecast is computed from the initial guess (typically different from the true initial condition), observations trigger local backward corrections only when forecast error exceeds the trigger threshold ($\tau$), and past trajectories are actively corrected within fixed-size correction windows, i.e., $N_{\text{window}}$ is fixed.

Since the Duffing oscillator is already low-dimensional ($n=2$ with $r=2$), the observation operator becomes $\mathbf{H} = \mathbf{\Phi} \in \mathbb{R}^{2 \times 2}$, meaning we observe the full state directly. This differs from high-dimensional systems, where the column-pivoted QR strategy (as explained in Section~\ref{sec:sensor_placement}) informative sparse measurement locations. We reiterate that no sensor-selection algorithm is used in this illustrative example.

The background state is initialised from the last snapshot of the training trajectory, which is intentionally different from the true initial condition used in the assimilation test case. This gives the ROM a realistic but imperfect starting point. Here, the training data are generated from a separate reference trajectory of the Duffing system, whereas the truth trajectory is reserved for the assimilation test case and for generating synthetic observations. We generate synthetic observations by: (i) forward-integrating the truth using the FOM, (ii) subsampling at regular intervals with $f_{\text{obs}} = 1/30$ (meaning that observations happen at every 30 time steps) from 2.5 time units, and (iii) adding zero-mean Gaussian noise with standard deviation $\sigma_{\text{obs}} = 0.5$. An observational covariance matrx  $\mathbf{R}=\sigma_{\text{obs}}^2\mathbf{I}$ and the background covariance used was taken to have a covariance of $\mathbf{B} = \mathbf{I}$ (Identity Matrix) are used during data assimilation.

Following the error-triggered strategy outlined in Section~\ref{sec:windowing}, when the forecast error exceeds the trigger threshold $\tau = 0.1$, a 4D-Var correction window is activated spanning the past 20 time steps ($N_{\text{window}} = 20$). Within each correction window, we solve a local optimisation problem: minimise the cost functional given in equation~\eqref{eqn:4dvar_cost} but only over the correction window time steps.

Solving this local optimisation requires computing gradients of the cost with respect to the initial condition at the correction-window start (denoted by time step $k_{\text{corr}}$, where $k_{\text{corr}} = \max(0, k_{\text{obs}} - N_{\text{window}}+1)$). Rather than using gradient computation methods such as finite differences or accessing the original full-order model's code, we leverage a key advantage of the learned OpInf model: we can explicitly access its polynomial form. This means we can differentiate it to obtain the required Jacobians.

The Jacobian of the cubic term with respect to the reduced state,
evaluated at $\widehat{\mathbf{x}}_k$, is obtained using
Kronecker-product identities as
\begin{equation}
\left.\frac{\partial (\widehat{\mathbf{x}}\widetilde{\otimes}\widehat{\mathbf{x}}\widetilde{\otimes}\widehat{\mathbf{x}})}{\partial \widehat{\mathbf{x}}}\right|_{\widehat{\mathbf{x}}_k} =
I_r \widetilde{\otimes}\widehat{\mathbf{x}}_k \widetilde{\otimes} \widehat{\mathbf{x}}_k + \widehat{\mathbf{x}}_k\widetilde{\otimes}I_r \widetilde{\otimes}\widehat{\mathbf{x}}_k + \widehat{\mathbf{x}}_k\widetilde{\otimes}\widehat{\mathbf{x}}_k\widetilde{\otimes}I_r,
\end{equation}
where $I_r$ is the $r \times r$ identity matrix. For a two-dimensional state $r=2$ with $\widehat{\mathbf{x}}_k = [x_1, x_2]^\top$, the symmetric cubic basis is the $4 \times 1$ vector $\widehat{\mathbf{x}}_k^{\widetilde{\otimes}3} = [x_1^3,\, x_1^2 x_2,\, x_1 x_2^2,\, x_2^3]^\top$. The resulting Jacobian is therefore a $4 \times 2$ matrix:
\begin{equation}  \left.\frac{\partial \widehat{\mathbf{x}}^{\widetilde{\otimes}3}}{\partial \widehat{\mathbf{x}}}\right\vert{}_{\widehat{\mathbf{x}}_k} = \begin{bmatrix} 3x_1^2 & 2x_1x_2 & x_2^2 & 0 \\ 0 & x_1^2 & 2x_1x_2 & 3x_2^2 \end{bmatrix}^\top
\label{eq:cubic_jacobian}
\end{equation}

The forward propagation yields the state Jacobian at each time step $k$:
\begin{equation}
\frac{\partial \widehat{\mathcal{M}}}{\partial \widehat{\mathbf{x}}_k} = \widehat{\mathbf{A}} + \widehat{\mathbf{F}}\left.\frac{\partial \widehat{\mathbf{x}}^{\widetilde{\otimes}3}}{\partial \widehat{\mathbf{x}}}\right|_{\widehat{\mathbf{x}}_k},
\end{equation}
where $\widehat{\mathbf{A}}$ and $\widehat{\mathbf{F}}$ are the learned OpInf operators from equation~\eqref{eq:duffing_rom_dynamics}.

After obtaining these Jacobians, we apply the adjoint method (Section~\ref{sec:proposed}) to propagate sensitivity information backwards through the correction window using the gradients computed via the Adjoint equation and backpropagation. Once the adjoint has propagated to the correction-window start, the gradient with respect to the correction-window 
 is passed to the L-BFGS optimiser for minimisation. The correction window is then optimised to minimise the local cost function, after which we update the trajectory and continue the forecast until the next trigger event.
\begin{figure}[htb]
\centering
\begin{tikzpicture}
    \begin{groupplot}[
    group style={
        group size=1 by 2,
        vertical sep=1.5cm,
        xlabels at=edge bottom,
        xticklabels at=edge bottom
    }, 
    height=5.5cm, 
    width=\linewidth, 
    grid=major,
    grid style={dotted, gray!50},
    xlabel={Time},
    unbounded coords=jump,
    table/col sep=comma,
    name=mainplot 
]

    \nextgroupplot[
        ylabel={Displacement $x_t$},
        title style={at={(0,1.2)}, anchor=north west, font=\bfseries},
        title={(a)}
    ]
    \addplot[color=truthColor, line width=3.2pt, opacity=0.6] 
        table[x=t, y=x_true] {Data_sheets/trajectory_data.csv};
    \node[color=truthColor, anchor=south east] at (axis cs: 0.2, 1.4) {\textbf{Truth}};

    \addplot[color=assimColor, line width=2pt] 
        table[x=t, y=x_opt] {Data_sheets/trajectory_data.csv};
    \node[color=assimColor, anchor=north east] at (rel axis cs: 1.001, 0.25) {\textbf{OpInf-DA}};

    \addplot[color=romForecastColor, line width=2pt, loosely dotted] 
        table[x=t, y=x_guess] {Data_sheets/trajectory_data.csv};
    \node[color=romForecastColor, anchor=north east] at (rel axis cs: 0.15, 0.4) {\textbf{OpInf}};

    \addplot[only marks, mark=o, mark size=2.4pt, line width=2pt, color=obsColor, fill=white] 
        table[x=t, y=x_obs] {Data_sheets/trajectory_data.csv};
        \node[color=obsColor, anchor=north east] at (rel axis cs: 0.83, 0.8) {\textbf{Obs}};

    
    \pgfplotstableread[col sep=comma]{Data_sheets/trigger_events.csv}\triggertable
    \pgfplotstablegetrowsof{\triggertable}
    \pgfmathtruncatemacro{\numtriggers}{\pgfplotsretval-1}
    \foreach \t in {0,...,\numtriggers}{
       \pgfplotstablegetelem{\t}{trigger_time}\of{\triggertable}
        \edef\trigTime{\pgfplotsretval}
        \pgfplotstablegetelem{\t}{correction_time}\of{\triggertable}
        \edef\corrTime{\pgfplotsretval}
      \addplot[forget plot, fill=corrWindowColor, fill opacity=0.15, draw=none] coordinates {
            (\corrTime,-2) (\corrTime,2) (\trigTime,2) (\trigTime,-2)
        };
    }


    \nextgroupplot[
        ylabel={Velocity $v_t$},
        title style={at={(0,1.2)}, anchor=north west, font=\bfseries},
        title={(b)}
    ]
    \addplot[color=truthColor, line width=3.2pt, opacity=0.6] 
        table[x=t, y=v_true] {Data_sheets/trajectory_data.csv};
    \node[color=truthColor, anchor=south east] at (axis cs: 0.5, -0.1) {\textbf{Truth}};

    \addplot[color=assimColor, line width=2pt] 
        table[x=t, y=v_opt] {Data_sheets/trajectory_data.csv};
    \node[color=assimColor, fill=white, opacity=0.8, anchor=west] at (rel axis cs: 0.88, 0.35) {\textbf{OpInf-DA}};
    
    \addplot[color=romForecastColor, line width=2pt, loosely dotted] 
        table[x=t, y=v_guess] {Data_sheets/trajectory_data.csv};
    \node[color=romForecastColor, anchor=north east] at (rel axis cs: 0.08, 0.9) {\textbf{OpInf}};

    \addplot[only marks, mark=o, mark size=2.4pt, line width=2pt, color=obsColor, fill=white] 
        table[x=t, y=v_obs] {Data_sheets/trajectory_data.csv};
    \node[color=obsColor, anchor=north east] at (rel axis cs: 0.8, 0.35) {\textbf{Obs}};

    \pgfplotstableread[col sep=comma]{Data_sheets/trigger_events.csv}\triggertable
    \pgfplotstablegetrowsof{\triggertable}
    \pgfmathtruncatemacro{\numtriggers}{\pgfplotsretval-1}
    \foreach \t in {0,...,\numtriggers}{
        \pgfplotstablegetelem{\t}{trigger_time}\of{\triggertable}
        \edef\trigTime{\pgfplotsretval}
        \pgfplotstablegetelem{\t}{correction_time}\of{\triggertable}
        \edef\corrTime{\pgfplotsretval}
        \addplot[forget plot, fill=corrWindowColor, fill opacity=0.15, draw=none] coordinates {
            (\corrTime,-2.5) (\corrTime,2.5) (\trigTime,2.5) (\trigTime,-2.5)
        };
    }


    \end{groupplot}

\end{tikzpicture}
\caption{Error-triggered correction-windowed 4D-Var assimilation for the
Duffing oscillator. (a) Displacement and (b) velocity trajectories comparing
the reference solution, the open-loop OpInf forecast, and the corrected
OpInf-DA trajectory. Circles denote the noisy observations available
intermittently in time, and the shaded regions indicate activated correction
windows. The state corrections repeatedly re-align the learned-model
trajectory with the reference solution after the open-loop forecast departs
from it.}
\label{fig:duffing_4dvar}
\end{figure}
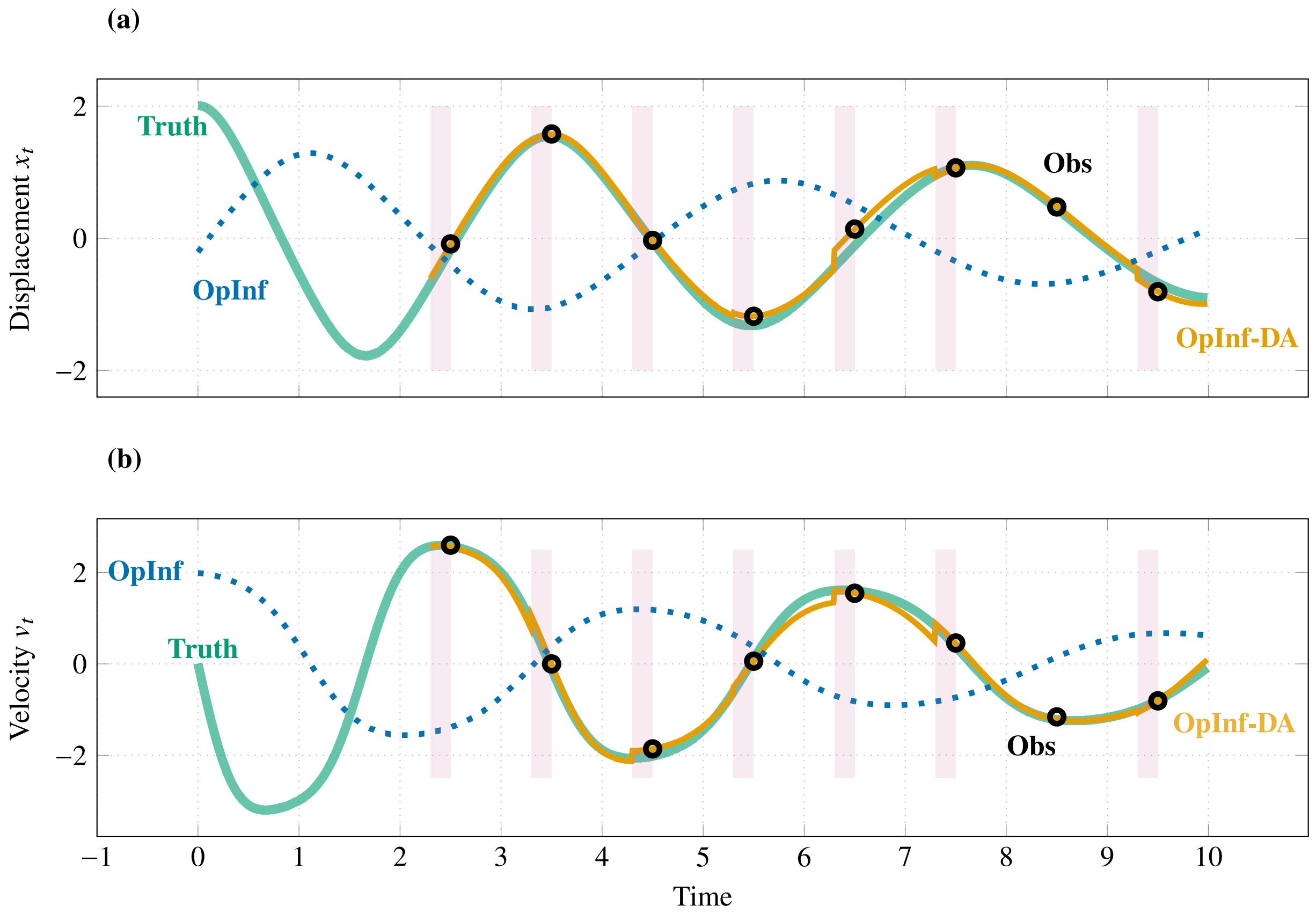
Figure~\ref{fig:duffing_4dvar} shows the displacement and velocity trajectories comparing the OpInf-based ROM background forecast, the assimilated trajectory after corrections, and the truth. The background forecast, initialised from the poor guess, diverges significantly from the truth in the early phase. The sparse observations (shown as circles) provide only intermittent information. Whenever forecast error exceeds the threshold $\tau = 0.1$, a correction window spanning the past $N_{\text{window}}=20$ steps is activated (shaded band). Within each window, 4D-Var retroactively adjusts the starting state to better fit observations. After each correction, the state is re-integrated forward, yielding an improved trajectory that better matches observations. As the simulation progresses, the assimilated trajectory successively reduces error through these adaptive corrections, demonstrating that sparse observations and backwards-looking corrections can guide a poor forecast toward the truth. Notably, around $t \approx 8.5$ time units, the forecast error remains below the trigger threshold $\tau$, correctly avoiding unnecessary corrections and demonstrating the efficiency of the error-triggered approach in skipping redundant optimisations during periods when the ROM maintains adequate forecast accuracy.

\subsection{Theoretical Interpretation of the 2D Duffing Oscillator}
\label{subsec:duffing_verification}

The 2D Duffing oscillator provides a direct, low-dimensional testbed ($n = r = 2$) where the theoretical assumptions and constituent quantities of the gradient error framework can be verified in closed form. Because model reduction is identity-equivalent ($\boldsymbol{\Phi}_r = \mathbf{I}_2$), several key terms simplify:

\begin{itemize}
    \item \textbf{Subspace Defect Collapse:} 
    Because $\boldsymbol{\Phi}_r \boldsymbol{\Phi}_r^T = \mathbf{I}_2$, both the out-of-subspace gradient component in Theorem~\ref{thm:decomp} and the tangent linear subspace invariance defect vanish identically:
    \begin{equation}
    \|(\mathbf{I}_2 - \boldsymbol{\Phi}_r \boldsymbol{\Phi}_r^T) \mathbf{g}_{\mathrm{FOM}}\|_2 \equiv 0, \qquad \eta_{\mathrm{sub}, k} \triangleq \|(\mathbf{I}_2 - \boldsymbol{\Phi}_r \boldsymbol{\Phi}_r^T)\mathbf{M}_k(\mathbf{x}_k)\boldsymbol{\Phi}_r\|_2 \equiv 0.
    \end{equation}
    Consequently, the full-space error equals the projected adjoint error: $\|\mathbf{g}_{\mathrm{FOM}} - \boldsymbol{\Phi}_r \mathbf{g}_{\mathrm{ROM}}\|_2 = \|\boldsymbol{\Phi}_r^T \mathbf{g}_{\mathrm{FOM}} - \mathbf{g}_{\mathrm{ROM}}\|_2 = \varepsilon_0$. Furthermore, since $\mathbf{x}^b \in \operatorname{range}(\boldsymbol{\Phi}_r) = \mathbb{R}^2$ and $\widehat{\mathbf{B}}^{-1} = \boldsymbol{\Phi}_r^T \mathbf{B}^{-1} \boldsymbol{\Phi}_r = \mathbf{B}^{-1}$, the background prior mismatch vanishes ($E_{\mathrm{prior}} = 0$).

    \item \textbf{Analytical Operators and Defect Quantification:}
    The forward discrete FOM map is given by the forward Euler discretization $\mathcal{M}(\mathbf{x}_k) = \mathbf{x}_k + \Delta t \, f(\mathbf{x}_k)$, while the learned cubic OpInf surrogate evaluates to $\widehat{\mathcal{M}}(\widehat{\mathbf{x}}_k) = \widehat{\mathbf{A}}\widehat{\mathbf{x}}_k + \widehat{\mathbf{F}}\widehat{\mathbf{x}}_k^{\tilde{\otimes} 3}$. The local one-step OpInf defect $\delta_{\mathrm{OpInf}}(k) = \|\mathcal{M}(\widehat{\mathbf{x}}_k) - \widehat{\mathcal{M}}(\widehat{\mathbf{x}}_k)\|_2 \le \bar{\delta}$ reflects polynomial regression truncation on the compact attractor $\mathcal{D} = \{ [s, v]^T : |s| \le 2, |v| \le 3.5 \}$. Using the explicit cubic Jacobian from Equation~\eqref{eq:cubic_jacobian}, the Tangent Linear Model mismatch evaluates directly along the trajectory:
    \begin{equation}
    \Delta_{\mathrm{TLM}, k} = \|\mathbf{M}_k(\widehat{\mathbf{x}}_k) - \widehat{\mathbf{M}}_k(\widehat{\mathbf{x}}_k)\|_2 \le \bar{\Delta}_{\mathrm{TLM}},
    \end{equation}
    where $\mathbf{M}_k(\widehat{\mathbf{x}}_k) = \mathbf{I}_2 + \Delta t \, \nabla f(\widehat{\mathbf{x}}_k)$ is the exact FOM derivative and $\widehat{\mathbf{M}}_k(\widehat{\mathbf{x}}_k) = \widehat{\mathbf{A}} + \widehat{\mathbf{F}}\frac{\partial \widehat{\mathbf{x}}^{\tilde{\otimes} 3}}{\partial \widehat{\mathbf{x}}}|_{\widehat{\mathbf{x}}_k}$ is the evaluated OpInf Jacobian.

\end{itemize}

The Duffing example illustrates the behavior predicted by the preceding analysis. The bounds derived above are conservative sufficient estimates and are not intended here as sharp trajectory-specific certificates. Instead, the numerical results demonstrate that restricting assimilation to local correction windows prevents the extrapolating OpInf trajectory from accumulating large long-horizon errors. Repeated corrections consistently return the learned-model trajectory toward the reference solution, supporting the proposed mechanism that limiting the forecast and adjoint horizons mitigates the accumulation and amplification of reduced-model errors during data assimilation.

In addition, we numerically assess the gradient-alignment result, we compared the ROM gradient with the full-order gradient projected onto the reduced subspace. The two gradients are nearly collinear, with \(\cos\theta=0.999986\) (\(\theta=0.302^\circ\)), confirming that the ROM gradient retains a descent direction for the projected full-order objective.

This example validates the error-triggered correction-windowed approach: (i) the framework correctly identifies when forecast error warrants correction via the trigger-threshold $\tau$ mechanism, (ii) backwards-looking correction windows efficiently correct accumulated drift without requiring global optimisation, and (iii) the analytically computed adjoint enables fast convergence within each correction window. Most importantly, this demonstrates that even with significant initial errors and sparse observations (in time), adaptive correction-windowed assimilation remains computationally feasible. We now test the approach on realistic high-dimensional systems across multiple scales.

\section{Results}
\label{sec:results}
We next evaluate the proposed state-correction framework on two systems that progressively test its ability to extend the predictive horizon of a fixed ROM beyond its training regime. Burgers' equation provides a controlled nonlinear PDE in which the ROM is deliberately propagated into post-shock dynamics, while sea-surface-temperature data provide a realistic high-dimensional forecasting problem.

\subsection{Burgers' Equation: Correction-Windowed Assimilation Analysis}
\label{sec:burgers}
The viscous Burgers' equation provides a controlled setting in which to test the central extrapolation problem considered in this work. Starting from a smooth initial condition, the solution undergoes nonlinear gradient steepening and develops a shock-like structure. We construct the POD basis and learn the OpInf operators using trajectory data through the smooth evolution up to the onset of shock formation. The reduced basis and learned operators are then held fixed, and the same ROM is tasked with predicting the subsequent post-shock evolution. This regime is challenging for the autonomous ROM because the sharp localized structures that develop after shock formation are not well represented by the low-dimensional model constructed from the preceding trajectory. The experiment therefore asks whether intermittent state correction from subsequently acquired observations can extend the useful predictive horizon of this fixed ROM without retraining the operators or modifying the reduced basis. The viscous Burgers' equation governs the velocity field $u(x,t)$:

\begin{equation}
    \frac{\partial u}{\partial t} + u \frac{\partial u}{\partial x} = \nu \frac{\partial^2 u}{\partial x^2}, \quad x \in [0, L], \quad t \in [0, T].
    \label{eq:burgers_pde}
\end{equation}

We set the viscosity coefficient to $\nu=0.005$ and impose homogeneous
Dirichlet boundary conditions,
$u(0,t)=u(L,t)=0$, on a domain of length $L=1$. The system is initialized
from the smooth profile $u(x,0)=\sin(2\pi x)$. The spatial domain is
discretized using $n_x=128$ uniformly spaced grid points, with central finite
differences used for the advective and diffusive terms. The resulting
semidiscrete system is integrated from $t=0$ to $T=0.6$ using a fourth-order
Runge--Kutta scheme with a time step of $\Delta t=0.001$.

The reduced model uses $r=4$ POD modes, which capture $99.98\%$ of the
snapshot energy. Consistent with the quadratic nonlinearity of Burgers'
equation, the reduced dynamics are learned using the linear--quadratic OpInf
formulation described in Section~\ref{sec:opinf}. The model-construction
interval consists of the first $40\%$ of the trajectory, corresponding to
$t\leq0.24$, through the smooth evolution up to the onset of shock formation.
Within this interval, a validation subset comprising $10\%$ of the available
snapshots is used to select the Tikhonov regularisation parameters through
grid search. After $t=0.24$, the POD basis and learned OpInf operators are held
fixed; no post-shock data are used to update the basis or retrain the reduced
dynamics.

For sparse assimilation, we use $m=r=4$ physical observation locations
selected using the CPQR procedure described in
Section~\ref{sec:sensor_placement}. This sparse configuration is compared
against assimilation using observations at all $n_x=128$ spatial locations,
which serves as a full-state observation benchmark. The same pre-trained ROM
is used in both cases; only the spatial density of observations available
during assimilation differs. An observational covariance matrx  $\mathbf{R}=\sigma_{\text{obs}}^2\mathbf{I}$, with $\sigma_{\text{obs}} = 2 \times 10^{-2}$ and the background covariance used was taken to have a covariance of $\widehat{\mathbf{B}} = \mathbf{I}_r$ (Identity Matrix, As per assumption 2), following the Assumption 2 from the previous section.

\subsubsection*{Assimilation Results and Comparative Analysis}
\label{subsubsec:Burgers_results}

\begin{figure}[H]
    \centering
    \begin{subfigure}[t]{0.49\linewidth}
        \centering
        \includegraphics[width=\linewidth, height=6cm]{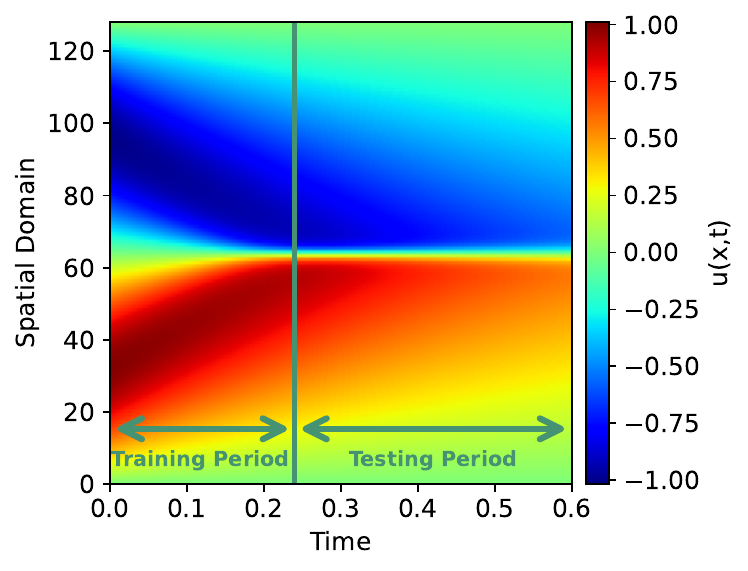}
        \caption{Ground truth solution evolution across the full time frame}
        \label{fig:burgers_truth}
    \end{subfigure}
    \hfill
    \begin{subfigure}[t]{0.49\linewidth}
        \centering
        \includegraphics[width=\linewidth, height=6cm]{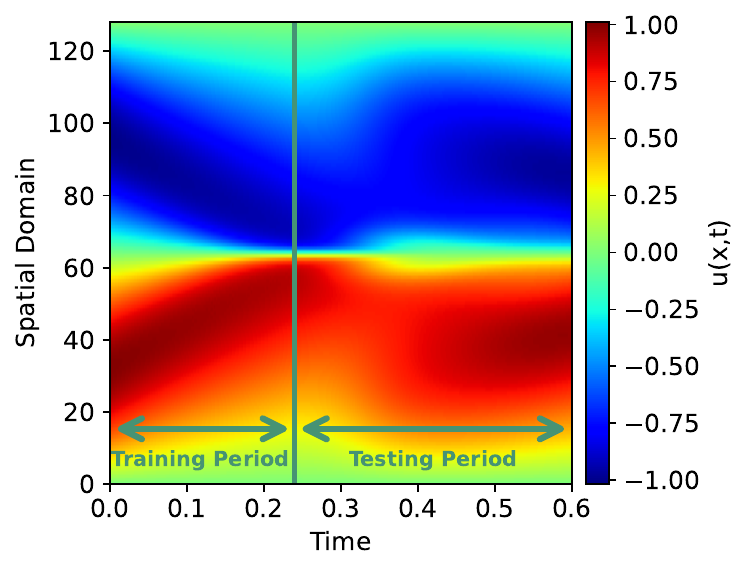}
        \caption{OpInf-ROM forecast without assimilation}
        \label{fig:burgers_rom}
    \end{subfigure}
\end{figure}

\begin{figure}[H]
    \ContinuedFloat
    \centering
    \begin{subfigure}[t]{0.49\linewidth}
        \centering
        \includegraphics[width=\linewidth, height=6cm]{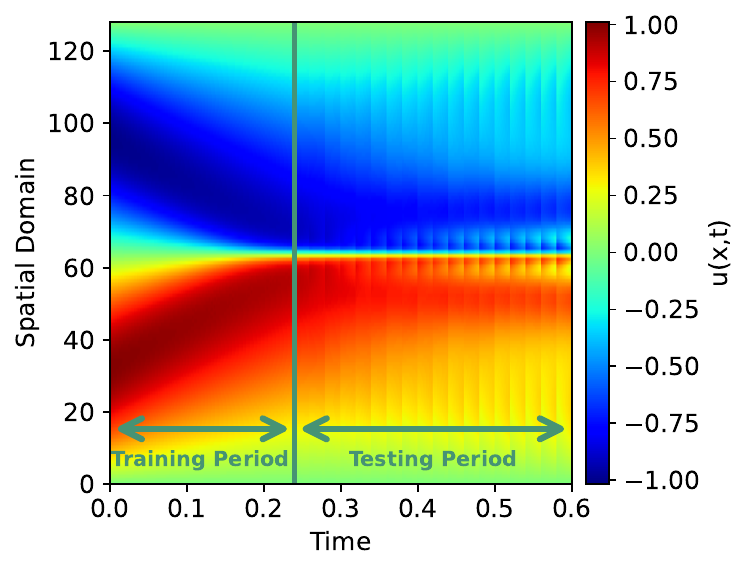}
        \caption{Windowed assimilation with full-state observations}
        \label{fig:burgers_fullstate}
    \end{subfigure}
    \hfill
    \begin{subfigure}[t]{0.49\linewidth}
        \centering
        \includegraphics[width=\linewidth, height=6cm]{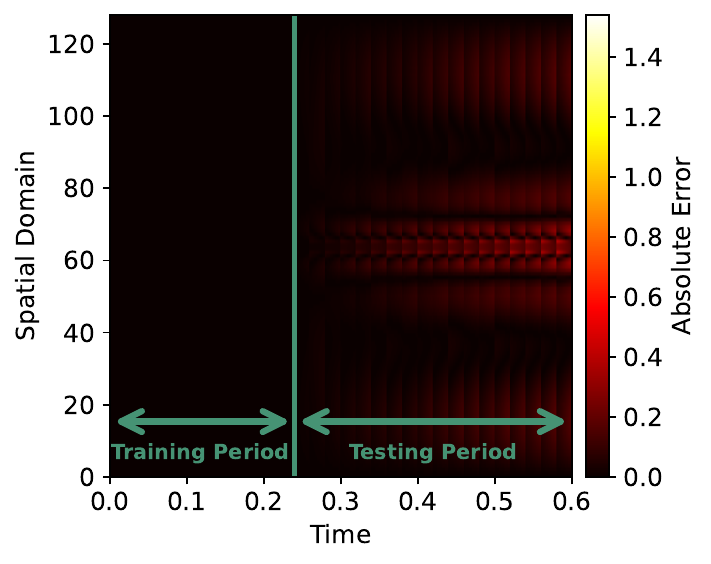}
        \caption{Prediction error with full-state observations}
        \label{fig:burgers_fullstate_error}
    \end{subfigure}

\end{figure}
\begin{figure}[H]
    \ContinuedFloat
    \centering
    \begin{subfigure}[t]{0.49\linewidth}
        \centering
        \includegraphics[width=\linewidth, height=6cm]{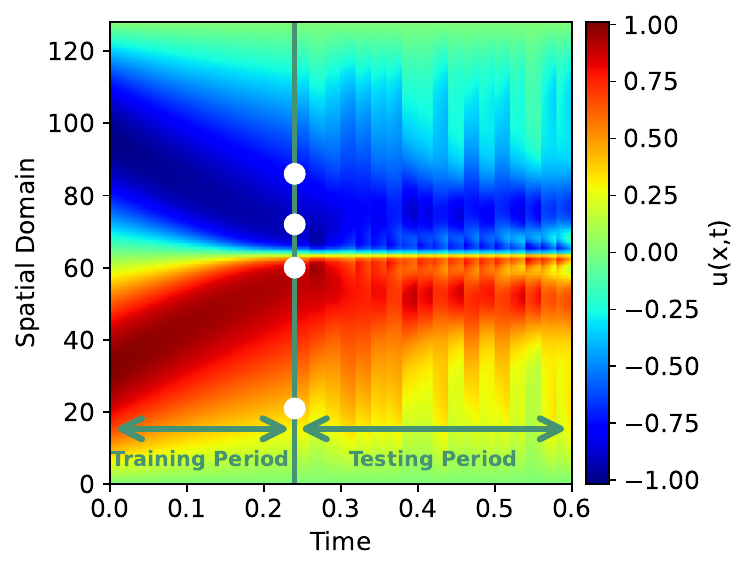}
        \caption{Windowed assimilation with QR-selected sensor locations}
        \label{fig:burgers_qr}
    \end{subfigure}
    \hfill
    \begin{subfigure}[t]{0.49\linewidth}
        \centering
        \includegraphics[width=\linewidth, height=6cm]{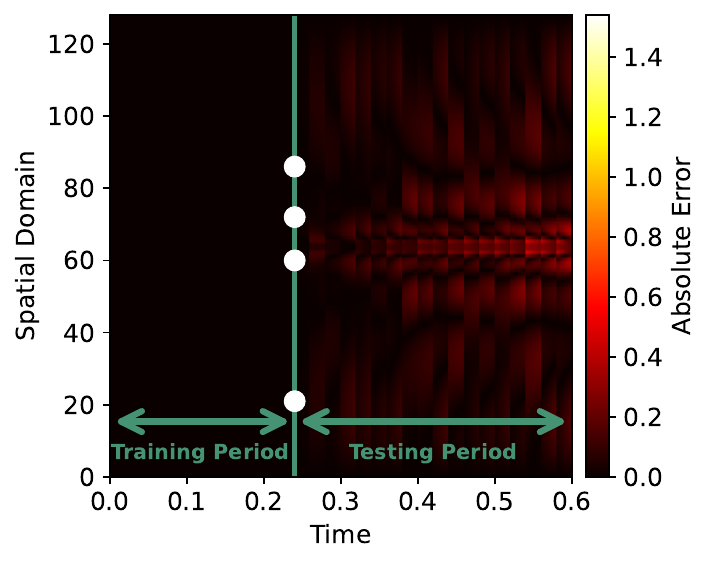}
        \caption{Prediction error with QR-selected sensor observations}
        \label{fig:burgers_qr_error}
    \end{subfigure}
    \hfill

    \caption{Burgers' equation correction-windowed assimilation results.
(a) Ground-truth evolution over the full simulation interval.
(b) Autonomous OpInf forecast, showing increasing error in the post-training
regime.
(c) Corrected trajectory using full-state observations and
(d) the corresponding absolute error.
(e) Corrected trajectory using the $m=4$ CPQR-selected observations and
(f) the corresponding absolute error.
Both observation configurations substantially reduce the post-training error
relative to the autonomous ROM, with the sparse configuration recovering
behavior close to the full-state reference.}
    \label{fig:burgers_dashboard}
\end{figure}

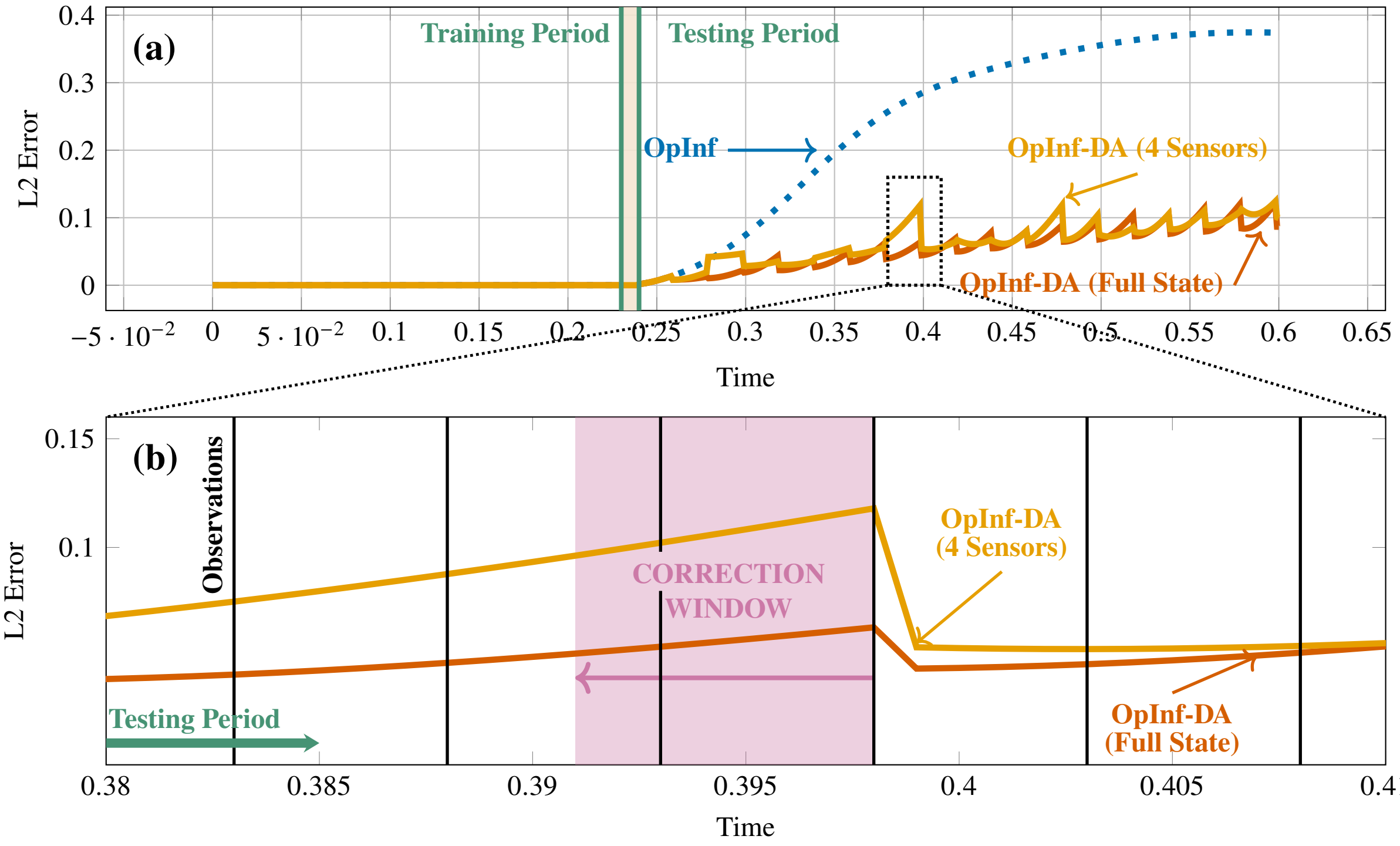
\begin{figure}[htb]
    \centering
    \begin{tikzpicture}
    \begin{axis}[
        name=mainplot,
        xlabel={Time},
        ylabel={L2 Error},
        legend style={draw=none},
        grid,
        width=16cm,
        height=5cm,
        font=\small,
    ]

    \addplot [color=romForecastColor, line width=2pt, mark=none, loosely dotted] table[x=time, y=ROM, col sep=comma] {Data_sheets/error_timeseries_ROM4.csv};

    \addplot [color=fullstatecolor, line width=2pt, mark=none] table[x=time, y=FullState_128, col sep=comma] {Data_sheets/error_timeseries_ROM4.csv};
    
    \fill[phaseboxxbrown, opacity=0.9] (axis cs:0.23,-1.0) rectangle (axis cs:0.24,0.6);
    
    \addplot [color=assimColor, line width=2pt, mark=none] table[x=time, y=QR_Picked_4, col sep=comma] {Data_sheets/error_timeseries_ROM4.csv};
    \draw [color=trainingtesting, line width=1.5pt] (axis cs: 0.23, -0.5) -- (axis cs: 0.23, 0.5);
    \draw [color=trainingtesting, line width=1.5pt] (axis cs: 0.24, -0.5) -- (axis cs: 0.24, 0.5);
    \node[anchor=east, color=trainingtesting, font=\small\bfseries] at (axis cs: 0.23, 0.37) {Training Period};
    \node[anchor=west, color=trainingtesting, font=\small\bfseries] at (axis cs: 0.25, 0.37) {Testing Period};
    
    \node[anchor=east, color=romForecastColor, font=\small\bfseries] (romLabel) at (axis cs: 0.29, 0.2) {OpInf};
    \draw[->, color=romForecastColor, line width=1pt] (romLabel.east) -- (axis cs: 0.34, 0.2);
    
    \node[anchor=east, color=fullstatecolor, font=\small\bfseries] (fsLabel) at (axis cs: 0.575, 0) {OpInf-DA (Full State)};
    \draw[->, color=fullstatecolor, line width=1pt] (fsLabel.east) -- (axis cs: 0.59, 0.08);
    
    \node[anchor=east, color=assimColor, font=\small\bfseries] (qrLabel) at (axis cs: 0.6, 0.2) {OpInf-DA (4 Sensors)};
    \draw[->, color=assimColor, line width=1pt] (qrLabel.south) -- (axis cs: 0.48, 0.13);

    \coordinate (zoomBoxSW) at (axis cs: 0.38, 0.0);
    \coordinate (zoomBoxNE) at (axis cs: 0.41, 0.16);
    \coordinate (zoomBoxSE) at (axis cs: 0.41, 0.0);
    \draw[densely dotted, color=black, line width=1pt] (zoomBoxSW) rectangle (zoomBoxNE);

    \end{axis}

    \node[anchor=north west, font=\large\bfseries, xshift=5pt, yshift=-5pt] at (mainplot.north west) {(a)};

    \begin{axis}[
        name=zoomplot,
        at={(mainplot.south west)},
        anchor=north west,
        yshift=-1.2cm,
        xlabel={Time},
        ylabel={L2 Error},
        legend style={draw=none},
        width=16cm,
        height=5.5cm,
        font=\small,
        xmin=0.38,
        xmax=0.41,
        ymin=0.0,
        ymax=0.16,
        clip=true,
        scaled x ticks=false,
        xticklabel={\pgfmathprintnumber[precision=3,fixed]{\tick}},
        xtick distance=0.005,
        ytick = {0.1, 0.15}
    ]

    \fill[corrWindowColor, opacity=0.35] (axis cs:0.391,0.0) rectangle (axis cs:0.398,0.16);

    \node[
        font=\small\bfseries,
        text=corrWindowColor,
        anchor=center,
        align=center,
    ] at (axis cs:0.3946,0.08) {CORRECTION\\WINDOW};

    \draw[<-, color=corrWindowColor, line width=1.5pt]
        (axis cs:0.391,0.04) -- (axis cs:0.398,0.04);


    \addplot [color=romForecastColor, line width=2pt, mark=none, loosely dotted] table[x=time, y=ROM, col sep=comma] {Data_sheets/error_timeseries_ROM4.csv};

    \addplot [color=fullstatecolor, line width=2pt, mark=none] table[x=time, y=FullState_128, col sep=comma] {Data_sheets/error_timeseries_ROM4.csv};

    \addplot [color=assimColor, line width=2pt, mark=none] table[x=time, y=QR_Picked_4, col sep=comma] {Data_sheets/error_timeseries_ROM4.csv};

    \node[anchor=center, color=fullstatecolor, font=\small\bfseries, align=center] (zoomFsLabel) at (axis cs: 0.405, 0.016) {OpInf-DA  \\[-2 pt] (Full State)};
    \draw[->, color=fullstatecolor, line width=1pt] (zoomFsLabel.north) -- (axis cs: 0.407, 0.05);

    \node[anchor=center, color=assimColor, font=\small\bfseries, align=center, line width=0.5pt] (zoomQrLabel) at (axis cs: 0.401, 0.106) {OpInf-DA \\[-2 pt] (4 Sensors)};
    \draw[->, color=assimColor, line width=1pt] (zoomQrLabel.south) -- (axis cs: 0.399, 0.056);

    \tikzset{obsline/.style={obsColor, solid, line width=1pt}}
    
    \draw[obsline] (axis cs:0.353,0.0) -- (axis cs:0.353,0.16);
    \node[anchor=center, color=obsColor, font=\small\bfseries, rotate=90] at (axis cs: 0.3825, 0.115) {Observations};
    \draw[obsline] (axis cs:0.358,0.0) -- (axis cs:0.358,0.16);
    \draw[obsline] (axis cs:0.363,0.0) -- (axis cs:0.363,0.16);
    \draw[obsline] (axis cs:0.368,0.0) -- (axis cs:0.368,0.16);
    \draw[obsline] (axis cs:0.373,0.0) -- (axis cs:0.373,0.16);
    \draw[obsline] (axis cs:0.378,0.0) -- (axis cs:0.378,0.16);
    \draw[obsline] (axis cs:0.383,0.0) -- (axis cs:0.383,0.16);
    \draw[obsline] (axis cs:0.388,0.0) -- (axis cs:0.388,0.16);
    \draw[obsline] (axis cs:0.393,0.0) -- (axis cs:0.393,0.08);
    \draw[obsline] (axis cs:0.393,0.098) -- (axis cs:0.393,0.16);
    \draw[obsline] (axis cs:0.398,0.0) -- (axis cs:0.398,0.16);
    \draw[obsline] (axis cs:0.403,0.0) -- (axis cs:0.403,0.16);
    \draw[obsline] (axis cs:0.408,0.0) -- (axis cs:0.408,0.16);
    \draw[obsline] (axis cs:0.413,0.0) -- (axis cs:0.413,0.16);
    \draw[obsline] (axis cs:0.418,0.0) -- (axis cs:0.418,0.16);
    \draw[obsline] (axis cs:0.423,0.0) -- (axis cs:0.423,0.16);
    \draw[obsline] (axis cs:0.428,0.0) -- (axis cs:0.428,0.16);
    \draw[obsline] (axis cs:0.433,0.0) -- (axis cs:0.433,0.16);
    \draw[obsline] (axis cs:0.438,0.0) -- (axis cs:0.438,0.16);
    \draw[obsline] (axis cs:0.443,0.0) -- (axis cs:0.443,0.16);

    \draw[-{Stealth[length=5pt,width=6pt]}, color=trainingtesting, line width=3.0pt]
  (axis cs:0.38,0.01) -- (axis cs:0.385,0.01);
    \node[font=\small\bfseries, text=trainingtesting, anchor=west]
  at (axis cs:0.3798,0.02) {Testing Period};

    \end{axis}

    \node[anchor=north west, font=\large\bfseries, xshift=5pt, yshift=-5pt] at (zoomplot.north west) {(b)};

    \draw[densely dotted, color=black, line width=0.9pt] (zoomBoxSW) -- (zoomplot.north west);
    \draw[densely dotted, color=black, line width=0.9pt] (zoomBoxSE) -- (zoomplot.north east);
\end{tikzpicture}
    \caption{Relative $L_2$ prediction error for the Burgers' equation across
training ($t\leq0.24$) and testing ($t>0.24$).
(a) Comparison of the autonomous OpInf forecast, correction-windowed
assimilation using full-state observations, and assimilation using
$m=4$ CPQR-selected sensors.
(b) Enlarged view of a representative testing interval showing observation
times and an activated correction window. The state correction reduces the
accumulated forecast error before the ROM resumes forward propagation.}
    \label{fig:burger_error_timeseries}
\end{figure}

Figure~\ref{fig:burgers_dashboard} shows the spatial evolution of the Burgers' solution after the ROM
enters the post-training regime. The autonomous OpInf forecast progressively
departs from the reference solution as the shock-like structure develops,
whereas correction-windowed assimilation repeatedly re-aligns the reduced
trajectory with the available observations. Importantly, the four-sensor
configuration produces a corrected field that remains qualitatively close to
the result obtained using observations at all $128$ spatial locations.

Figure~\ref{fig:burger_error_timeseries} quantifies this behavior using the relative $L_2$ prediction error.
The autonomous ROM begins with small error near the training boundary but
accumulates substantial error as the post-shock trajectory evolves. Both
assimilation configurations limit this growth through repeated state
corrections, with the CPQR-selected four-sensor case closely following the
full-state reference over much of the testing interval. The enlarged view of a
representative correction event illustrates the mechanism directly: error
grows between observations, a finite correction window is activated, and the
updated state reduces the accumulated forecast discrepancy before propagation
resumes.

The OpInf regularisation parameters are selected from the validation trajectory
as described above and are held fixed throughout the subsequent assimilation
experiments.

The post-shock regime also clarifies the role of the local correction window.
The ROM is constructed using data through $t=0.24$, up to the onset of shock
formation, after which the POD basis and learned OpInf operators are held
fixed. As sharper spatial gradients develop during the subsequent evolution,
these features are represented less accurately by the fixed reduced model,
leading to increasing forward-model error and tangent-linear-model mismatch.
Over a long assimilation horizon, these discrepancies can propagate through
both the forward trajectory and the reduced adjoint and thereby degrade the
fidelity of the correction gradient. The analysis in Section~\ref{sec:bounds_and_stability} shows that
restricting the correction horizon limits the amplification of these error
terms and therefore promotes satisfaction of the sufficient descent condition
derived in Corollary~\ref{cor:descent}. Windowing does not eliminate projection or model-form
error; rather, it restricts how strongly these errors contaminate each local
variational update. 

Finally, we observe that the projected full-order and ROM gradients also exhibit strong alignment for the Burgers problem, with \(\cos\theta=0.996651\) (\(\theta=4.69^\circ\)). Since \(\theta<\pi/2\), the reduced-order gradient retains a descent direction for the full-order objective restricted to the POD subspace, consistent with the preceding analysis in Section~\ref{sec:bounds_and_stability}. 

Having established that intermittent state correction limits post-training
forecast drift, we next examine the sensitivity of the Burgers' results to the
principal assimilation parameters: observation frequency, trigger threshold,
and correction-window length. The OpInf model and the CPQR-selected sensor
locations are held fixed throughout this study, so that the analysis isolates
the influence of the online assimilation settings.

\subsubsection{Assimilation Parameter Sensitivity}
\label{subsubsec:param_opt}

We examine the sensitivity of the correction-windowed assimilation to the
three online parameters introduced in Section~III.A.1: observation frequency
$f_{\mathrm{obs}}$, trigger threshold $\tau$, and correction-window length
$N_{\mathrm{window}}$. Throughout this study, the learned OpInf model and the
CPQR-selected sensor locations are held fixed. Parameter combinations are
compared using the cumulative observation-space discrepancy
$\sum_{k=0}^{T}\|\mathbf{y}_k-\mathbf{H}\widehat{\mathbf{x}}_k\|_F$
over the testing interval.

Within the tested parameter sweep, the lowest cumulative error is obtained for
observations every five time steps ($f_{\mathrm{obs}}=1/5$), a relative-error
trigger threshold of $\tau=0.1$, and a correction-window length of
$N_{\mathrm{window}}=7$. These values are used for the Burgers' experiment
and should be interpreted as problem-specific settings identified within the
tested range rather than as universally optimal assimilation parameters.

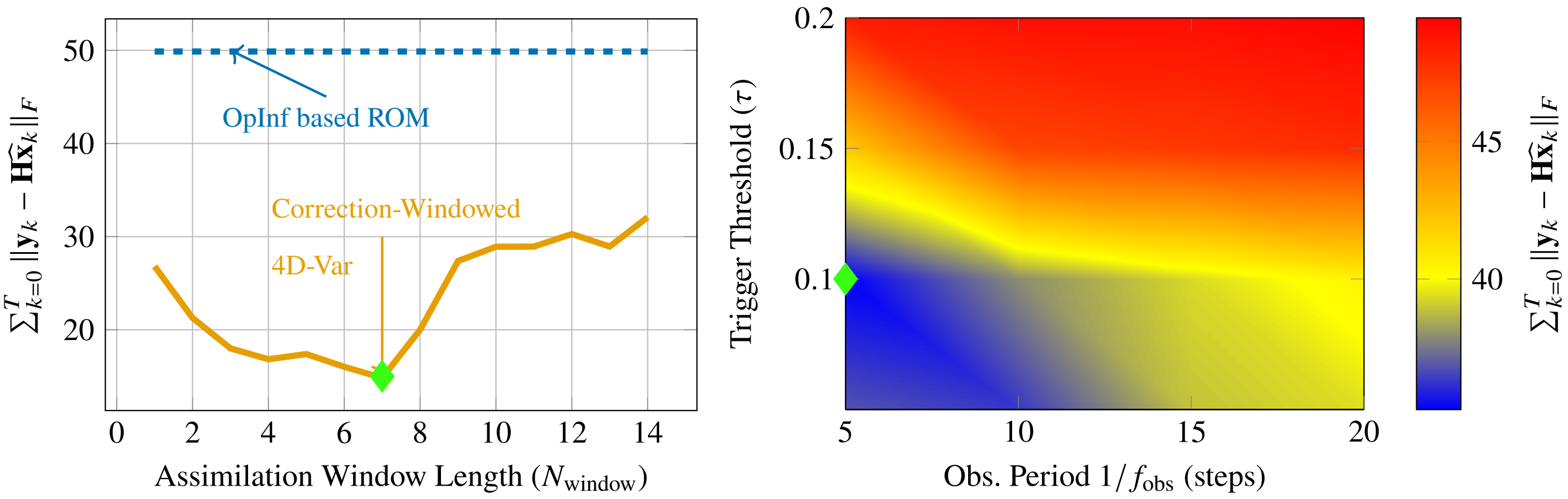
\begin{figure}[H]
    \hspace{-1.5em}
    \begin{subfigure}[t]{0.50\linewidth}
        \centering
        \begin{tikzpicture}
            \begin{axis}[
                width=\linewidth, height=6cm,
                xlabel={Assimilation Window Length ($N_{\text{window}}$)},
                ylabel={$\sum_{k=0}^{T} \| \mathbf{y}_k - \mathbf{H}\widehat{\mathbf{x}}_k \|_F$},
                grid=major,
                mark size=2pt
            ]
            \addplot[romForecastColor, line width= 2.0pt, dashed] table [x=assim_window, y=err_rom, col sep=comma] {Data_sheets/sweep_assim_window.csv};

            \addplot[assimColor, line width= 2.0pt, solid] table [x=assim_window, y=err_win, col sep=comma] {Data_sheets/sweep_assim_window.csv};
            
            \definecolor{neonGreen}{HTML}{39FF14}
            \addplot[only marks, mark=diamond*, mark size=5pt, neonGreen] coordinates {(7, 15.0)};
            \node[font=\small, text=romForecastColor, anchor=west, align=left] (romlbl) at (axis cs:2.5,42.5) {OpInf based ROM};
            \draw[->, romForecastColor, thick] (romlbl.north) -- (axis cs:3.0,50.0);

            \node[font=\small, text=assimColor, anchor=west, align=left] (winlbl) at (axis cs:3.8,30.0) {Correction-Windowed \\ 4D-Var};
            \draw[->, assimColor, thick] (axis cs:7.0,30.0) -- (axis cs:7.0,15.0);
            \end{axis}
        \end{tikzpicture}
        \caption{Cumulative error vs. window length}
        \label{fig:sweep_window}
    \end{subfigure}
    \begin{subfigure}[t]{0.5\linewidth}
        \centering
        \begin{tikzpicture}
            \begin{axis}[
                view={0}{90},
                width=0.9\linewidth, height=6cm,
                xlabel={Obs.\ Period $1/f_{\text{obs}}$ (steps)},
                ylabel={Trigger Threshold ($\tau$)},
                colorbar,
                colorbar style={ylabel={$\sum_{k=0}^{T} \| \mathbf{y}_k - \mathbf{H}\widehat{\mathbf{x}}_k \|_F$}},
                enlargelimits=false,
                axis on top,
                ytick = {0.1, 0.15, 0.2},
            ]
            \addplot3[
                surf, shader=interp,
                mesh/cols=4
            ] table [x=obs_interval, y=trigger_threshold, z=windowed_error, col sep=comma] {Data_sheets/heatmap_obs_trigger.csv};
            \definecolor{neonGreen}{HTML}{39FF14}
            \addplot3[only marks, mark=diamond*, mark size=5pt, neonGreen] coordinates {(5, 0.1, 33)};
            
            \end{axis}
        \end{tikzpicture}
        \caption{Error heatmap for observation and trigger parameters}
        \label{fig:heatmap_interactions}
    \end{subfigure}
    \definecolor{neonGreen}{HTML}{39FF14}
    \caption{Sensitivity of the Burgers' assimilation results to the online
correction parameters.
(a) Cumulative observation-space discrepancy as a function of correction-window
length $N_{\mathrm{window}}$; the lowest value among the tested configurations
occurs at $N_{\mathrm{window}}=7$.
(b) Dependence of the cumulative discrepancy on observation period and trigger
threshold $\tau$. The \textcolor{neonGreen}{Neon Green} marker denotes the configuration
$f_{\mathrm{obs}}=1/5$ and $\tau=0.1$ used in the Burgers' experiments.
The selected parameters represent the best-performing settings within the
tested sweep and are not intended as universal optima.}
    \label{fig:parameter_optimization}
\end{figure}

Figure~\ref{fig:sweep_window} examines the influence of correction-window
length. All of the tested correction-window configurations substantially
improve upon the open-loop OpInf-based ROM. The cumulative discrepancy decreases as the
window is increased up to $N_{\mathrm{window}}=7$, after which it increases
again for the longer windows considered. Thus, the best-performing value
within this sweep is $N_{\mathrm{window}}=7$. This result indicates that,
for the present Burgers' problem, extending the correction horizon beyond a
certain range does not provide additional benefit and can instead degrade the
quality of the correction. This behavior is consistent with the analysis in
Section~\ref{sec:bounds_and_stability}, which shows that increasing the physical correction horizon
allows greater amplification of forward-model and adjoint errors. The theory,
however, does not predict a particular optimal value of
$N_{\mathrm{window}}$.

Figure~\ref{fig:heatmap_interactions} shows the corresponding sensitivity to
observation availability and the trigger threshold. As expected, infrequent
observations combined with a permissive trigger allow larger discrepancies to
develop between corrective updates. More frequent observations provide
additional opportunities for correction, while the threshold determines
whether the available observation actually activates a 4D-Var solve. For the
present Burgers' configuration, $f_{\mathrm{obs}}=1/5$ and $\tau=0.1$
provide low cumulative error without requiring the most aggressive trigger
setting considered in the sweep.

Taken together, these results show that the proposed framework does not depend
on a single precisely tuned correction horizon. Instead, effective performance
is obtained over a range of local assimilation settings, with the selected
configuration providing a representative operating point for the Burgers'
experiment. The dominant result remains that intermittent corrections prevent
the extrapolating fixed ROM from accumulating the large errors observed in the
open-loop OpInf-based forecast.
\subsection{Sea Surface Temperature}
\label{sec:sst}
The SST case extends the proposed framework to a realistic high-dimensional
geophysical system with a state dimension of $n=33{,}528$. Unlike the
controlled Burgers' experiment, this example tests whether intermittent state
correction can preserve the predictive utility of a fixed learned ROM when the
underlying state is large and observational coverage is necessarily limited.
Such settings are representative of geophysical forecasting problems, where
high-dimensional state estimation must be performed from incomplete
measurements.

We use the CMC $0.1^{\circ}$ Global Foundation SST Level-4 Analysis dataset,
a daily gap-filled product constructed from satellite observations. The spatial
domain covers the Indian Ocean region spanning $5^{\circ}$N--$22^{\circ}$N
and $75^{\circ}$E--$95^{\circ}$E, including the Bay of Bengal and Arabian Sea.
After excluding land points, each SST snapshot contains $n=33{,}528$ valid
ocean grid points. The data are sampled daily from January 1 to February 28,
2026, yielding a 59-day trajectory. Temperatures are converted to $^{\circ}$C
for visualization. Before constructing the POD snapshot matrix and training
the OpInf operators, the SST values used by the model are min--max normalized
to the interval $[0,1]$. The same scaling is applied to the validation,
testing, and assimilated observations, while reported temperatures and figures
are converted back to $^{\circ}$C.

\begin{figure}[H]
    \centering
    \begin{subfigure}[b]{0.35\textwidth}
        \centering
        \input{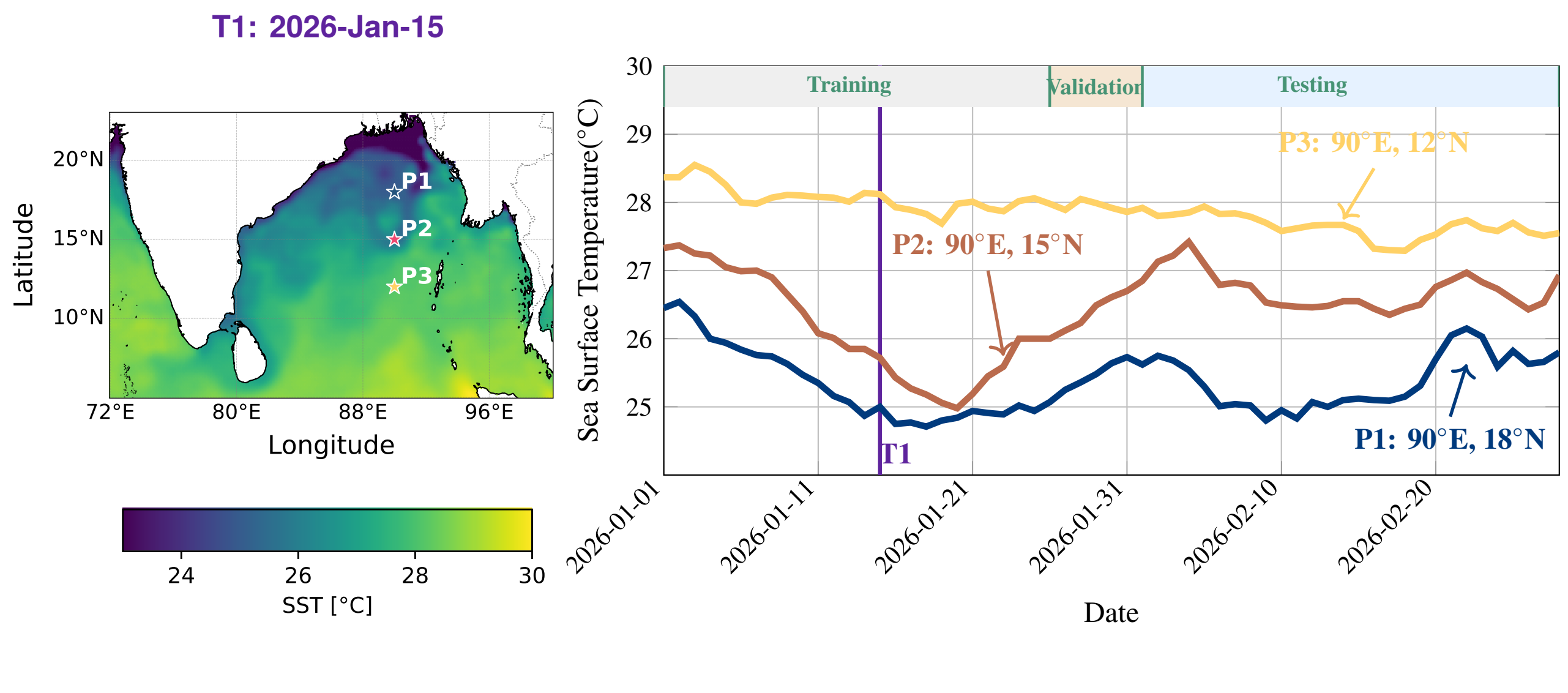}
        \caption{Spatial field of SST at reference time T1. }
        \label{fig:sst_truth_plots}
    \end{subfigure}
    \begin{subfigure}[b]{0.62\textwidth}
        \centering
\begin{tikzpicture}
\def\trainStart{2026-01-01}
\def\trainEnd{2026-01-26}
\def\trainMid{2026-01-13}
\def\validStart{2026-01-26}
\def\validEnd{2026-02-01}
\def\validMid{2026-01-29}
\def\testStart{2026-02-01}
\def\testEnd{2026-02-28}
\def\testMid{2026-02-12}
\def\boxstarts{29.4}
\def\boxends{30}
\def\boxmids{29.7}
\def\boxstart{29.4}
\def\boxend{30}
\def\boxmid{29.7}

\begin{axis}[
    name=mainplot,
    width=1.1\textwidth,
    height=6cm,
    grid=both,
    xlabel={Date},
    ylabel={Sea Surface Temperature($^\circ$C)},
    ymin=24,
    ytick={25, 26, 27, 28, 29, 30},
    ymax=30,
    date coordinates in=x,
    xticklabel=\year-\month-\day,
    xticklabel style={rotate=45, anchor=east, font=\footnotesize},
    tick label style={font=\footnotesize},
    label style={font=\small},
    legend style={at={(0.5,-0.22)}, anchor=north, legend columns=1, draw=none},
    xtick distance = 10,
    xmin = 2026-01-01,
    xmax = 2026-02-28,
]


\draw[timestamps, line width=1.2pt] (axis cs:2026-01-15,0) -- (axis cs:2026-01-15,50);
\node[anchor=south, font=\small\bfseries, color=timestamps] at (axis cs:2026-01-16,24.0) {T1};

\path[fill=phaseboxgray]
    (axis cs:\trainStart,\boxstarts) rectangle (axis cs:\trainEnd,\boxends);
\draw[draw=trainingtesting, line width=0.8pt] 
    (axis cs:\trainStart,\boxstarts) -- (axis cs:\trainStart,\boxends)
    (axis cs:\trainEnd,\boxstarts) -- (axis cs:\trainEnd,\boxends);
\node[font=\scriptsize\bfseries, text=trainingtesting]
    at (axis cs:\trainMid,\boxmids) {Training};

\path[fill=phaseboxxbrown]
    (axis cs:\validStart,\boxstarts) rectangle (axis cs:\validEnd,\boxends);
\draw[draw=trainingtesting, line width=0.8pt]
    (axis cs:\validStart,\boxstarts) -- (axis cs:\validStart,\boxends)
    (axis cs:\validEnd,\boxstarts) -- (axis cs:\validEnd,\boxends);
\node[font=\scriptsize\bfseries, text=trainingtesting]
    at (axis cs:\validMid,\boxmids) {Validation};

\path[fill=obslabelblue]
    (axis cs:\testStart,\boxstart) rectangle (axis cs:\testEnd,\boxend);
\draw[draw=trainingtesting, line width=0.8pt]
    (axis cs:\testStart,\boxstart) -- (axis cs:\testStart,\boxend)
    (axis cs:\testEnd,\boxstart) -- (axis cs:\testEnd,\boxend);
\node[font=\scriptsize\bfseries, text=trainingtesting]
    at (axis cs:\testMid,\boxmid) {Testing};

\addplot[P1Color, line width=2.0pt] table [x=time, y=temp_lon90_lat18, col sep=comma] {Data_sheets/truth_temperature_timeseries_90E_18N_15N_12N.csv};
\node[anchor=west, color=P1Color, font=\small\bfseries] (label18) at (axis cs: 2026-02-14, 24.50) {P1: 90$^\circ$E, 18$^\circ$N};
\draw[->, color=P1Color, line width=1pt] (label18.north) -- (axis cs: 2026-02-22, 25.625);

\addplot[P2Color, line width=2.0pt] table [x=time, y=temp_lon90_lat15, col sep=comma] {Data_sheets/truth_temperature_timeseries_90E_18N_15N_12N.csv};
\node[anchor=south, color=P2Color, font=\small\bfseries, align=center] (label15) at (axis cs: 2026-01-22, 27.0) {P2: 90$^\circ$E, 15$^\circ$N};
\draw[->, color=P2Color, line width=1pt] (label15.south) -- (axis cs: 2026-01-23, 25.75);

\addplot[P3Color, line width=2.0pt] table [x=time, y=temp_lon90_lat12, col sep=comma] {Data_sheets/truth_temperature_timeseries_90E_18N_15N_12N.csv};
\node[anchor=south, color=P3Color, font=\small\bfseries, align=center] (label12) at (axis cs: 2026-02-16, 28.50) {P3: 90$^\circ$E, 12$^\circ$N};
\draw[->, color=P3Color, line width=1pt] (label12.south) -- (axis cs: 2026-02-14, 27.75);

\end{axis}
\end{tikzpicture}
        \caption{Temporal evolution of the SST at the designated probe locations.  }
        \label{fig:sst_true_evolution}
    \end{subfigure}
    \caption{Spatial and temporal characteristics of the SST dataset used in the assimilation experiment. 
(a) Representative SST field at reference time $T_1$ with probe locations P1, P2, and P3 along the $90^\circ$E meridian. 
(b) SST evolution at the three probe locations over the full study interval. 
The shaded regions indicate the training, validation, and testing periods.}
    \label{fig:sst_true_combined}
\end{figure}

We use this contiguous two-month interval to retain a reasonably coherent
seasonal setting while still providing a sufficiently long temporal separation
between model construction and testing. Figure~\ref{fig:sst_true_combined}
provides spatial and temporal context for the experiment. The spatial snapshot
shows a representative SST field together with three probe locations along the
$90^{\circ}$E meridian: P1 at $18^{\circ}$N, P2 at $15^{\circ}$N, and P3 at
$12^{\circ}$N. These locations sample the north--south temperature variation
along a common longitude and are subsequently used to compare the temporal
behavior of the different forecast and assimilation strategies.

The 59-day trajectory is partitioned into training (January 1--27),
validation (January 28--31), and testing (February 1--28), as indicated in
Figure~\ref{fig:sst_true_evolution}. A POD basis of rank $r=12$ is constructed
from the training snapshots, with the retained modes capturing approximately
$99.9\%$ of the cumulative snapshot energy. The corresponding linear and
quadratic OpInf operators are learned from the same training interval, while
the four-day validation period is used to select the Tikhonov regularisation
parameters.

\ak{For $r=12$, the linear--quadratic OpInf model contains
$r+r(r+1)/2=90$ regressors per reduced equation, so the SST regression
is underdetermined relative to the available training trajectory. We
deliberately retain this data-limited setting to evaluate OpInf in a
regime representative of realistic applications, where long,
high-quality training trajectories may not be available. Accordingly,
we do not interpret the inferred operators as uniquely identified
physical coefficients. Instead, Eq.~\eqref{eq:opinf_opt} is solved as a
Tikhonov-regularized predictive regression, with the regularization
parameters selected using validation-trajectory error. The resulting
ROM is then held fixed throughout the testing period and treated as a
predictive surrogate. The limited training interval also preserves a
temporally separated out-of-sample test period, allowing us to assess
whether the proposed data-assimilation framework can extend the
predictive utility of the fixed ROM beyond the interval used for its
construction.}

\ak{For the SST assimilation experiments, observations are available at the daily sampling interval of the dataset. We tested multiple correction-window lengths and observed broadly similar forecast performance across the short horizons considered. The best performance was obtained using a one-day correction horizon, in which an observation at the current day is used to correct the reduced state one day prior before the trajectory is re-integrated forward. We therefore use this one-day correction horizon for the SST results reported below.} \ak{No additional synthetic observation noise is added in the SST experiments; the Level-4 SST fields are assimilated as provided, with observational uncertainty represented through the observation-error covariance $\mathbf{R}=\sigma_{\text{obs}}^2\mathbf{I}$, with $\sigma_{\text{obs}} = 10^{-2}$ and the background covariance used was $\widehat{\mathbf{B}} = \mathbf{I}_r$ (As per assumption 2)}. \ak{For the SST experiments, the same relative observation-space trigger criterion defined in Eq.~\eqref{eqn:rel_error} is used, with a trigger threshold of $\tau = 0.5$.}

After validation, the POD basis and learned OpInf operators are held fixed
throughout the February testing interval. Thus, the ROM is required to forecast
state evolution outside the temporal interval used for its construction
without retraining the operators or updating the reduced basis. This provides
a substantially higher-dimensional test of the same central question examined
with Burgers' equation: whether subsequently acquired observations can be used
to correct the evolving reduced state and extend the useful predictive horizon
of an otherwise deteriorating fixed ROM.

Once the fixed ROM is propagated into the February testing interval, the
open-loop OpInf forecast progressively departs from the reference SST evolution.
This provides a high-dimensional realization of the same extrapolation problem
examined with Burgers' equation: although the learned ROM remains inexpensive
to propagate, its long-horizon predictive accuracy deteriorates outside the
temporal interval used for model construction. The question is therefore
whether observations acquired during testing can be used to correct the
evolving reduced state while leaving the POD basis and learned OpInf operators
unchanged.

We compare three forecasting configurations during the testing period:
the open-loop OpInf ROM, reduced-order assimilation using
$m=r=12$ CPQR-selected measurement locations, and reduced-order assimilation
using observations at all $n=33{,}528$ spatial locations as a full-state
reference. The CPQR-selected locations are obtained using the procedure in
Section~\ref{sec:sensor_placement}. The same fixed POD basis and OpInf
operators are used in all three cases; only the observational information
available to the assimilation procedure differs. This comparison isolates
whether sparse measurements that constrain the reduced state can recover much
of the forecast improvement obtained when observations are available across
the full spatial domain.

Drift-correction performance is quantified across three scenarios: the
open-loop OpInf ROM, sparse-sensor assimilation, and full-state assimilation.
Table~\ref{tab:sst_metrics} reports the mean absolute error (MAE), relative
$L_2$ error, and the Frobenius norm of the prediction error over the testing
period. These metrics provide complementary measures of average forecast
error, relative state error, and aggregate spatiotemporal discrepancy,
respectively. The Efficiency column measures how much of the error reduction
obtained using full-state observations is recovered by the sparse
configuration.

\begin{table}[H]
\centering
\small
\begin{tabular}{lllll}
\hline
\textbf{Metric} & \textbf{ROM} & \textbf{Assimilated (Sparse, $m$=12)} & \textbf{Assimilated (Full-State)} & \textbf{Efficiency} \\
\hline
Mean Absolute Error (MAE) & 2.8$\unit{\degreeCelsius}$ & 1.2$\unit{\degreeCelsius}$ & 1.1$\unit{\degreeCelsius}$ & \ak{$94.12\%$} \\
Relative $L_2$ (\%) & 3.2 & 1.2& 1.0 & \ak{$90.91\%$} \\
Frobenius norm of error (full domain) & 2847$\unit{\degreeCelsius}$ & 658$\unit{\degreeCelsius}$ & 587$\unit{\degreeCelsius}$ & \ak{$96.86\%$} \\
\hline
\end{tabular}
\caption{SST forecast-error metrics over the 28-day testing period for the
unassimilated ROM, sparse-sensor assimilation with $m=12$ CPQR-selected
measurements, and full-state assimilation. Efficiency measures the fraction
of the error reduction obtained with full-state observations that is recovered
using sparse measurements.}
\label{tab:sst_metrics}
\end{table}
\vspace{-30pt}
To quantify how much of the improvement obtained with full-state observations
is recovered using sparse sensing, we define the efficiency as
$100(E_{\mathrm{ROM}}-E_{\mathrm{sparse}})
 /(E_{\mathrm{ROM}}-E_{\mathrm{full}})$. A particularly notable result is the degree of observational sparsity required
for the reduced-order correction. The sparse configuration supplies
measurements at only $12$ of the $33{,}528$ spatial locations, corresponding
to approximately $0.036\%$ of the full-state measurement density, or a
$99.964\%$ reduction in measurement locations relative to the full-state
reference. Despite this substantial reduction, the CPQR-selected measurements
recover most of the error reduction obtained when observations are available
over the complete spatial domain. This result reflects an important advantage of performing the correction in
the $r=12$ dimensional reduced space: measurements need primarily to constrain
the retained reduced coordinates rather than directly observe every component
of the high-dimensional state.

 Across all three error measures, sparse assimilation substantially improves
upon the open-loop ROM and approaches the performance obtained with full-state
observations. The MAE decreases from $2.8^\circ$C for the open-loop OpInf ROM to
$1.2^\circ$C with the 12-location configuration, compared with
$1.1^\circ$C using full-state observations. More broadly, the sparse
configuration recovers approximately $91$--$97\%$ of the error reduction
achieved by the full-state reference across the three reported metrics.
Thus, for this SST experiment, the principal benefit of assimilation is
retained despite the large reduction in spatial measurement density.
\begin{figure}[!t]
  \centering
      \input{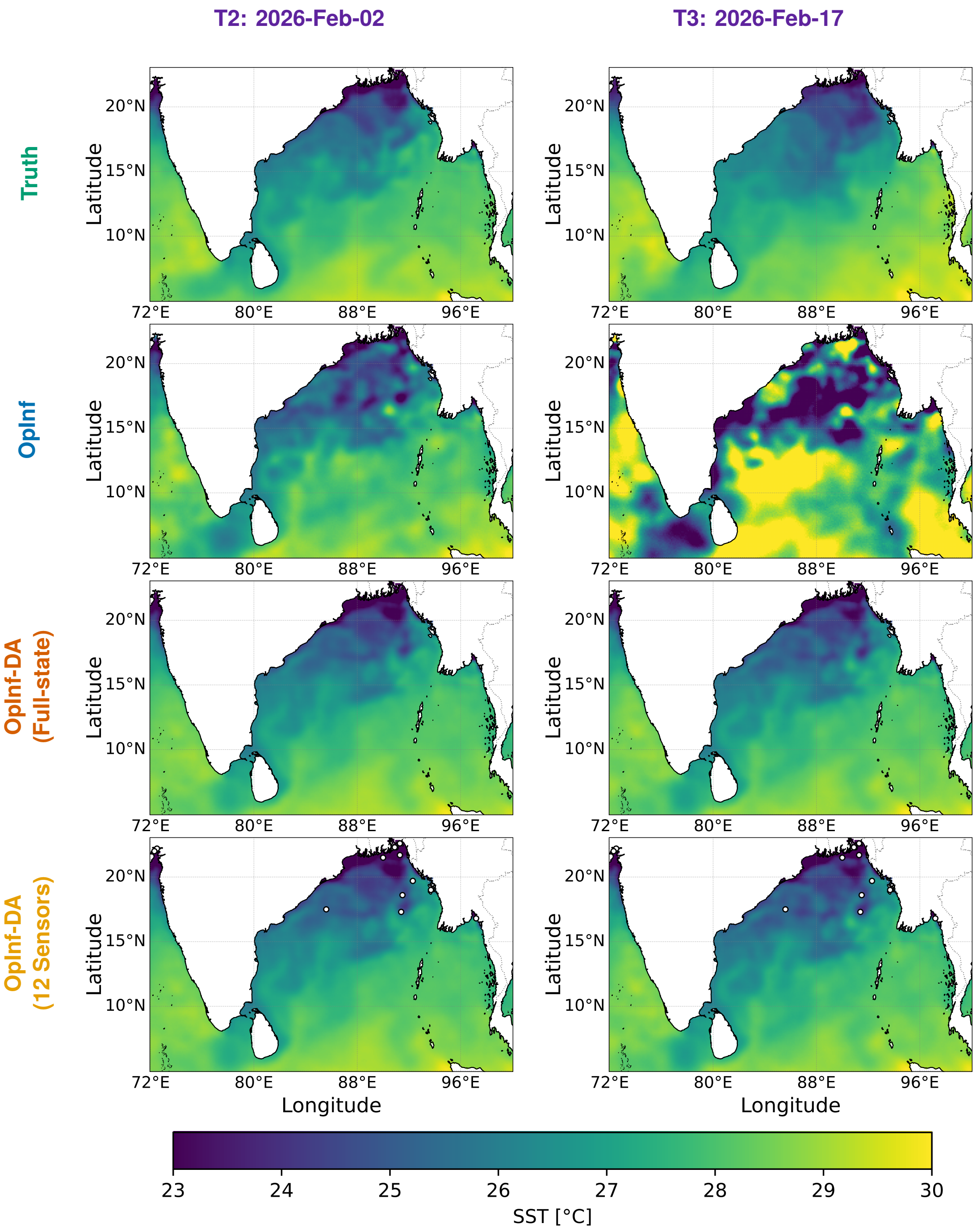}
       \caption{Spatial SST fields at two representative times, $T_2$ and $T_3$,
during the February testing period. The rows show the reference SST field,
the open-loop OpInf forecast, correction using full-state observations, and
correction using $m=12$ CPQR-selected measurement locations, respectively.
The selected measurement locations are indicated by white circles in the
sparse-assimilation panels. Both assimilation configurations recover the
large-scale spatial temperature structure more accurately than the open-loop
ROM forecast.}
  \label{fig:sst_dashboard}
\end{figure}

Figure~\ref{fig:sst_dashboard} provides the corresponding spatial view of
forecast performance at two representative times in the testing interval.
As the forecast horizon increases, the open-loop OpInf prediction progressively
departs from the reference SST field and exhibits a visibly smoother spatial
structure. Both assimilation configurations substantially improve the spatial
agreement with the reference field, recovering the dominant temperature
gradients and basin-scale structure more accurately than the open-loop ROM.
Despite using measurements at only $12$ spatial locations, the sparse
assimilation produces fields that remain qualitatively close to those obtained
with full-state observations. Residual discrepancies are more apparent at the
later time $T_3$, consistent with the increasing difficulty of correcting the
trajectory as the fixed ROM moves farther beyond the interval used for its
construction.
\begin{figure}[H]
    \centering
\begin{tikzpicture}
\def\trainStart{2026-01-01}
\def\trainEnd{2026-01-26}
\def\trainMid{2026-01-13}
\def\validStart{2026-01-26}
\def\validEnd{2026-02-01}
\def\validMid{2026-01-29}
\def\testStart{2026-02-01}
\def\testEnd{2026-02-28}
\def\testMid{2026-02-12}
\def\boxstarts{29.4}
\def\boxends{30}
\def\boxmids{29.7}
\def\boxstart{29.4}
\def\boxend{30}
\def\boxmid{29.7}

\begin{axis}[
    name=mainplot,
    width=0.98\linewidth,
    height=7cm,
    grid=both,
    xlabel={Date},
    ylabel={Sea Surface Temperature($^\circ$C)},
    ymin=24,
    ymax=30,
    date coordinates in=x,
    xticklabel=\year-\month-\day,
    xticklabel style={rotate=45, anchor=east, font=\footnotesize},
    tick label style={font=\footnotesize},
    label style={font=\small},
    legend style={at={(0.5,-0.28)}, anchor=north, legend columns=4, draw=none, font=\footnotesize},
    xtick distance = 10,
    xmin= 2026-01-01,
    xmax= 2026-02-28,
    ytick = {26, 28, 30},
]

\draw[timestamps, line width=1.2pt] (axis cs:2026-02-02,0) -- (axis cs:2026-02-02,50);
\node[anchor=south, font=\small\bfseries, color=timestamps] at (axis cs:2026-02-03,28.20) {T2};

\draw[timestamps, line width=1.2pt] (axis cs:2026-02-17,0) -- (axis cs:2026-02-17,50);
\node[anchor=south, font=\small\bfseries, color=timestamps] at (axis cs:2026-02-18,28.20) {T3};

\path[fill=phaseboxgray]
    (axis cs:\trainStart,\boxstarts) rectangle (axis cs:\trainEnd,\boxends);
\draw[draw=trainingtesting, line width=0.8pt] 
    (axis cs:\trainStart,\boxstarts) -- (axis cs:\trainStart,\boxends)
    (axis cs:\trainEnd,\boxstarts) -- (axis cs:\trainEnd,\boxends);
\node[font=\scriptsize\bfseries, text=trainingtesting]
    at (axis cs:\trainMid,\boxmids) {Training};

\path[fill=phaseboxxbrown]
    (axis cs:\validStart,\boxstarts) rectangle (axis cs:\validEnd,\boxends);
\draw[draw=trainingtesting, line width=0.8pt]
    (axis cs:\validStart,\boxstarts) -- (axis cs:\validStart,\boxends)
    (axis cs:\validEnd,\boxstarts) -- (axis cs:\validEnd,\boxends);
\node[font=\scriptsize\bfseries, text=trainingtesting]
    at (axis cs:\validMid,\boxmids) {Validation};

\path[fill=obslabelblue]
    (axis cs:\testStart,\boxstart) rectangle (axis cs:\testEnd,\boxend);
\draw[draw=trainingtesting, line width=0.8pt]
    (axis cs:\testStart,\boxstart) -- (axis cs:\testStart,\boxend)
    (axis cs:\testEnd,\boxstart) -- (axis cs:\testEnd,\boxend);
\node[font=\scriptsize\bfseries, text=trainingtesting]
    at (axis cs:\testMid,\boxmid) {Testing};

\addplot[P1Color, solid, mark=*, mark options={solid, scale=0.7}, line width=1.2pt, forget plot] table [x=time, y=truth_lon90_lat18, col sep=comma, each nth point=12] {Data_sheets/combined_truth_rom_fullstate_sensor_timeseries_90E_18N_15N_12N.csv};
\addplot[P1Color, dashed, mark=square*, mark options={solid, scale=0.7}, line width=1.2pt, forget plot] table [x=time, y=rom_lon90_lat18, col sep=comma, each nth point=12] {Data_sheets/combined_truth_rom_fullstate_sensor_timeseries_90E_18N_15N_12N.csv};
\addplot[P1Color, dotted, mark=triangle*, mark options={solid, scale=0.8}, line width=1.2pt, forget plot] table [x=time, y=full_assim_lon90_lat18, col sep=comma, each nth point=12] {Data_sheets/combined_truth_rom_fullstate_sensor_timeseries_90E_18N_15N_12N.csv};
\addplot[P1Color, dashdotted, mark=diamond*, mark options={solid, scale=0.8}, line width=1.2pt, forget plot] table [x=time, y=sensor_assim_lon90_lat18, col sep=comma, each nth point=12] {Data_sheets/combined_truth_rom_fullstate_sensor_timeseries_90E_18N_15N_12N.csv};

\addplot[P2Color, solid, mark=*, mark options={solid, scale=0.7}, line width=1.2pt, forget plot] table [x=time, y=truth_lon90_lat15, col sep=comma, each nth point=12] {Data_sheets/combined_truth_rom_fullstate_sensor_timeseries_90E_18N_15N_12N.csv};
\addplot[P2Color, dashed, mark=square*, mark options={solid, scale=0.7}, line width=1.2pt, forget plot] table [x=time, y=rom_lon90_lat15, col sep=comma, each nth point=12] {Data_sheets/combined_truth_rom_fullstate_sensor_timeseries_90E_18N_15N_12N.csv};
\addplot[P2Color, dotted, mark=triangle*, mark options={solid, scale=0.8}, line width=1.2pt, forget plot] table [x=time, y=full_assim_lon90_lat15, col sep=comma, each nth point=12] {Data_sheets/combined_truth_rom_fullstate_sensor_timeseries_90E_18N_15N_12N.csv};
\addplot[P2Color, dashdotted, mark=diamond*, mark options={solid, scale=0.8}, line width=1.2pt, forget plot] table [x=time, y=sensor_assim_lon90_lat15, col sep=comma, each nth point=12] {Data_sheets/combined_truth_rom_fullstate_sensor_timeseries_90E_18N_15N_12N.csv};

\addplot[P3Color, solid, mark=*, mark options={solid, scale=0.7}, line width=1.2pt, forget plot] table [x=time, y=truth_lon90_lat12, col sep=comma, each nth point=12] {Data_sheets/combined_truth_rom_fullstate_sensor_timeseries_90E_18N_15N_12N.csv};
\addplot[P3Color, dashed, mark=square*, mark options={solid, scale=0.7}, line width=1.2pt, forget plot] table [x=time, y=rom_lon90_lat12, col sep=comma, each nth point=12] {Data_sheets/combined_truth_rom_fullstate_sensor_timeseries_90E_18N_15N_12N.csv};
\addplot[P3Color, dotted, mark=triangle*, mark options={solid, scale=0.8}, line width=1.2pt, forget plot] table [x=time, y=full_assim_lon90_lat12, col sep=comma, each nth point=12] {Data_sheets/combined_truth_rom_fullstate_sensor_timeseries_90E_18N_15N_12N.csv};
\addplot[P3Color, dashdotted, mark=diamond*, mark options={solid, scale=0.8}, line width=1.2pt, forget plot] table [x=time, y=sensor_assim_lon90_lat12, col sep=comma, each nth point=12] {Data_sheets/combined_truth_rom_fullstate_sensor_timeseries_90E_18N_15N_12N.csv};

\addlegendimage{black, solid, mark=*, mark options={solid, scale=0.7}, line width=1.2pt} \addlegendentry{Truth}
\addlegendimage{black, dashed, mark=square*, mark options={solid, scale=0.7}, line width=1.2pt} \addlegendentry{OpInf}
\addlegendimage{black, dotted, mark=triangle*, mark options={solid, scale=0.8}, line width=1.2pt} \addlegendentry{OpInf-DA (Full-state)}
\addlegendimage{black, dashdotted, mark=diamond*, mark options={solid, scale=0.8}, line width=1.2pt} \addlegendentry{OpInf-DA (12 Sensors)}

\addlegendimage{P1Color, solid, line width=2.0pt} \addlegendentry{P1 (18$^\circ$N)}
\addlegendimage{P2Color, solid, line width=2.0pt} \addlegendentry{P2 (15$^\circ$N)}
\addlegendimage{P3Color, solid, line width=2.0pt} \addlegendentry{P3 (12$^\circ$N)}

\end{axis}
\end{tikzpicture}
    \caption{Temporal evolution of SST at the three probe locations along the
$90^{\circ}$E meridian: P1 ($18^{\circ}$N), P2 ($15^{\circ}$N), and
P3 ($12^{\circ}$N). The reference trajectories are compared with the
open-loop OpInf forecast, full-state assimilation, and assimilation using
$m=12$ CPQR-selected measurements. The shaded regions indicate the training,
validation, and testing intervals, while $T_2$ and $T_3$ correspond to the
spatial fields shown in Figure~\ref{fig:sst_dashboard}. During testing, the
corrected trajectories remain closer to the reference evolution than the
open-loop forecast over much of the prediction interval.}
    \label{fig:model_evolution_time_series}
\end{figure}
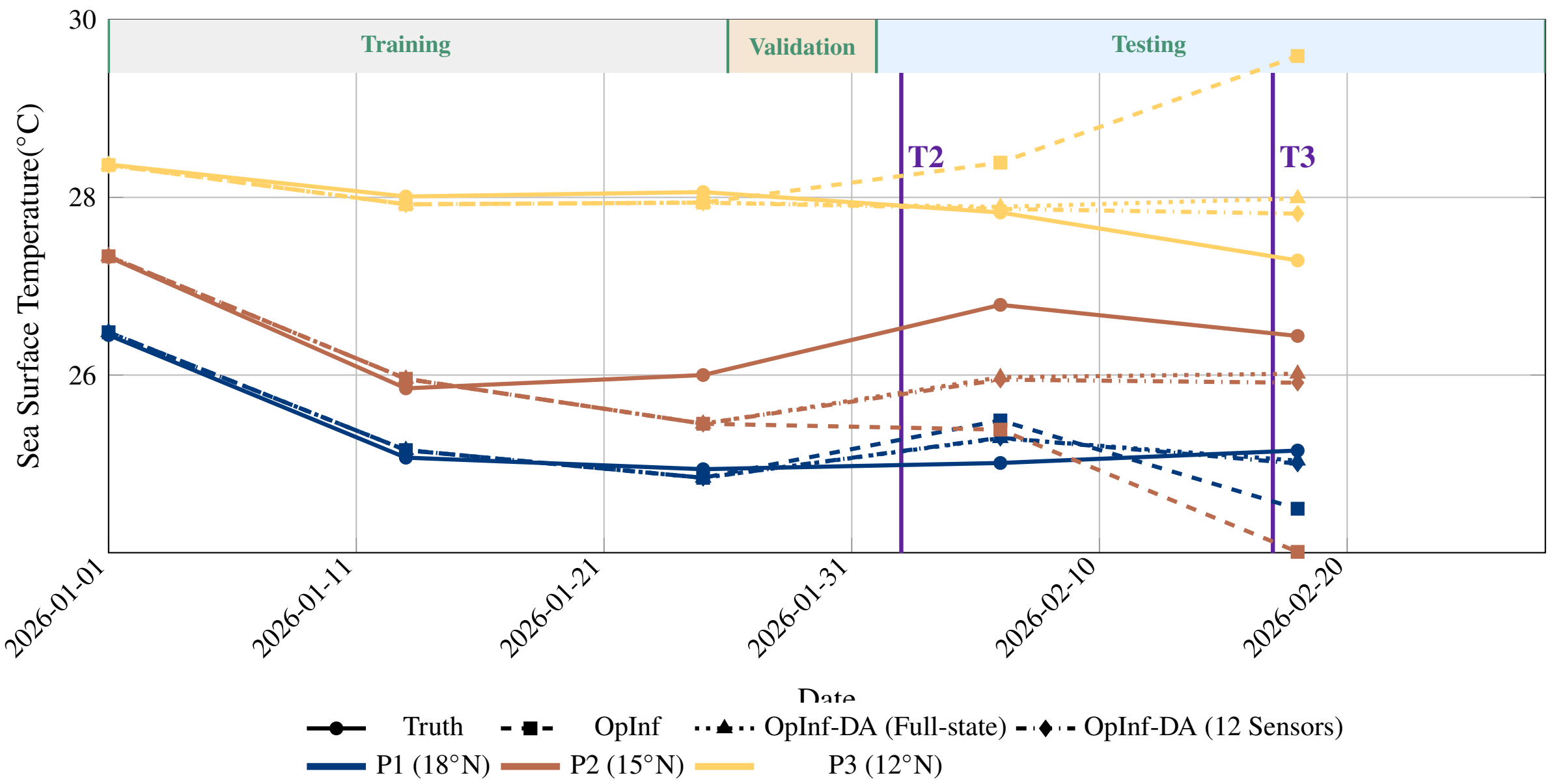

Figure~\ref{fig:model_evolution_time_series} provides a local temporal view of the
forecast behavior at the three probe locations. As the fixed OpInf model is
propagated beyond the interval used for its construction, the open-loop
forecast progressively departs from the reference SST evolution. In contrast,
both assimilation configurations remain substantially closer to the reference
trajectory over much of the testing interval. The sparse $m=12$ configuration
also follows the full-state assimilation result closely at these locations,
consistent with the aggregate error reductions reported in
Table~\ref{tab:sst_metrics}.

The corrected trajectories nevertheless begin to show increasing discrepancies
at later forecast times. This is an important limitation rather than a failure
of the correction strategy: state assimilation can repeatedly re-align the
trajectory, but it cannot recover dynamics that are increasingly poorly
represented by the fixed reduced basis and learned operators. The SST results
therefore reinforce the central distinction of this work---intermittent state
correction can extend the useful predictive horizon of a fixed ROM, but does
not make the reduced model indefinitely accurate outside its training regime.
The global evolution of this error is examined next.

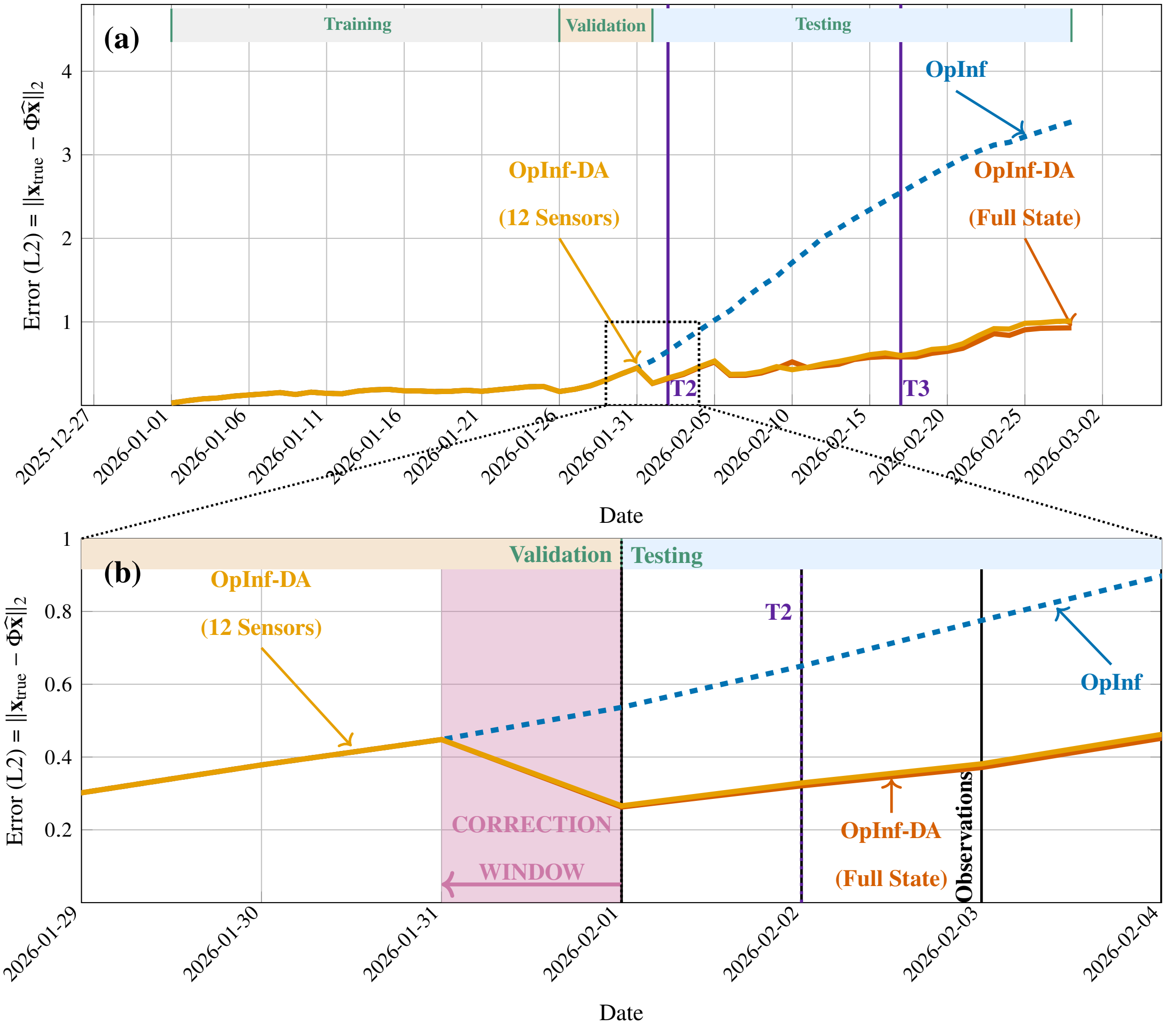
\begin{figure}[H]
  \centering
\begin{tikzpicture}
\def\trainStart{2026-01-01}
\def\trainEnd{2026-01-26}
\def\trainMid{2026-01-13}
\def\validStart{2026-01-26}
\def\validEnd{2026-02-01}
\def\validMid{2026-01-29}
\def\testStart{2026-02-01}
\def\testEnd{2026-02-28}
\def\testMid{2026-02-12}
\def\boxstarts{4.35}
\def\boxends{4.75}
\def\boxmids{4.55}
\def\boxstart{4.35}
\def\boxend{4.75}
\def\boxmid{4.55}

\begin{axis}[
    name=mainplot,
    width=0.98\linewidth,
    height=7cm,
    grid=both,
    xlabel={Date},
    ylabel={Error (L2) = $\| \mathbf{x}_{\text{true}} - \Phi \widehat{\mathbf{x}} \|_2$},
    ymin=0,
    ymax=4.8,
    date coordinates in=x,
    xticklabel=\year-\month-\day,
    xticklabel style={rotate=45, anchor=east, font=\footnotesize},
    tick label style={font=\footnotesize},
    label style={font=\small},
    legend style={at={(0.5,-0.22)}, anchor=north, legend columns=1, draw=none},
    ytick = {1, 2, 3, 4},
]

\draw[timestamps, line width=1.2pt] (axis cs:2026-02-02,0) -- (axis cs:2026-02-02,4.8);
\node[anchor=south, font=\small\bfseries, color=timestamps] at (axis cs:2026-02-03,0.0) {T2};

\draw[timestamps, line width=1.2pt] (axis cs:2026-02-17,0) -- (axis cs:2026-02-17,4.8);
\node[anchor=south, font=\small\bfseries, color=timestamps] at (axis cs:2026-02-18,0.0) {T3};

\path[fill=phaseboxgray]
    (axis cs:\trainStart,\boxstarts) rectangle (axis cs:\trainEnd,\boxends);
\draw[draw=trainingtesting, line width=0.8pt] 
    (axis cs:\trainStart,\boxstarts) -- (axis cs:\trainStart,\boxends)
    (axis cs:\trainEnd,\boxstarts) -- (axis cs:\trainEnd,\boxends);
\node[font=\scriptsize\bfseries, text=trainingtesting]
    at (axis cs:\trainMid,\boxmids) {Training};

\path[fill=phaseboxxbrown]
    (axis cs:\validStart,\boxstarts) rectangle (axis cs:\validEnd,\boxends);
\draw[draw=trainingtesting, line width=0.8pt]
    (axis cs:\validStart,\boxstarts) -- (axis cs:\validStart,\boxends)
    (axis cs:\validEnd,\boxstarts) -- (axis cs:\validEnd,\boxends);
\node[font=\scriptsize\bfseries, text=trainingtesting]
    at (axis cs:\validMid,\boxmids) {Validation};

\path[fill=obslabelblue]
    (axis cs:\testStart,\boxstart) rectangle (axis cs:\testEnd,\boxend);
\draw[draw=trainingtesting, line width=0.8pt]
    (axis cs:\testStart,\boxstart) -- (axis cs:\testStart,\boxend)
    (axis cs:\testEnd,\boxstart) -- (axis cs:\testEnd,\boxend);
\node[font=\scriptsize\bfseries, text=trainingtesting]
    at (axis cs:\testMid,\boxmid) {Testing};

\addplot[romForecastColor, line width=2.0pt, dashed] table [x=date, y=ROM_only, col sep=comma] {Data_sheets/temporal_evolution_error_lines.csv};
\node[anchor=west, color=romForecastColor, font=\small\bfseries] (romLabel) at (axis cs: 2026-02-18, 4.0) {OpInf};
\draw[->, color=romForecastColor, line width=1pt] (romLabel.south) -- (axis cs: 2026-02-25, 3.25);

\addplot[fullstatecolor, line width=2.0pt] table [x=date, y=Assimilated_Full_state, col sep=comma] {Data_sheets/temporal_evolution_error_lines.csv};
\node[anchor=south, color=fullstatecolor, font=\small\bfseries, align=center] (fullStateLabel) at (axis cs: 2026-02-25, 2.0) {OpInf-DA \\ (Full State)};
\draw[->, color=fullstatecolor, line width=1pt] (fullStateLabel.south) -- (axis cs: 2026-02-28, 0.95);

\addplot[assimColor, line width=2.0pt] table [x=date, y=Assimilated_Sensors_12, col sep=comma] {Data_sheets/temporal_evolution_error_lines.csv};
\node[anchor=south, color=assimColor, font=\small\bfseries, align=center] (sensorLabel) at (axis cs: 2026-01-26, 2.0) {OpInf-DA \\ (12 Sensors)};
\draw[->, color=assimColor, line width=1pt] (sensorLabel.south) -- (axis cs: 2026-01-31, 0.5);

\coordinate (zoomBoxSW) at (axis cs: 2026-01-29, 0.0);
\coordinate (zoomBoxNE) at (axis cs: 2026-02-04, 1.0);
\coordinate (zoomBoxSE) at (axis cs: 2026-02-04, 0.0);
\draw[densely dotted, color=black, line width=1pt] (zoomBoxSW) rectangle (zoomBoxNE);

\end{axis}

\node[anchor=north west, font=\large\bfseries, xshift=5pt, yshift=-5pt] at (mainplot.north west) {(a)};

\begin{axis}[
    name=zoomplot,
    at={(mainplot.south west)},
    anchor=north west,
    yshift=-1.8cm,
    width=0.98\linewidth,
    height=6.5cm,
    grid=both,
    xlabel={Date},
    ylabel={Error (L2) = $\| \mathbf{x}_{\text{true}} - \Phi \widehat{\mathbf{x}} \|_2$},
    ymin=0,
    ymax=1,
    xmin=2026-01-29,
    xmax=2026-02-04,
    date coordinates in=x,
    xticklabel=\year-\month-\day,
    xticklabel style={rotate=45, anchor=east, font=\footnotesize},
    tick label style={font=\footnotesize},
    label style={font=\small},
    legend style={draw=none},
    clip=true,
    xtick distance = 1,
    ytick = {0.2, 0.4, 0.6, 0.8, 1.0}
]

\tikzset{obsline/.style={obsColor, solid, line width=1pt}}

\node[anchor=south, color=obsColor, font=\small\bfseries, rotate=90] at (axis cs: 2026-02-03, 0.18) {Observations};
\draw[obsline] (axis cs:2026-02-01,0) -- (axis cs:2026-02-01,1.0);
\draw[obsline] (axis cs:2026-02-02,0) -- (axis cs:2026-02-02,1.0);
\draw[obsline] (axis cs:2026-02-03,0) -- (axis cs:2026-02-03,1.0);
\draw[obsline] (axis cs:2026-02-04,0) -- (axis cs:2026-02-04,1.0);

\fill[corrWindowColor, opacity=0.35] (axis cs:2026-01-31,0.0) rectangle (axis cs:2026-02-01,1.0);

\node[    font=\small\bfseries,
    text=corrWindowColor,
    anchor=east,
    align=center,
] at (axis cs:2026-02-01,0.15) {CORRECTION\\WINDOW};
\draw[<-, color=corrWindowColor, line width=1.5pt] (axis cs:2026-01-31,0.05) -- (axis cs:2026-02-01,0.05);

\draw[timestamps, line width=1.2pt, dotted] (axis cs:2026-02-02,0) -- (axis cs:2026-02-02,1.0);
\node[anchor=east, font=\small\bfseries, color=timestamps] at (axis cs:2026-02-02,0.8) {T2};


\pgfmathsetmacro{\boxstartsDiv}{\boxstarts/4.75}
\pgfmathsetmacro{\boxendsDiv}{\boxends/4.75}

\path[fill=phaseboxxbrown]
    (axis cs:\validStart,\boxstartsDiv) rectangle (axis cs:\validEnd,\boxendsDiv);
\draw[draw=trainingtesting, line width=0.8pt]
    (axis cs:\validStart,\boxstartsDiv) -- (axis cs:\validStart,\boxendsDiv)
    (axis cs:\validEnd,\boxstarts) -- (axis cs:\validEnd,\boxends);

\path[fill=obslabelblue]
    (axis cs:\testStart,\boxstartsDiv) rectangle (axis cs:\testEnd,\boxendsDiv);
\draw[draw=trainingtesting, line width=0.8pt]
    (axis cs:\testStart,\boxstartsDiv) -- (axis cs:\testStart,\boxendsDiv)
    (axis cs:\testEnd,\boxstartsDiv) -- (axis cs:\testEnd,\boxendsDiv);
    \draw[obsColor, line width=1pt, densely dotted] (axis cs:2026-02-01,0) -- (axis cs:2026-02-01,1.0);
\node[anchor=south east, font=\small\bfseries, color=trainingtesting] at (axis cs:2026-02-01,0.91) {Validation};
\node[anchor=south west, font=\small\bfseries, color=trainingtesting] at (axis cs:2026-02-01,0.895) {Testing};


\addplot[romForecastColor, line width=2.0pt, dashed] table [x=date, y=ROM_only, col sep=comma] {Data_sheets/temporal_evolution_error_lines.csv};
\node[anchor=west, color=romForecastColor, font=\small\bfseries] (romLabel) at (axis cs: 2026-02-03 12:00, 0.6) {OpInf};
\draw[->, color=romForecastColor, line width=1pt] (romLabel.north) -- (axis cs: 2026-02-03 10:00, 0.81);

\addplot[fullstatecolor, line width=2.0pt] table [x=date, y=Assimilated_Full_state, col sep=comma] {Data_sheets/temporal_evolution_error_lines.csv};
\node[anchor=south, color=fullstatecolor, font=\small\bfseries, align=center] (fullStateLabel) at (axis cs: 2026-02-02 12:00, 0.01) {OpInf-DA \\ (Full State)};
\draw[->, color=fullstatecolor, line width=1pt] (fullStateLabel.north) -- (axis cs: 2026-02-02 12:00, 0.34);

\addplot[assimColor, line width=2.0pt] table [x=date, y=Assimilated_Sensors_12, col sep=comma] {Data_sheets/temporal_evolution_error_lines.csv};
\node[anchor=south, color=assimColor, font=\small\bfseries, align=center] (sensorLabel) at (axis cs: 2026-01-30, 0.7) {OpInf-DA \\ (12 Sensors)};
\draw[->, color=assimColor, line width=1pt] (sensorLabel.south) -- (axis cs: 2026-01-30 12:00, 0.43);

\end{axis}

\node[anchor=north west, font=\large\bfseries, xshift=5pt, yshift=-5pt] at (zoomplot.north west) {(b)};

\draw[densely dotted, color=black, line width=0.9pt] (zoomBoxSW) -- (zoomplot.north west);
\draw[densely dotted, color=black, line width=0.9pt] (zoomBoxSE) -- (zoomplot.north east);

\end{tikzpicture}
     \caption{Relative $L_2$ error across the training, validation, and testing
intervals for the open-loop OpInf forecast, assimilation using
$m=12$ CPQR-selected measurements, and assimilation using full-state
observations. (a) Error evolution over the complete study period; $T_2$ and
$T_3$ denote the representative testing times shown in
Figure~\ref{fig:sst_dashboard}. (b) Enlarged view of the transition into
the testing interval, showing the observation times and a representative
activated correction window.}
      \label{fig:sst_dashboard_error}

\end{figure}

Figure~\ref{fig:sst_dashboard_error} shows the global evolution of the relative
$L_2$ error and complements the spatial and pointwise comparisons in
Figures~\ref{fig:sst_dashboard} and~\ref{fig:model_evolution_time_series}. During the
testing interval, the open-loop ROM error increases progressively as the fixed
model is propagated beyond the data used for its construction. In contrast,
both assimilation configurations limit this error growth through repeated
state corrections. The $m=12$ sparse configuration remains close to the
full-state reference over much of the testing period, consistent with the
aggregate error measures reported in Table~\ref{tab:sst_metrics}. The enlarged
view in Figure~\ref{fig:sst_dashboard_error}(b) illustrates a representative
correction event, in which observational information is incorporated over the
local correction window before forecasting resumes with the same fixed ROM.

The spatial comparisons in Figure~\ref{fig:sst_dashboard} also show why the
aggregate error measures should be interpreted together with the reconstructed
fields. The assimilated predictions preserve the large-scale SST structure
more faithfully than the open-loop ROM, while residual spatial discrepancies
become more apparent at later testing times. These errors are consistent with
the fundamental limitation of the present framework: all state corrections
remain confined to the fixed POD subspace and cannot recover features that are
poorly represented by that basis.

\section{Conclusion}\label{sec:conclusion}

This work investigates whether a fixed data-driven reduced-order model can
remain useful after its open-loop forecast begins to deteriorate outside the
regime represented by its training data. Rather than retraining the reduced
operators, modifying the POD basis, or returning to the full-order model, we
use intermittently available observations to correct the evolving reduced
state through correction-windowed 4D-Var. The learned OpInf operators and
reduced basis remain fixed throughout deployment. In this setting, the
polynomial structure of Operator Inference is particularly advantageous
because it provides an explicit and inexpensive reduced-order Jacobian and
adjoint, allowing both forecasting and variational correction to be performed
entirely in reduced coordinates. Sparse measurements are incorporated through
CPQR-selected observation locations that are informative with respect to the
retained POD subspace.

The analysis establishes a connection between the correction-window horizon
and the reliability of the reduced-order variational update. We derive a bound
on the discrepancy between the OpInf gradient and the full-order gradient
projected onto the POD subspace, together with a sufficient condition under
which the reduced gradient defines a descent direction for the full-order
objective restricted to that subspace. The bound contains contributions from
forward-model error, tangent-linear-model mismatch, and backward adjoint
propagation. Restricting the physical correction horizon limits the
amplification of these errors through both the forward and adjoint dynamics.
Short correction windows therefore do not guarantee descent by themselves;
rather, they reduce the accumulation of the error terms that determine whether
the sufficient descent condition is satisfied. This provides a mathematical
basis for exploiting short-horizon ROM fidelity even when long-horizon
open-loop predictions have become inaccurate.

The numerical experiments demonstrate this behavior across problems of
increasing dimension and complexity. The Duffing oscillator provides a
controlled nonlinear setting in which the correction mechanism and the
quantities entering the theoretical analysis can be examined directly. For
Burgers' equation, the fixed open-loop OpInf model progressively loses
accuracy as it evolves into the post-shock regime, whereas intermittent
state correction repeatedly re-aligns the reduced trajectory with available
observations and limits the accumulation of forecast error. The SST example
extends the framework to a state dimension of $n=33{,}528$ represented using
$r=12$ POD modes. Using measurements at only 12 CPQR-selected spatial
locations, the corrected prediction reduces the mean absolute error from
approximately $2.8^\circ$C for the unassimilated ROM to $1.2^\circ$C,
compared with $1.1^\circ$C using full-state observations. Across the reported
error metrics, the sparse configuration recovers most of the improvement
obtained using observations over the complete spatial domain.

The gradient direction analysis was further examined numerically for both the Duffing oscillator and Burgers equation by directly comparing the ROM gradient with the projected full-order gradient. The observed angles of \(0.30^\circ\) and \(4.69^\circ\), respectively, demonstrate strong directional agreement and, importantly, satisfy the descent condition \(\theta<\pi/2\) in both cases. These results provide numerical support for the central theoretical argument: accurate variational correction does not require exact recovery of the full-order gradient, but rather sufficient preservation of its descent direction within the reduced subspace

Several limitations define the scope of the present formulation. All
corrections remain confined to the fixed POD subspace, so features that are
poorly represented by this basis cannot be recovered through state
assimilation alone. In addition, the strong-constraint formulation treats the
learned OpInf dynamics as exact within each correction window despite the
model discrepancy that motivates assimilation. Weak-constraint formulations,
joint state--parameter estimation, and adaptive or partitioned reduced spaces
therefore provide natural extensions when model-form error, parameter drift,
or unresolved dynamics become significant. The present study also assumes
linear observations and a polynomial OpInf representation; extensions to
nonlinear observation operators, broader reduced-model classes, and adaptive
triggering provide additional directions for future work.

Taken together, the results show that a data-driven ROM need not remain
accurate as an open-loop long-horizon predictor to remain useful during
deployment. If its dynamics and sensitivities retain sufficient fidelity over
a local horizon, intermittent variational state correction from sparse
observations can repeatedly limit accumulated forecast error and extend the
useful predictive horizon of the existing reduced model without retraining its
operators or modifying its reduced basis.

\appendix
\appendix
\section{Burgers' Equation Experiment 1 — Initial Condition Recovery}
\label{sec:appendix_burgers_ic}

Most Burgers-model construction details are given in Section~\ref{sec:burgers}. This appendix provides two complementary experiments based on single-window OpInf-DA initial-condition correction. The first experiment uses a separate smooth-regime initial condition to demonstrate successful initial-condition recovery, whereas the second uses the Burgers configuration from the main study to examine the limitations of single-window correction in the mixed-regime forecast.

\subsubsection*{Single-Window OpInf-DA Methodology}

The core methodology applied in both experiments is single-window OpInf-DA, which adjusts the initial reduced-order state to minimise a cost functional over the entire forecast horizon:
\begin{equation}
\begin{aligned}
  & \min_{\widehat{\mathbf{x}}_0} \mathcal{J}(\widehat{\mathbf{x}}_0) = \|\widehat{\mathbf{x}}_0 - \widehat{\mathbf{x}}_{0,\text{bg}}\|^2_{P_0^{-1}} + \sum_{k=0}^{T} \|\mathbf{y}_k - \mathbf{H}\widehat{\mathbf{x}}_k\|^2_{R^{-1}} \\
  & \text{subject to} \quad \widehat{\mathbf{x}}_{k+1} = \mathcal{M}(\widehat{\mathbf{x}}_k)
\end{aligned}
\label{eqn:global_4dvar}
\end{equation}
The adjoint propagates observation information backward over the full horizon to update the initial reduced state $\widehat{\mathbf{x}}_0$ at the start of testing. In this single-window formulation, all observations are weighted within a single objective over the entire window. This is effective when ROM fidelity is approximately uniform in time and space. When fidelity changes across the window, the same formulation can produce corrections that are optimal for the objective but not for the underlying physics.

\subsection{Smooth-Regime Success: Information Recovery from Equidistant Sensors}
\label{subsubsec:ic_recovery_success}

The first experiment tests whether adjoint-based correction can recover unobserved single-window structure from a small set of equidistant measurements. The true initial condition is a sine wave, $u(x,0)=\sin(\pi x)$, while the initial guess is clipped:
\begin{equation}
    u_{\text{guess}}(x,0) = \min(\sin(\pi x), 0.5)
\end{equation}
This clipping suppresses amplitude and introduces a biased forecast if no correction is applied. The assimilation horizon is $T=2.0$ s with three equidistant sensors (indices 15, 55, 95), observation noise $\sigma_{\text{obs}}=0.05$, and the same OpInf-ROM configuration used in the main Burgers study.

\begin{figure}[H]
    \centering
\begin{tikzpicture}[font=\small]
    \begin{groupplot}[
        group style={
            group size=2 by 1,
            horizontal sep=2cm,
        },
        width=8cm, height=6cm,
    ]

    \nextgroupplot[
        title={\textbf{(c) \textcolor{romForecastColor}{OpInf}}},
        xlabel={Time ($t$)},
        ylabel={Position ($x$)},
        xmin=0, xmax=2.0,
        ymin=0, ymax=1.0,
        enlargelimits=false,
        axis on top,
    ]
    \addplot graphics [xmin=0, xmax=2.0, ymin=0, ymax=1.0] {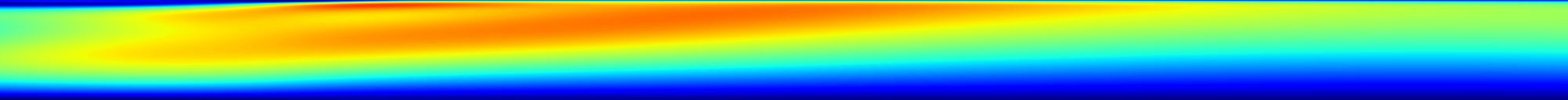};

\nextgroupplot[
        title={\textbf{(d)} \textcolor{assimColor}{\textbf{single-window}} \textcolor{assimColor}{\textbf{OpInf-DA}}},
        title style={align=center}, 
        xlabel={Time ($t$)},
        ylabel={Position ($x$)},
        xmin=0, xmax=2.0,
        ymin=0, ymax=1.0,
        enlargelimits=false,
        axis on top,
        colorbar,
        colorbar style={
            ylabel={$u(x,t)$},
            ytick={0,0.5,1.0}
        },
        point meta min=0, point meta max=1
    ]
    \addplot graphics [xmin=0, xmax=2.0, ymin=0, ymax=1.0] {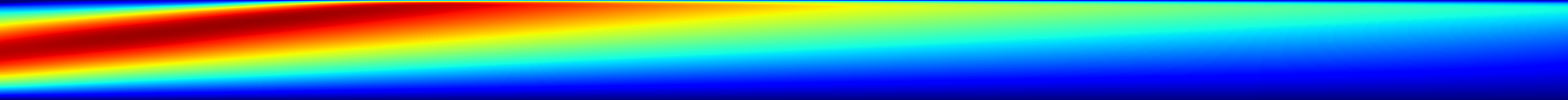};

\end{groupplot}
\end{tikzpicture}
    \caption{Initial-condition recovery via OpInf-DA in the smooth regime. (a) From a clipped initial guess, adjoint-based OpInf-DA with three sparse sensors restores the missing sine amplitude. (b) Ground-truth spatiotemporal evolution from the true initial condition. (c) Unassimilated evolution from the clipped guess, showing persistent amplitude deficit. (d) Assimilated evolution from the corrected initial condition, recovering dynamics consistent with ground truth. This demonstrates successful information recovery when ROM fidelity is sustained throughout the assimilation window.}
    \label{fig:burgers_da_1}
\end{figure}

Figure~\ref{fig:burgers_da_1} summarizes the result. Panel (a) shows that OpInf-DA recovers the missing sine amplitude, while the unassimilated guess stays clipped. Panels (b)--(d) show that this corrected initial condition yields trajectories consistent with the ground truth across the window. In this case, the full assimilation interval remains in the smooth (pre-shock) regime, where ROM training coverage is dense and model error is small. Under these conditions, sparse measurements provide enough information for effective single-window initial-condition recovery.

\begin{figure}[!htb]
    \centering
\begin{tikzpicture}[font=\small]
    \begin{groupplot}[
        group style={
            group size=2 by 1,
            horizontal sep=2cm,
        },
        width=8cm, height=6cm,
    ]

    \nextgroupplot[
        title={\textbf{(a) Initial Condition ($t=0$)}},
        xlabel={Position $x$},
        ylabel={$u(x)$},
        grid=major,
        grid style={dotted, gridcolor},
        ymin=-0.2, ymax=1.4
    ]
    
    \addplot[truthColor, ultra thick] table[x=x, y=u_true, col sep=comma] {Data_sheets/ic_data.csv};

    \addplot[romForecastColor, dashed, thick] table[x=x, y=u_guess, col sep=comma] {Data_sheets/ic_data.csv};

    \addplot[assimColor, dashdotted, thick] table[x=x, y=u_assim, col sep=comma] {Data_sheets/ic_data.csv};

    \addplot[only marks, mark=x, mark size=3pt, thick, obsColor] table[x=x_obs, y=y_obs, col sep=comma] {Data_sheets/obs_data.csv};

    \node[anchor=west, text=truthColor] (lblTruth) at (rel axis cs:0.5,0.88) {Truth};
    \draw[->, truthColor, thick] (lblTruth.south) -- (axis cs:0.62,0.93);

    \node[anchor=west, text=romForecastColor] (lblGuess) at (rel axis cs:0.4,0.25) {OpInf};
    \draw[->, romForecastColor, thick] (lblGuess.north) -- (axis cs:0.60,0.50);

    \node[anchor=west, text=assimColor, align=center] (lblAssim) at (rel axis cs:0.05,0.8) {single-window \\ OpInf-DA};
    \draw[->, assimColor, thick] (lblAssim.east) -- (axis cs:0.4,0.89);

    \node[anchor=center, text=obsColor] (lblObs) at (rel axis cs:0.78,0.9) {Observations};
    \draw[->, obsColor, thick] (lblObs.south) -- (axis cs:0.74,0.66);

    \nextgroupplot[
        title={\textbf{(b) \textcolor{truthColor}{Truth Propagation}}},
        xlabel={Time ($t$)},
        ylabel={Position ($x$)},
        xmin=0, xmax=2.0,
        ymin=0, ymax=1.0,
        enlargelimits=false,
        axis on top,
    ]
    \addplot graphics [xmin=0, xmax=2.0, ymin=0, ymax=1.0] {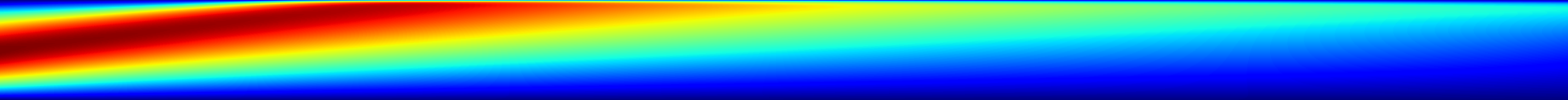};

\end{groupplot}
\end{tikzpicture}
\end{figure}

\subsection{Mixed-Regime Forecast: When single-window Correction Fails}
\label{subsubsec:global_4dvar}

The first experiment stayed entirely within a smooth regime where ROM fidelity is high. The second experiment tests the contrasting case: the ROM is still trained only on smooth-phase data, but the forecast window extends into shock formation, where error increases sharply. In this cross-regime setting, the same single-window OpInf-DA formulation does not improve the forecast quality.

The second experiment uses the same Burgers configuration as the main study,
including the initial condition $u(x,0)=\sin(2\pi x)$ and the ROM trained only
on the smooth-phase data. The assimilation is performed as a single-window
initial-condition correction over a horizon that extends into the shock-forming
regime. The observations are again obtained from three sensors (indices 15, 55,
and 95) with $\sigma_{\text{obs}}=0.05$.

\begin{figure}[H]
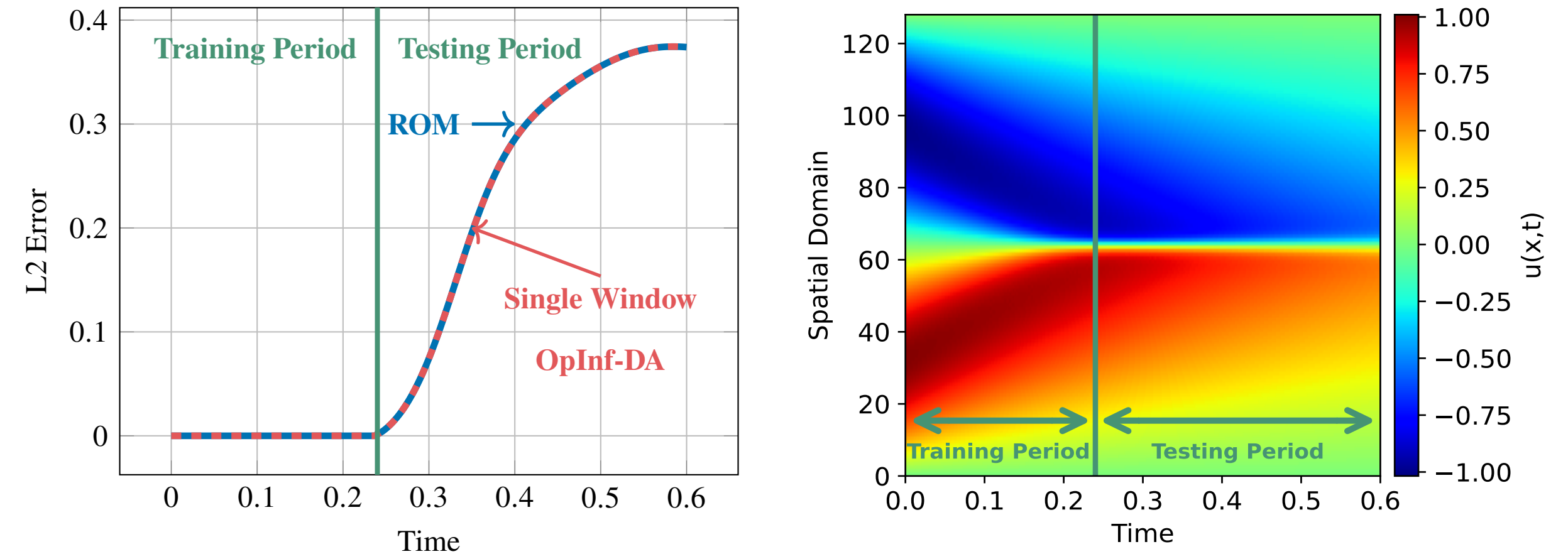
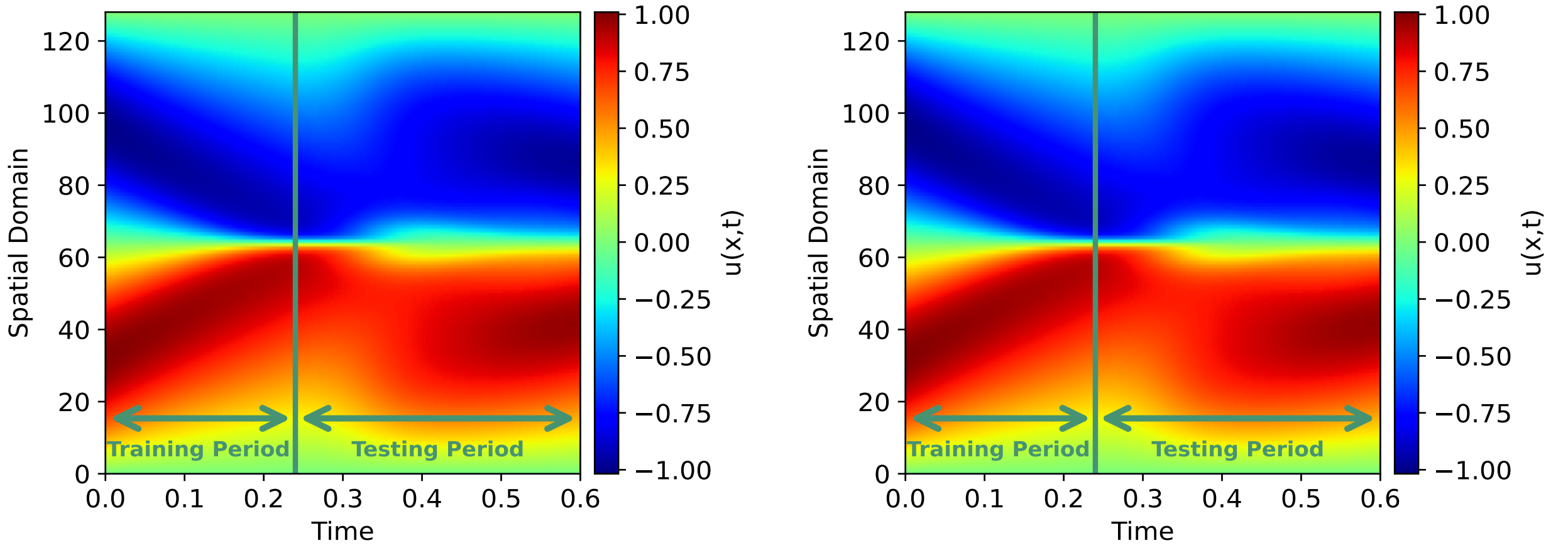

    \centering
    \begin{subfigure}[t]{0.49\linewidth}
        \centering

\begin{tikzpicture}
    \begin{axis}[
        xlabel={Time},
        ylabel={L2 Error},
        legend style={draw=none},
        grid,
        width=\textwidth,
        height=6.5cm,
        font=\small,
    ]

    \addplot [color=romForecastColor, line width=2pt, mark=none] table[x=time, y=ROM, col sep=comma] {Data_sheets/error_timeseries_ROM4.csv};




    \addplot [color=ICDAColor, line width=2pt, dashed] table[x=time, y=IC_DA, col sep=comma] {Data_sheets/error_timeseries_ROM4.csv};
    
    \draw [color=trainingtesting, line width=1.5pt] (axis cs: 0.24, -0.5) -- (axis cs: 0.24, 0.5);
    \node[anchor=east, color=trainingtesting, font=\small\bfseries] at (axis cs: 0.23, 0.37) {Training Period};
    \node[anchor=west, color=trainingtesting, font=\small\bfseries] at (axis cs: 0.25, 0.37) {Testing Period};
    
    \node[anchor=east, color=romForecastColor, font=\small\bfseries] (romLabel) at (axis cs: 0.35, 0.3) {ROM};
    \draw[->, color=romForecastColor, line width=1pt] (romLabel.east) -- (axis cs: 0.4, 0.3);
    
    
    
    
    \node[anchor=center, color=ICDAColor, font=\small\bfseries, align=center] (icLabel) at (axis cs: 0.5, 0.1) {Single Window \\ OpInf-DA};
    \draw[->, color=ICDAColor, line width=1pt] (icLabel.north) -- (axis cs: 0.35, 0.2);


    \end{axis}
\end{tikzpicture}
        \caption{Time-resolved relative $L^2$ error for the mixed-regime forecast}
        \label{fig:burgers_ic_da_error}
    \end{subfigure}
    \hfill
    \begin{subfigure}[t]{0.49\linewidth}
        \centering
        \includegraphics[width=\linewidth, height=6cm]{figures/burgers_ground_truth_ROM4.pdf}
        \caption{Ground truth solution evolution across the full time frame}
        \label{fig:burgers_truth_app}
    \end{subfigure}

    \vspace{0.5em}

    \begin{subfigure}[t]{0.49\linewidth}
        \centering
        \includegraphics[width=\linewidth, height=6cm]{figures/burgers_ROM_ROM4.pdf}
        \caption{OpInf forecast without assimilation}
        \label{fig:burgers_rom_app}
    \end{subfigure}
    \hfill
    \begin{subfigure}[t]{0.49\linewidth}
        \centering
        \includegraphics[width=\linewidth, height=6cm]{figures/burgers_initial_condition_ROM4.pdf}
        \caption{Single-window OpInf-DA Forecast}
        \label{fig:burgers_ic}
    \end{subfigure}
\caption{Single-window OpInf-DA for the mixed-regime Burgers' forecast.
(a) Time-resolved relative $L_2$ error for the open-loop ROM and the
single-window corrected forecast.
(b)--(d) Ground-truth evolution, open-loop OpInf forecast, and
single-window OpInf-DA forecast, respectively.
Although the optimized initial condition improves agreement over the earlier
portion of the window, the correction provides little improvement once the
trajectory evolves beyond the regime accurately represented by the fixed ROM.}
    \label{fig:burgers_ic_da}
\end{figure}

Figure~\ref{fig:burgers_ic_da_error} reports the time-resolved relative $L^2$ error $\frac{\|\mathbf{u}_{\text{true}} - \mathbf{u}_{\text{pred}}\|_2}{\|\mathbf{u}_{\text{true}}\|_2}$. The cumulative observation error of single-window OpInf-DA is $\sim 0.354$, effectively the same as the ROM-only case ($\sim 0.356$). The optimizer identifies an initial condition that improves the fit in the early smooth interval, but this does not translate to better performance after shock emergence. Both trajectories diverge similarly after $t \approx 0.24$ s. The limiting factor is the model form: when shock physics are outside the ROM representation, initial-condition adjustment alone cannot recover the missing dynamics.

\subsection{Analysis and Implications}

The two experiments differ mainly in the relationship between the assimilation window and ROM validity. In Equation~\ref{eqn:global_4dvar}, all observations enter a single-window objective. In the smooth regime case, this is appropriate because the ROM remains valid over the full window, so the assimilation aligns with the forecast improvement. In the mixed-regime case, shock-phase observations are filtered through a model that cannot represent relevant physics. The optimizer still reduces the chosen objective, but the resulting correction has a limited predictive value once the trajectory enters the shock regime.

The main implication is that ROM assimilation should be matched to local model reliability. A single-window correction is effective only when the training coverage and model fidelity are sustained over the entire horizon. When fidelity varies, corrections should be applied in reliability-aware windows rather than across a long interval. This motivates the windowed strategy used in the main text: partition the horizon and assimilate only where the ROM is demonstrably reliable.

\section{Equidistant vs. QR-Selected Sensor Placement}
\label{sec:appendix_sensor_compare}

Appendix \ref{sec:appendix_burgers_ic} used three equidistant sensors and showed that the single-window OpInf-DA succeeds in reliable regimes but fails across regime boundaries. This section examines whether these conclusions depend on the sensor placement. Specifically, we compared four observation configurations under the same ROM and windowed-correction setup to isolate the placement effects from the model effects. This comparison distinguishes two possibilities: performance loss from poor measurement geometry versus performance loss from limited ROM fidelity.

\begin{itemize}
    \item ROM-only forecast (no assimilation),
    \item Equidistant sensors ($m=4$ uniformly spaced points),
    \item QR-selected sensors ($m=4$ from column-pivoted QR on the POD basis),
    \item Full-state observations ($m=128$).
\end{itemize}

\begin{figure}[H]
    \centering

\begin{tikzpicture}
    \begin{axis}[
        xlabel={Time},
        ylabel={L2 Error},
        legend style={draw=none},
        grid,
        width=16cm,
        height=5cm,
        font=\small,
    ]

    \addplot [color=romForecastColor, line width=2pt, mark=none] table[x=time, y=ROM, col sep=comma] {Data_sheets/error_timeseries_ROM4.csv};

    \addplot [color=fullstatecolor, line width=2pt, mark=none] table[x=time, y=FullState_128, col sep=comma] {Data_sheets/error_timeseries_ROM4.csv};

    \addplot [color=assimColor, line width=2pt, mark=none] table[x=time, y=QR_Picked_4, col sep=comma] {Data_sheets/error_timeseries_ROM4.csv};

    \addplot [color=EqColor, line width=2pt, mark=none] table[x=time, y=Equidistant_4, col sep=comma] {Data_sheets/error_timeseries_ROM4.csv};

    
    \draw [color=trainingtesting, line width=1.5pt] (axis cs: 0.24, -0.5) -- (axis cs: 0.24, 0.5);
    \node[anchor=east, color=trainingtesting, font=\small\bfseries] at (axis cs: 0.23, 0.37) {Training Period};
    \node[anchor=west, color=trainingtesting, font=\small\bfseries] at (axis cs: 0.25, 0.37) {Testing Period};
    
    \node[anchor=east, color=romForecastColor, font=\small\bfseries] (romLabel) at (axis cs: 0.29, 0.2) {ROM};
    \draw[->, color=romForecastColor, line width=1pt] (romLabel.east) -- (axis cs: 0.34, 0.2);
    
    \node[anchor=east, color=fullstatecolor, font=\small\bfseries] (fsLabel) at (axis cs: 0.45, 0) {Full State};
    \draw[->, color=fullstatecolor, line width=1pt] (fsLabel.east) -- (axis cs: 0.5, 0.06);
    
    \node[anchor=east, color=assimColor, font=\small\bfseries] (qrLabel) at (axis cs: 0.6, 0.2) {QR-Picked};
    \draw[->, color=assimColor, line width=1pt] (qrLabel.west) -- (axis cs: 0.48, 0.13);
    
    \node[anchor=east, color=EqColor, font=\small\bfseries] (eqLabel) at (axis cs: 0.6, 0.275) {Equidistant};
    \draw[->, color=EqColor, line width=1pt] (eqLabel.north) -- (axis cs: 0.515, 0.346);
    


    \end{axis}
\end{tikzpicture}
    \caption{Effect of sensor placement on Burgers' assimilation performance.
The open-loop ROM forecast is compared with correction using $m=4$
equidistant sensors, $m=4$ CPQR-selected sensors, and full-state observations.
For the same sensor count, the CPQR-selected configuration substantially
reduces the relative $L_2$ error and closely approaches the full-state
reference, demonstrating the importance of measurement placement with respect
to the retained POD subspace.}
    \label{fig:appendix_burger_error_timeseries}
\end{figure}
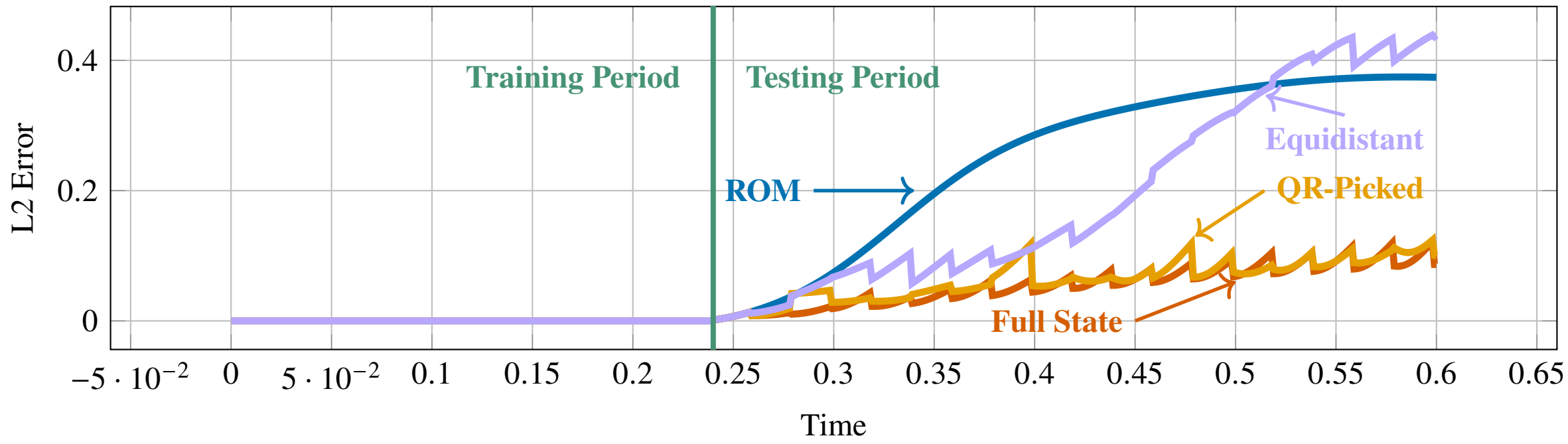

Figure~\ref{fig:appendix_burger_error_timeseries} reports the time-resolved $L^2$ error. The ROM-only forecast diverges rapidly after training. Equidistant sensing improves performance but still accumulates substantial testing error. QR-selected sensing, using the same number of sensors ($m=4$), closely tracks full-state behavior. The result shows that measurement informativeness, not only sensor count, controls correction quality. POD-informed CPQR sparse placement can recover most of the benefit of dense observations.

\bibliography{references}

\end{document}